\documentclass{article}

\usepackage{microtype}
\usepackage{multirow}
\usepackage{graphicx}
\usepackage{subfigure}
\usepackage{pifont}
\usepackage{transparent}
\usepackage{booktabs} 
\usepackage{threeparttable}
\usepackage{hyperref}
\usepackage{longtable}
\usepackage{threeparttablex}

\usepackage[accepted]{icml2024}

\usepackage{amsmath}
\usepackage{amssymb}
\usepackage{mathtools}
\usepackage{amsthm}
\usepackage{setspace}
\usepackage{verbatim}

\usepackage[capitalize,noabbrev]{cleveref}

\theoremstyle{plain}

\theoremstyle{definition}

\theoremstyle{remark}

\usepackage[textsize=tiny]{todonotes}

\icmltitlerunning{Timer: Generative Pre-trained Transformers Are Large Time Series Models}

\begin{document}

\twocolumn[
\icmltitle{Timer: Generative Pre-trained Transformers Are Large Time Series Models}

\icmlsetsymbol{equal}{*}




\begin{icmlauthorlist}
\icmlauthor{Yong Liu}{equal,software}
\icmlauthor{Haoran Zhang}{equal,software}
\icmlauthor{Chenyu Li}{equal,software}
\icmlauthor{Xiangdong Huang}{software}
\icmlauthor{Jianmin Wang}{software}
\icmlauthor{Mingsheng Long}{software}
\end{icmlauthorlist}

\icmlaffiliation{software}{School of Software, BNRist, Tsinghua University. Yong Liu $<$liuyong21@mails.tsinghua.edu.cn$>$. Haoran Zhang $<$z-hr20@mails.tsinghua.edu.cn$>$. Chenyu Li $<$lichenyu20@mails.tsinghua.edu.cn$>$}
\icmlcorrespondingauthor{Mingsheng Long}{mingsheng@tsinghua.edu.cn}

\icmlkeywords{Large time series models, Data-centric AI, Self-supervised pre-training, Transformers}

\vskip 0.3in
]



\printAffiliationsAndNotice{\icmlEqualContribution}  

\begin{abstract}
Deep learning has contributed remarkably to the advancement of time series analysis. Still, deep models can encounter performance bottlenecks in real-world data-scarce scenarios, which can be concealed due to the performance saturation with small models on current benchmarks. Meanwhile, large models have demonstrated great powers in these scenarios through large-scale pre-training. Continuous progress has been achieved with the emergence of large language models, exhibiting unprecedented abilities such as few-shot generalization, scalability, and task generality, which are however absent in small deep models. To change the status quo of training scenario-specific small models from scratch, this paper aims at the early development of \emph{large time series models} (LTSM). During pre-training, we curate large-scale datasets with up to 1 billion time points, unify heterogeneous time series into \emph{single-series sequence} (S3) format, and develop the GPT-style architecture toward LTSMs. To meet diverse application needs, we convert forecasting, imputation, and anomaly detection of time series into a unified \emph{generative task}. The outcome of this study is a Time Series Transformer (Timer), which is generative pre-trained by next token prediction and adapted to various downstream tasks with promising capabilities as an LTSM. Code and datasets are available at: \href{https://github.com/thuml/Large-Time-Series-Model}{https://github.com/thuml/Large-Time-Series-Model}.

\end{abstract}

\section{Introduction}\label{sec:intro}
\begin{figure}[t]
\begin{center}
    \centerline{\includegraphics[width=.98\columnwidth]{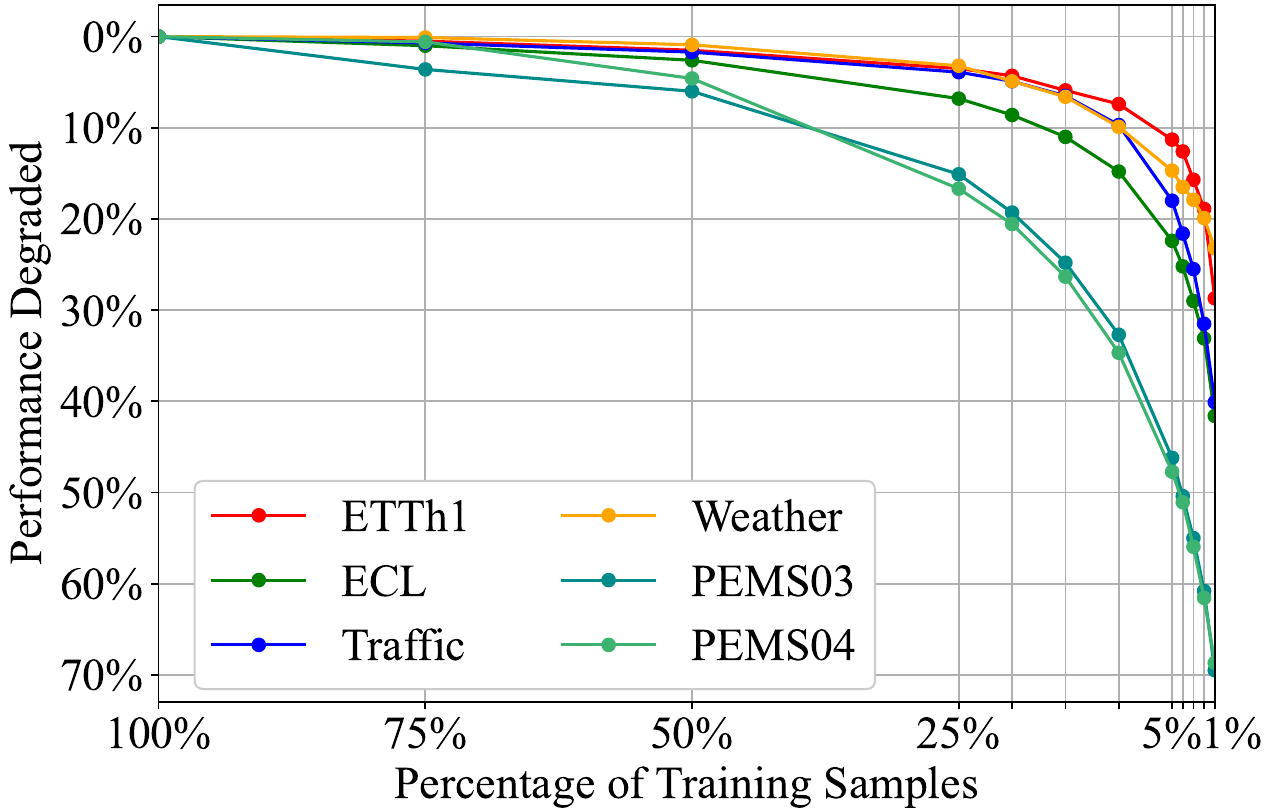}}
    \vspace{-10pt}
	\caption{Performance of PatchTST~\citeyearpar{nie2022time} on different data scarcities. The degradation is reported as the relative increase in MSE compared with training on full samples.}
	\label{fig:saturation}
\end{center}
\vspace{-30pt}
\end{figure}

Time series analysis encompasses a broad range of critical tasks, including time series forecasting~\cite{box2015time}, imputation~\cite{friedman1962interpolation}, anomaly detection~\cite{breunig2000lof}, etc. Despite the ubiquity of real-world time series, training samples can be scarce in specific applications. While remarkable advances have been made in deep time series models~\cite{wu2022timesnet, zeng2023transformers, liu2023itransformer}, the accuracy of state-of-the-art deep models~\cite{nie2022time} can still deteriorate drastically in such scenarios, even within prevalent benchmarks as shown in Figure~\ref{fig:saturation}. Concurrently, we are witnessing rapid progress of large language models~\cite{radford2018improving}, involving training on large-scale text corpora and exhibiting remarkable few-shot and zero-shot capabilities~\cite{radford2019language}. It can be indicative for the community to develop large time series models (LTSM) that are transferable on various data-scarce scenarios by pre-training on numerous time series data.

Further, large models evolved by generative pre-training (GPT) have demonstrated several advanced capabilities that are absent in small models: the generalization ability that one model fits many domains, the versatility that one model copes with various scenarios and tasks, and the scalability that performance improves with the scale of parameters and pre-training corpora. Fascinating capabilities have fostered the advancement of artificial general intelligence~\cite{openai2023gpt}. Time series holds comparable practical value to natural language. Essentially, they exhibit inherent similarities in generative modeling~\cite{bengio2000neural} and autoregression~\cite{box2013box}. Consequently, the unprecedented success of the generative pre-trained large language models~\cite{zhao2023survey} serves as a blueprint for the progress of LTSMs.

Although unsupervised pre-training on time series data has been widely explored, yielding breakthroughs based on the masked modeling~\cite{zerveas2021transformer} and contrastive learning~\cite{woo2022cost}, there are still unsolved fundamental issues for developing LTSMs. Firstly, the dataset infrastructure and unified treatment for heterogeneous time series are lagging behind other fields. As a result, prior unsupervised pre-training methods are typically constrained to a small scale and primarily focus on in-dataset transfer~\cite{zhang2022self, nie2022time}. Secondly, the architecture of scalable large models remains underexplored in the field of time series. It is observed that non-autoregressive structures, which are prevalent and effective in small time series models, may not be suitable for LTSMs. Thirdly, existing large-scale pre-trained models~\cite{woo2023pushing, das2023decoder} primarily concentrated on a single task (e.g., forecasting), and have scarcely addressed task unification. Consequently, the applicability of LTSMs remains elevatable.

In this paper, we dive into the pre-training and adaptation of large time series models. By aggregating publicly available time series datasets and following curated data processing, we construct \emph{Unified Time Series Dataset (UTSD)} of hierarchical capacities to facilitate the research on the scalability of LTSMs. To pre-train large models on heterogeneous time series data, we propose the \emph{single-series sequence (S3)} format that converts multivariate series with reserved patterns into unified token sequences. For better generalization and versatility, we adopt the GPT-style objective that predicts the next token~\cite{bengio2000neural}. Eventually, we present \textbf{Timer}, a large-scale pre-trained \textbf{Time} Series Transfor\textbf{mer}. Unlike prevalent encoder-only architecture~\cite{nie2022time, wu2022timesnet, das2023long}, Timer exhibits similar characteristics as large language models such as flexible context length and autoregressive generation. It also presents notable few-shot generalization, scalability, and task generality, outperforming state-of-the-art task-specific models on forecasting, imputation, and anomaly detection. Overall, our contributions can be summarized as follows:

\begin{itemize}
    \item We delve into the LTSM development by curating large-scale datasets comprised of 1B time points, proposing a unified sequence format to cope with data heterogeneity, and presenting Timer, a generative pre-trained Transformer for general time series analysis.
    \item We apply Timer on various tasks, which is realized in our unified generative approach. Timer exhibits notable feasibility and generalization in each task, achieving state-of-the-art performance with few samples.
    \item By pre-training on increasing available time series data, Timer exhibits zero-shot forecasting capability. Quantitative evaluations and quality assessments are provided among concurrent large time series models.
\end{itemize}

\begin{figure*}[ht]
\begin{center}
    \center{\includegraphics[width=\textwidth]{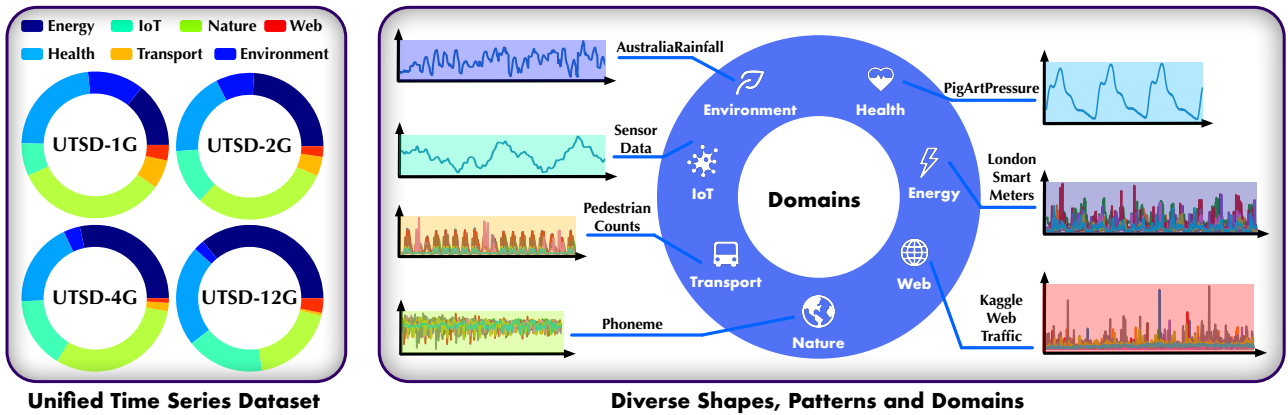}}
    \vspace{-20pt}
	\caption{Illustration of Unified Time Series Dataset (UTSD) that is composed of various time series domains with hierarchical capacities.}
	\label{fig:utsd}
\end{center}
\vspace{-10pt}
\end{figure*}

\section{Related Work}
\label{sec:related}

\subsection{Unsupervised Pre-training on Sequences}
Unsupervised pre-training on large-scale data is the essential step for modality understanding for downstream applications, which has achieved substantial success in sequences, covering natural language~\cite{radford2021learning}, patch-level image~\cite{bao2021beit} and video~\cite{yan2021videogpt}. Supported by powerful backbones~\cite{vaswani2017attention} for sequential modeling, the paradigms of unsupervised pre-training on sequences have been extensively studied in recent years, which can be categorized into the masked modeling~\cite{devlin2018bert}, contrastive learning~\cite{chen2020simple}, and generative pre-training~\cite{radford2018improving}.

Inspired by significant progress achieved in relevant fields, masked modeling and contrastive learning have been well-developed for time series. TST~\cite{zerveas2021transformer} and PatchTST~\cite{nie2022time} adopt the BERT-style masked pre-training to reconstruct several time points and patches respectively. LaST~\cite{wang2022learning} proposes to learn the representations of decomposed time series based on variational inference. Contrastive learning is also well incorporated in prior works~\cite{woo2022cost, yue2022ts2vec}. TF-C~\cite{zhang2022self} constrains the time-frequency consistency by temporal variations and frequency spectrums. SimMTM~\cite{dong2023simmtm} combines masked modeling and contrastive approach within the neighbors of time series.

However, generative pre-training has received relatively less attention in the field of time series despite its prevalence witnessed in developing large language models~\cite{touvron2023llama, openai2023gpt}. Most large language models are generative pre-trained~\cite{zhao2023survey} with token-level supervision, where each token is generated based on the previous context and independently supervised~\cite{bengio2000neural}. Consequently, they are not constrained by specific input and output lengths and excel at multi-step generation. Furthermore, prior studies~\cite{wang2022language, dai2022can} have demonstrated that scalability and generalization largely stem from generative pre-training, which requires more training data than other pre-training paradigms. Thus, our work aims to investigate and revitalize generative pre-training towards LTSMs, facilitated by extensive time series and deftly designed adaptation on downstream tasks.

\subsection{Large Time Series Models}
Pre-trained models with scalability can evolve to large foundation models~\cite{bommasani2021opportunities}, featured by increasing model capacity and pre-training scale to solve various data and tasks. Large language models even demonstrate advanced capabilities such as in-context learning and emergent abilities~\cite{wei2022emergent}. As of present, research on large time series models remains at a nascent stage. Existing efforts towards LTSMs can be categorized into two groups, with one being large language models for time series. FPT~\cite{zhou2023one} regards GPT-2 as a representation extractor of sequences, which is respectively fine-tuned on different downstream tasks. LLMTime~\cite{chang2023llm4ts} encodes time series into numerical tokens for LLMs, exhibiting model scalability in the forecasting task. Time-LLM~\cite{jin2023time} investigates prompting techniques to enhance prediction, demonstrating the generalization ability of LLMs. Unlike these methods, Timer is pre-trained natively on time series and free from extra modality alignment.

Another category includes pre-trained models on large-scale time series. ForecastFPN~\cite{dooley2023forecastpfn} is trained on synthetic series for zero-shot forecasting. CloudOps~\cite{woo2023pushing} adopts masked modeling on Transformer for domain-specific forecaster. Lag-Llama~\cite{rasul2023lag} is a probabilistic univariate forecaster that adopts lags as covariates. PreDcT~\cite{das2023decoder} is a decoder-only Transformer pre-trained on Google Trends, exhibiting notable zero-shot ability. TimeGPT-1~\cite{garza2023timegpt} releases the first commercial API for zero-shot forecasting. Different from prior works, our UTSD contains 1B real-world time points, which is not a simple aggregation but follows curated data processing. Timer is applicable to downstream tasks beyond forecasting and exhibits promising scalability. We are also the first to establish a zero-shot forecasting benchmark on concurrent LTSMs. 

\section{Approach}\label{sec:approach}
Inspired by the sequential structure inherent in language and time series, we leverage the advancement of large language models for developing LTSMs. In this paper, we advocate the development of large models for time series with (1) the utilization of extensive time series corpora, (2) the adoption of a standardized format for diverse time series data, and (3) the generative pre-training on the decoder-only Transformer that autoregressively predict the next time series token.

\subsection{Data}
Large-scale datasets are of paramount importance for pre-training large models. However, the curation of time series datasets can be prohibitively challenging. In spite of their ubiquity, there are numerous data of low quality, including missing values, unpredictability, variance in shape, and irregular frequencies, which significantly impact the efficacy of pre-training. Therefore, we establish the criteria for filtering high-quality data and stacking up the hierarchy of time series corpora. Concretely, we record the statistics of each dataset, including (1) basic properties, such as time steps, variate number, file size, frequency, etc; and (2) time series characteristics: periodicity, stationarity, and predictability. This also allows us to assess the complexity of different datasets and progressively conduct scalable pre-training. 

We curate Unified Time Series Dataset (UTSD) as shown in Figure~\ref{fig:utsd}. UTSD is constructed with hierarchical capacities to facilitate the scalability research of large models. UTSD encompasses seven domains with up to 1 billion time points (UTSD-12G), covering typical scenarios of time series analysis. Following the principle of keeping pattern diversity, we include as diverse datasets as possible in each hierarchy, ensure the data size of each domain is nearly balanced when scaling up, and the complexity gradually increases in accordance with the calculated statistics. We release four volumes on \href{https://huggingface.co/datasets/thuml/UTSD}{https://huggingface.co/datasets/thuml/UTSD}.

Notably, we make our curation applicable to the increasing open-source datasets, which is beneficial for the continuous expansion of time series corpora. Particularly, we conduct the same procedure on the recent LOTSA~\cite{woo2024unified}, a great endeavor with 27B time points, to explore zero-shot forecasting and establish the benchmark of LTSMs. Detailed construction and statistics are provided in Appendix~\ref{sec:datasets_detail}.

\subsection{Training Strategy}
Different from natural language, which has been facilitated by the well-established discrete tokenization and sequential structure with the regular shape, constructing unified time series sequences is not straightforward due to the heterogeneity of series such as amplitude, frequency, stationarity, and disparities of the datasets in the variate number, series length and purpose. To promote pre-training on extensive time series, we propose to convert heterogeneous time series into \emph{single-series sequence (S3)}, which reserves the patterns of series variations with the unified context length.

\begin{figure}[ht]
\begin{center}
    \centerline{\includegraphics[width=1\columnwidth]{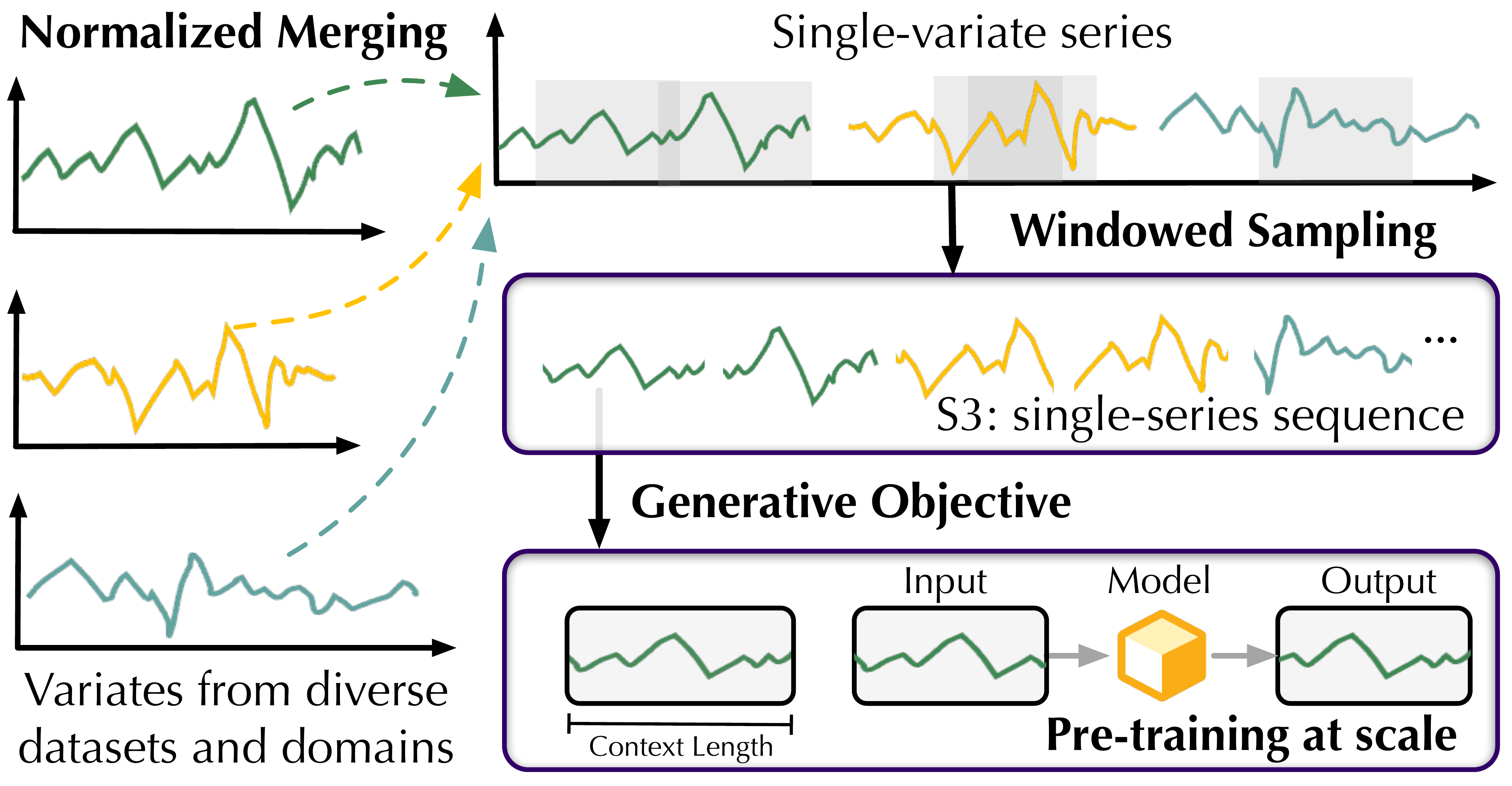}}
    \vspace{-10pt}
	\caption{Pre-training strategy for heterogeneous time series.}
	\label{fig:s3}
\end{center}
\vspace{-20pt}
\end{figure}

As depicted in Figure~\ref{fig:s3}, our initial step involves normalizing and merging at the level of variates. Each series representing a variate will be divided into training and validation splits at a ratio of 9:1 for pre-training. We apply the statistics of the training split to normalize the entire series. The normalized time series are merged into a pool of single-variate series. The time points of single-variate series for training follow the normal distribution, which mitigates the discrepancies in the amplitude and variate numbers across multiple datasets.

We uniformly sample sequences from the pool by a window, obtaining single-series sequences with a fixed context length, as the format of S3. The proposed format is essentially an extension of Channel Independence CI~\cite{nie2022time}. However, CI necessitates time-aligned multivariate series and flattens the variate dimension to the same batch, thereby requiring the batch of samples to originate from the same dataset. Based on our format, the model observes sequences from different periods and different datasets, thus increasing the pre-training difficulty and directing more attention to the temporal variation. S3 does not require time alignment, which applies to broad univariate and irregular time series. We then employ generative pre-training, where single-series sequences are regarded as standard sentences of time series. 

\subsection{Model Design}
\begin{figure}[ht]
\begin{center}
    \centerline{\includegraphics[width=\columnwidth]{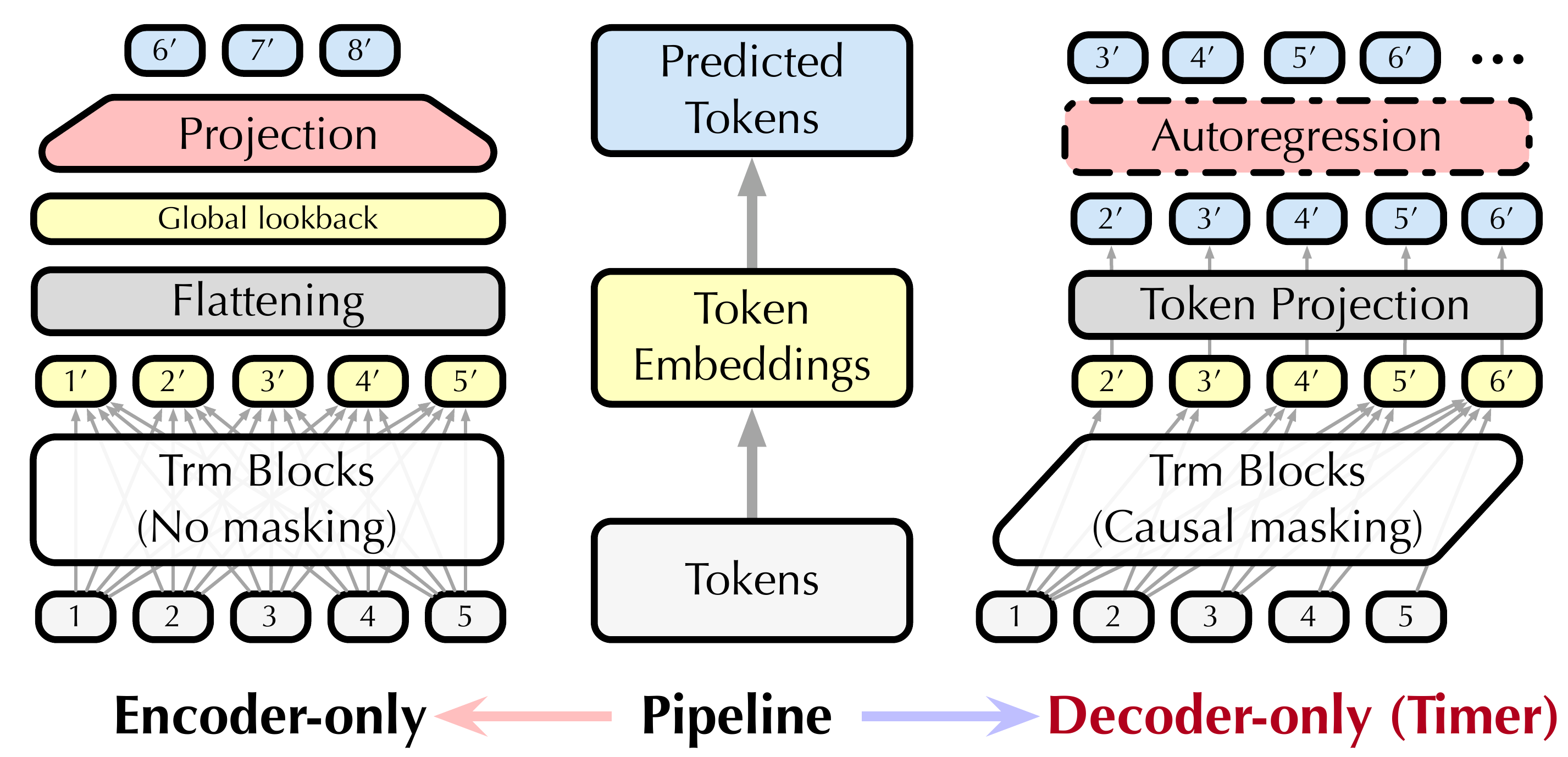}}
	\vspace{-10pt}
    \caption{Architectures of typical Transformer-based forecasters.}
	\label{fig:compare_arch}
\end{center}
\vspace{-15pt}
\end{figure}

Given the limited exploration of the backbone for large time series models, we extensively evaluate candidate backbones on the same pre-training scale in Section~\ref{sec:arch_compare}, which validates Transformer as the scalable choice. Further, we review Transformer-based models in time series forecasting, which have experienced notable development in recent years. They can be categorized into encoder-only and decoder-only architectures following a similar pipeline. As illustrated in Figure~\ref{fig:compare_arch}, prevalent small time series forecasters, the encoder-only non-autoregressive models, generate predictions with the globally flattened representation of lookback series. Although direct projection may benefit from end-to-end supervision, flattening can also wipe out sequential dependencies modeled by attention, thereby weakening Transformer layers to reveal the patterns of temporal variations. 

Inspired by the substantial progress of decode-only LLMs with the ability for iterative generation, we opt for an underexplored autoregressive approach for generative pre-training. As language models autoregressively predict the next token: 
\begin{equation}
P(\mathcal{U}) = \prod_{i=1}^N p(u_i|u_{<i})
\end{equation}
on the token sequence $\mathcal{U}=\{u_1, \dots, u_N\}$, we first establish the tokenization of the given single-series sequence (S3) $\mathbf{X}=\{x_1, \dots, x_{NS}\}$ with the unified context length $NS$. We define the time series token as consecutive time points (segment) of length $S$ that encompass the series
variations:
\begin{equation}
\mathbf{s}_i=\{x_{(i-1)S+1},\dots,x_{iS}\}\in\mathbb{R}^{S}, \ i=1, \dots, N.
\end{equation}
We adopt the decoder-only Transformer with dimension $D$ and $L$ layers and apply generative pre-training (GPT) on $N$ tokens in the single-series sequence (sentence):
\begin{equation}
\begin{aligned}
\mathbf{h}_i^0 & = \mathbf{W}_e \mathbf{s}_i + \mathbf{TE}_i, \ i=1, \dots, N, \\
\mathbf{H}^l   & = \operatorname{TrmBlock}(\mathbf{H}^{l-1}), \ l=1, \dots, L, \\
\{\hat{\mathbf{s}}_{i+1}\} & = \mathbf{H}^L \mathbf{W}_d, \ i=1, \dots, N,
\end{aligned}
\end{equation}
where $\mathbf{W}_e, \mathbf{W}_d \in \mathbb{R}^{D\times S}$ encode and decode token embeddings in $\mathbf{H}=\{\mathbf{h}_i\}\in\mathbb{R}^{N\times D}$ indepedently, and $\mathbf{TE}_i$ denotes the optional timestamp embedding. Via the causal attention of the decoder-only Transformer, the autoregressively generated $\hat{\mathbf{s}}_{i+1}$ is obtained as the next token of $\mathbf{s}_{i}$. Thus, we formulate the pre-training objective as follows:
\begin{equation}\label{equ:objective}
\mathcal{L}_{\text{MSE}} = \frac{1}{NS} \sum ||\mathbf{s}_i-\hat{\mathbf{s}}_i||_2^2,\ i=2, ... ,N.
\end{equation}
Equation~\ref{equ:objective} yields token-wise supervising signals, where generated tokens of each position are independently supervised. Consequently, generative pre-trained models are endowed with the flexibility to address unfixed context length during inference and excel at multi-step generation by iteratively sliding and enlarging input tokens. While small time series models generally refrain from iterative multi-step prediction to mitigate error accumulations, our experiments reveal that autoregressive models pre-trained on large-scale data can also perform as competitively as direct multi-step predictors.

\section{Experiments}\label{sec:exp}
\begin{figure}[ht]
\begin{center}
    \centerline{\includegraphics[width=\columnwidth]{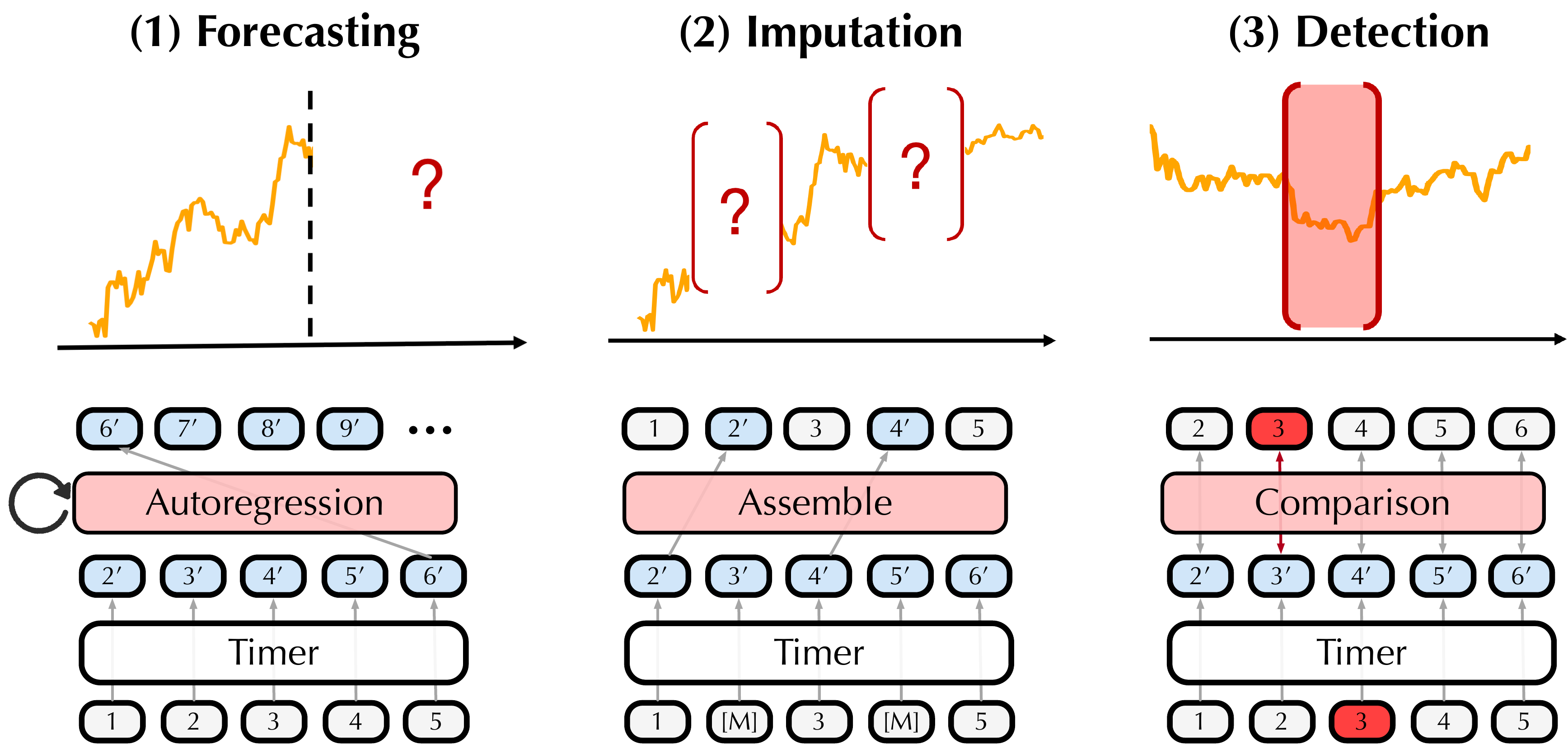}}
	\vspace{-5pt}
    \caption{Illustration of our generative task unification: (1) Generative pre-trained Timer can naturally predict the next series by the iterative autoregression; (2) By introducing masked tokens during adaptation, Timer generates imputations with the previous context and assemble them with the observed part; (3) We propose predictive anomaly detection by predicting normal series in advance.}
	\label{fig:tasks}
\end{center}
\vspace{-30pt}
\end{figure}

We demonstrate Timer as a large time series model in time series forecasting, imputation, and anomaly detection by tackling them in a unified generative scheme, which is described in Figure~\ref{fig:tasks}. We compare Timer with state-of-the-art task-specific models and present the benefit of pre-training on data-scarce scenarios, known as the few-shot ability of large models. Furthermore, we delve into the model scalability, including model/data size, and try to build a comprehensive zero-shot evaluation across concurrent large time series models. All the downstream datasets are not included in the pre-training stage to prevent data leakage. We provide the detailed implementation and model configurations of pre-training and adaptation in Appendix~\ref{sec:pretrain_detail} and~\ref{sec:finetune_detail}.

\vspace{-5pt}

\subsection{Time Series Forecasting}
\paragraph{Setups} Time series forecasting is essential and presents challenges in real-world applications. To thoroughly evaluate the performance, we elaborately establish the benchmark, including ETT, ECL, Traffic, Weather, and PEMS adopted in~\citet{liu2023itransformer}. We adopt the unified lookback length as $672$ and the forecast length as $96$. We pre-training Timer on UTSD-12G with the segment length $S=96$ and the number of tokens $N=15$, such that Timer can deal with time series with a context length up to $1440$. The downstream forecasting task can be naturally completed as the next token prediction, which is detailed in Appendix~\ref{sec:forecast_detail}.

\begin{figure}[ht]
\begin{center}
    \centerline{\includegraphics[width=0.5\textwidth]{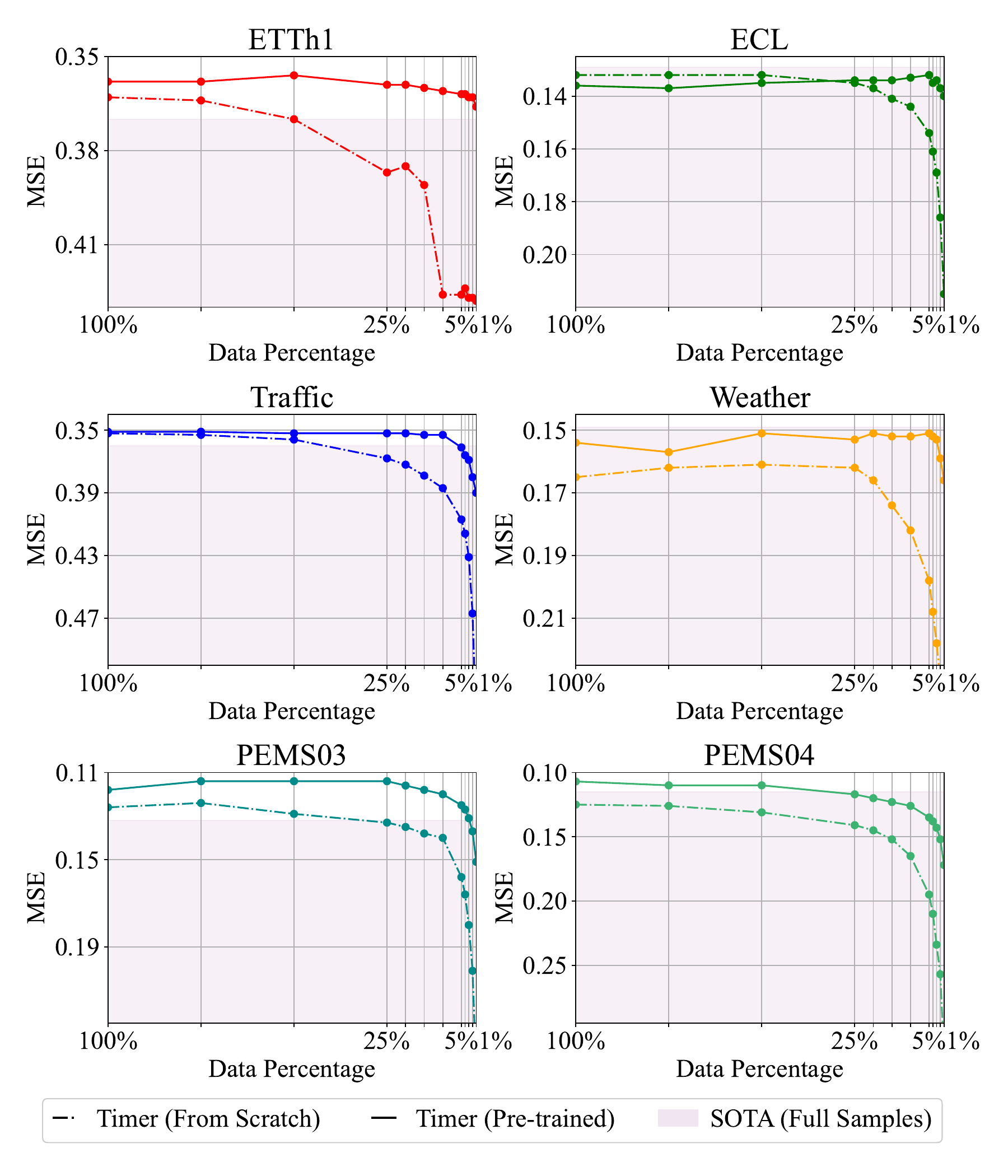}}
    \vspace{-5pt}
	\caption{Forecasting performance of Timer obtained by training from scratch and fine-tuning from the pre-trained model on different data scarcities. State-of-the-art small deep forecasters trained on full samples are provided the SOTA baseline. A smaller MSE indicates better results. Detailed results are provided in Table~\ref{tab:forecast_promotion_full}.} 
	\label{fig:forecast_promotion}
\end{center}
\vspace{-30pt}
\end{figure}

\begin{figure*}[ht]
\begin{center}
    \center{\includegraphics[width=\textwidth]{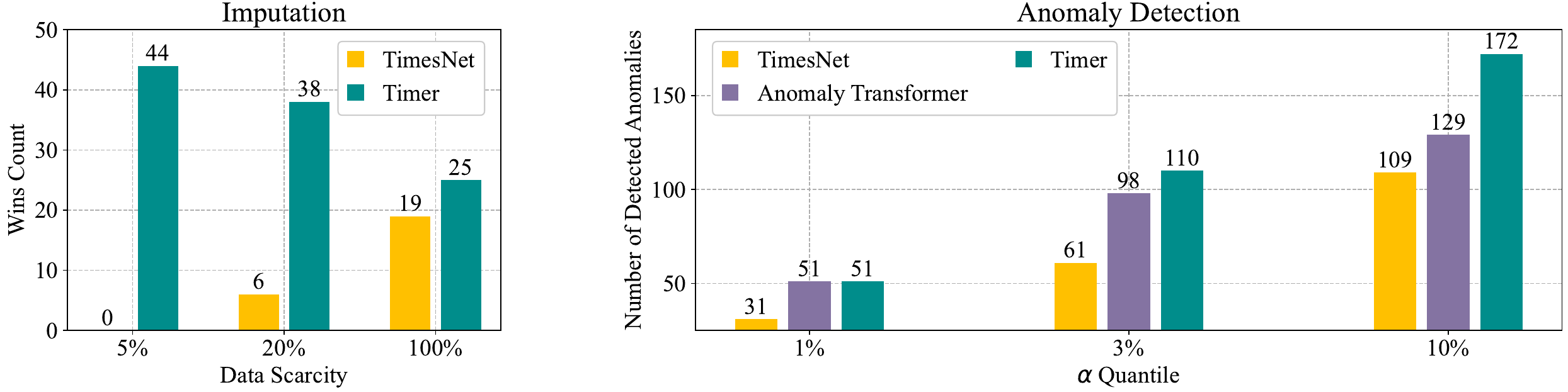}}
    \vspace{-25pt}
	\caption{Performance comparison with state-of-the-art small deep models. For imputation, Time is compared with TimesNet~\cite{wu2022timesnet} under different data scarcities, each of which contains $44$ imputation scenarios. In UCR Anomaly Detection Archive~\cite{wu2021current}, we compare the number of detected anomalies under given confidence quantiles. Detailed results are provided in Table~\ref{tab:imputation_full}-\ref{tab:anomaly_detection}.} 
	\label{fig:imputation_anomaly_detection}
\end{center}
\vspace{-10pt}
\end{figure*}

\paragraph{Results} As depicted in Figure~\ref{fig:forecast_promotion}, we present the results of the pre-trained Timer (solid line) and Timer trained from scratch (dashed line) under different data scarcities. We also evaluate state-of-the-art forecasters by training them on full samples as a competitive baseline. Concretely, we train PatchTST~\cite{nie2022time} and iTransformer~\cite{liu2023itransformer} on each dataset and report the better one as SOTA. Timer fine-tuned on few training samples demonstrates competitive results as advanced small deep forecasters, specifically achieving better results with only $1\%$ available samples from ETTh1, $5\%$ from Traffic, $3\%$ from PEMS03, and $25\%$ from PEMS04 and exhibiting remarkable few-shot ability.

To assess pre-training benefit, we compare solid and dashed lines, differing by whether to load the pre-trained checkpoint. Concretely, the performance of training a random-initialized Timer on full samples can be achieved by fine-tuning our pre-trained Timer with only $2\%$ of the training samples in ETTh1, $5\%$ in ECL, $1\%$ in Weather, and $4\%$ in PEMS03, which exemplifies the transferable knowledge acquired by pre-training on UTSD. When all samples are available, the performance of the pre-trained Timer can also outperform training it from scratch: the prediction error is reduced as $0.165\to 0.154$ on Weather, $0.126\to 0.118$ on PEMS03, and $0.125\to 0.107$ on PEMS04. Overall, in widespread data-scarce scenarios, the performance degradation can be alleviated by the few-shot generalization of LTSMs.

\subsection{Imputation}

\paragraph{Setups} Imputation is ubiquitous in real-world applications, aiming to fill corrupted time series based on partially observed data. However, while various machine learning algorithms and simple linear interpolation can effectively cope with the corruptions randomly happening at the point level, real-world corruptions typically result from prolonged monitor shutdowns and require a continuous period of recovery. Consequently, imputation can be ever challenging when attempting to recover a span of time points encompassing intricate series variations. In this task, we conduct the segment-level imputation. Each time series is divided into $8$ segments and each segment has the length of $24$ and the possibility of being completely masked. We obtain Timer on UTSD-4G by generative pre-training with the segment length $S=24$ and the token number $N=15$. For downstream adaptation, we conduct the denoising autoencoding in T5~\cite{raffel2020exploring} as detailed in Appendix~\ref{sec:imputation_detail} to recover the masked spans in a generative way.

\begin{figure}[ht]
\begin{center}
    \centerline{\includegraphics[width=0.5\textwidth]{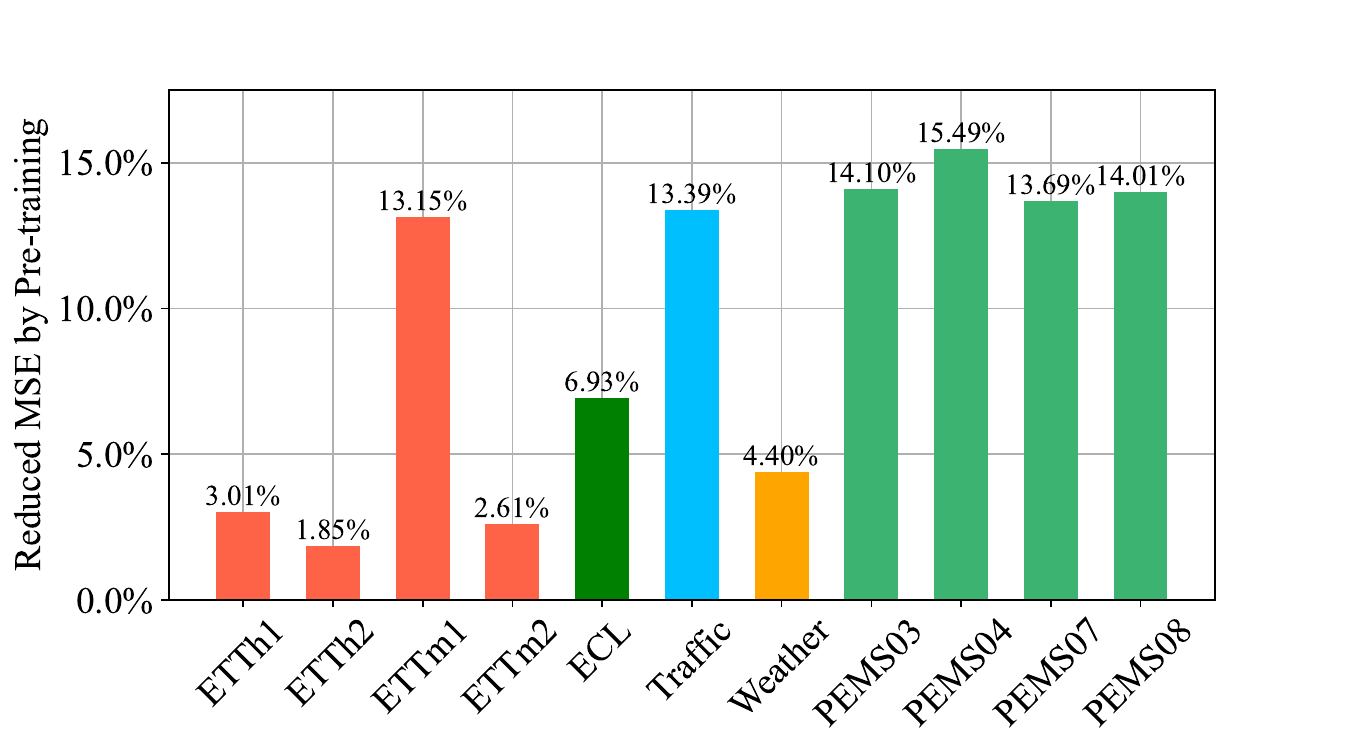}}
    \vspace{-12pt}
	\caption{Pre-training benefit of Timer on the downstream imputation task with $5\%$ available samples. Following TimesNet~\cite{wu2022timesnet}, each dataset is imputed with four mask ratios in $\{12.5\%, 25\%, 37.5\%, 50\%\}$ and we calculate the average reduced imputation error in MSE relative to training from scratch. Additional results of other data scarcities are provided in Figure~\ref{fig:imputation_full}.} 
	\label{fig:imputation}
\end{center}
\vspace{-23pt}
\end{figure}

\paragraph{Results} 
We establish a comprehensive segment-level imputation benchmark, which includes $11$ datasets with four mask ratios each. Timer is compared with the previous state-of-the-art imputation model~\cite{wu2022timesnet}. As shown in the left of Figure~\ref{fig:imputation_anomaly_detection}, Timer outperforms in respectively $100.0\%$, $86.4\%$, and $56.8\%$ of $44$ imputation scenarios under the data scarcities of $\{5\%, 20\%, 100\%\}$, validating the effectiveness of Timer on the challenging imputation task. Regarding the benefit of pre-training, we present the promotion as the reduction ratio of imputation errors in Figure~\ref{fig:imputation}, where pre-training consistently yields positive effects with $5\%$ downstream samples. Additional experiments on $20\%$ and $100\%$ available samples are provided in Figure~\ref{fig:imputation_full}, which still present notable performance improvement.

\subsection{Anomaly Detection}

\paragraph{Setups} Anomaly detection is vital in industry and operations. Previous methods~\cite{xu2021anomaly, wu2022timesnet} typically tackle the unsupervised scenario in a reconstructive approach, where a model is trained to reconstruct the input series, and the output is regarded as the normal series. Based on our generative model, we cope with anomaly detection in a predictive approach, which utilizes the observed segments to predict the future segment, and the predicted segment will be established as the standard to be compared with the actual value received. Unlike the previous method requiring to collect time series of a period for reconstruction, our predictive approach allows for segment-level anomaly detection on the fly. Thus, the task is converted into a next token prediction task as detailed in Appendix~\ref{sec:anomaly_detail}. 

We introduce UCR Anomaly Archive~\cite{wu2021current} that contains $250$ tasks. In each task, a single normal time series is provided for training, and the model should locate the position of an anomaly in the test series. We first train a predictive model on the training set and calculate the MSE between the predicted series and ground truth on the test set. By regarding the MSE of all segments as the confidence level, the segments with higher than $\alpha$ quantile of confidence are labeled as potential positions of anomalies. 

\begin{figure}[ht]
\begin{center}
    \centerline{\includegraphics[width=0.5\textwidth]{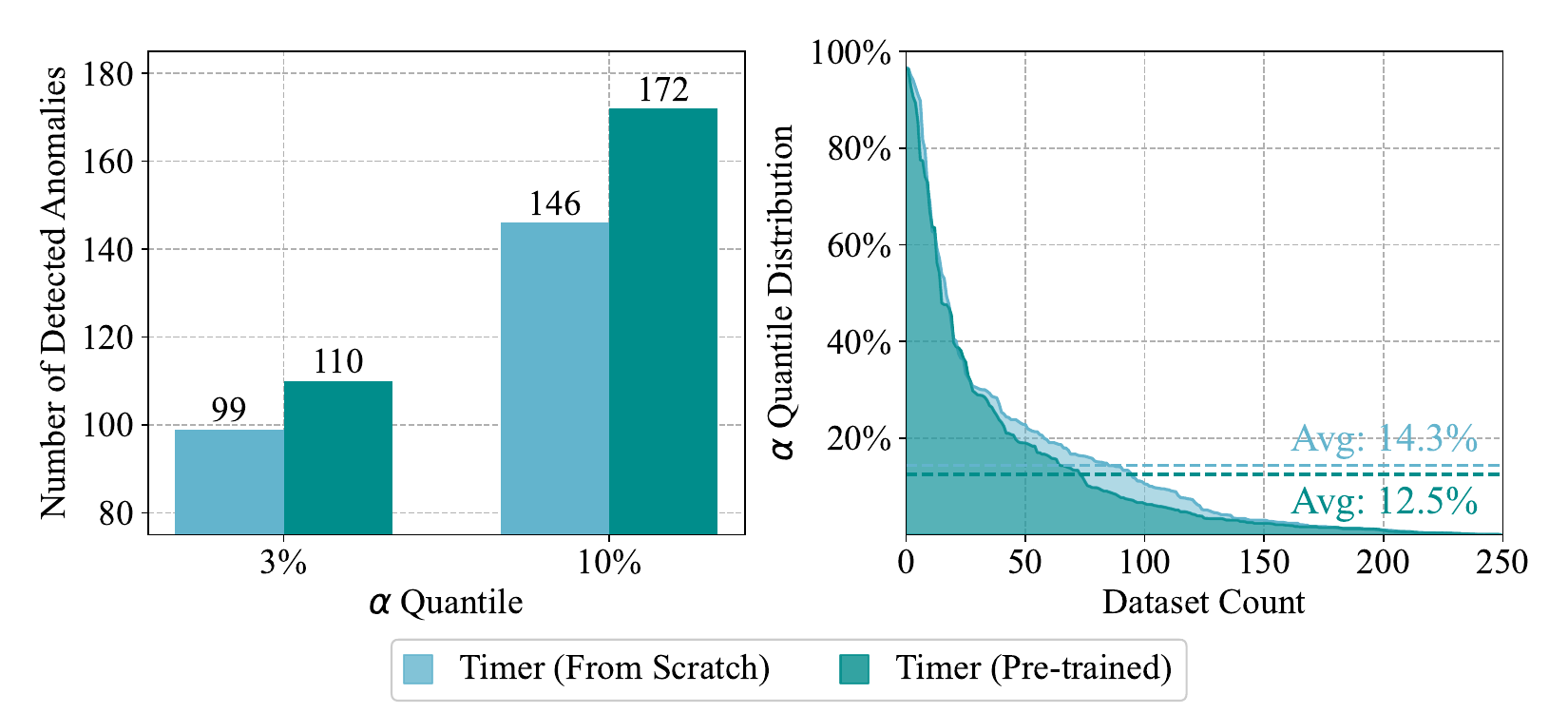}}
    \vspace{-8pt}
	\caption{Downstream anomaly detection results of Timer obtained by training from scratch and adapting with the pre-trained model.}
	\label{fig:anomaly_detection}
\end{center}
\vspace{-20pt}
\end{figure}

\paragraph{Results} 
We evaluate well-acknowledged anomaly detection models, including TimesNet~\cite{wu2022timesnet} and Anomaly Transformer~\cite{xu2021anomaly}. As shown
in the right of Figure~\ref{fig:imputation_anomaly_detection}, we present the number of detected anomalies with given quantiles, where Timer outperforms other advanced anomaly detection models, exhibiting the versatility of our generative time series model. Moreover, Figure~\ref{fig:anomaly_detection} compares the detection performance of pre-trained models and from scratch using two indicators. The left figure shows the number of datasets that the model has completed detection within the quantile of $3\%$ and $10\%$, and the right figure shows the quantile distribution and the averaged quantile of all $250$ UCR datasets, where the pre-trained Timer with the smaller averaged quantile works as a more accurate detector.

\subsection{Scalability}
Scalability is the essential property that emerges from pre-trained models to large models. To investigate the scaling behavior of Timer, we pre-train Timer with increased model size and data size as detailed in Appendix~\ref{sec:pretrain_detail} and evaluate it in downstream forecasting on all subsets of PEMS.

\begin{figure}[ht]
\vspace{-5pt}
\begin{center}
    \center{\includegraphics[width=0.48\textwidth]{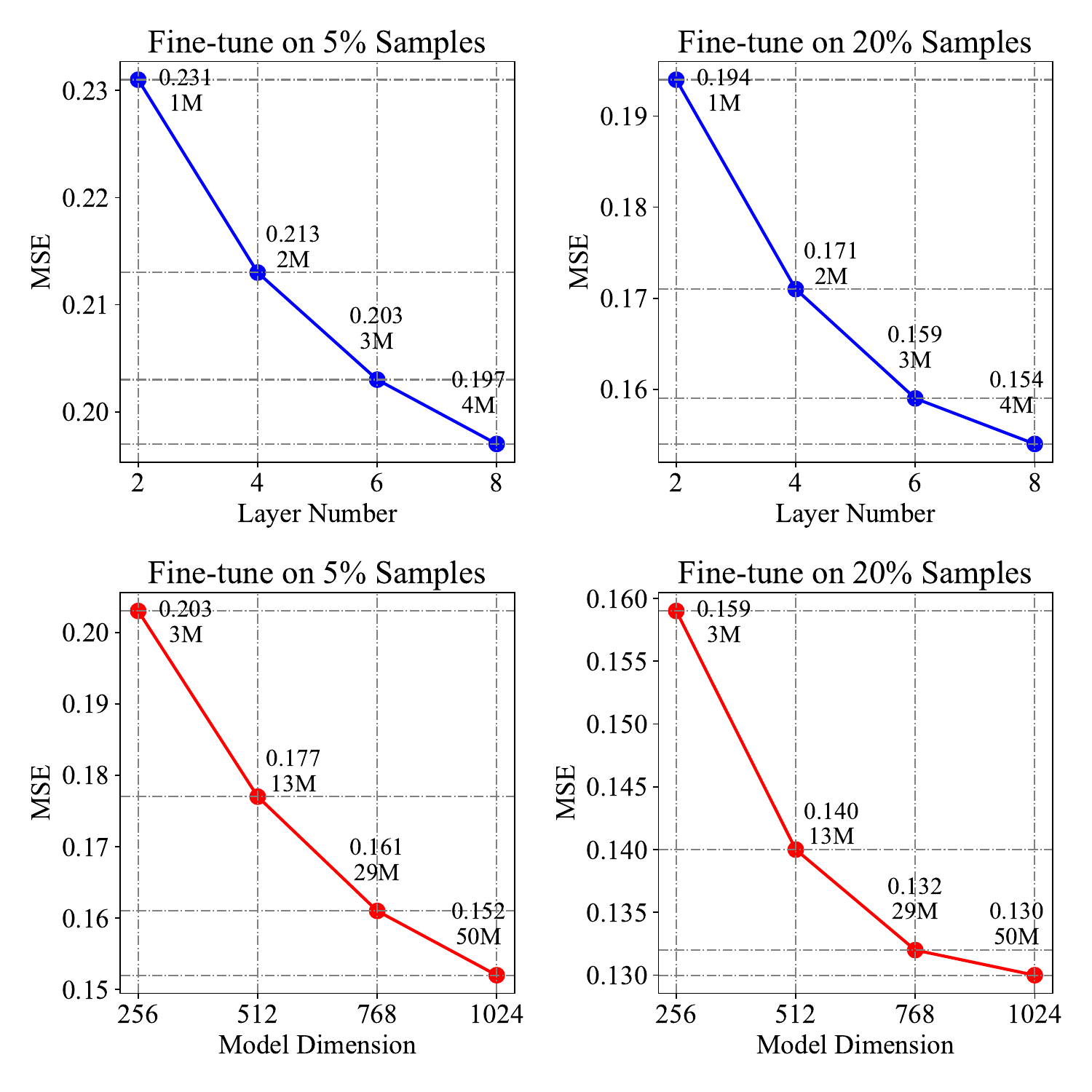}}
    \vspace{-18pt}
	\caption{Larger Timer demonstrates better performance on downstream forecasting. Models are all pre-trained on UTSD-4G. Detailed results of all PEMS subsets are provided in Table~\ref{tab:scale_up_full}.}
	\label{fig:scaleup_model}
\end{center}
\end{figure}

\paragraph{Model size} We keep UTSD-4G as the pre-training set. Results are presented in Figure~\ref{fig:scaleup_model}. While keeping model dimension $D=256$, we increase the number of layers. The growth of parameters from 1M to 4M leads to the decrease in forecasting errors in two few-shot scenarios by an average of $14.7\%$ and $20.6\%$ respectively. Subsequently, we increase the model dimension under the fixed layer number $L=6$, enlarging parameters from 3M to 50M, resulting in further improved performance of $25.1\%$ and $18.2\%$, validating the efficacy of scaling up the model size.

\begin{figure}[ht]
\begin{center}
    \centerline{\includegraphics[width=0.48\textwidth]{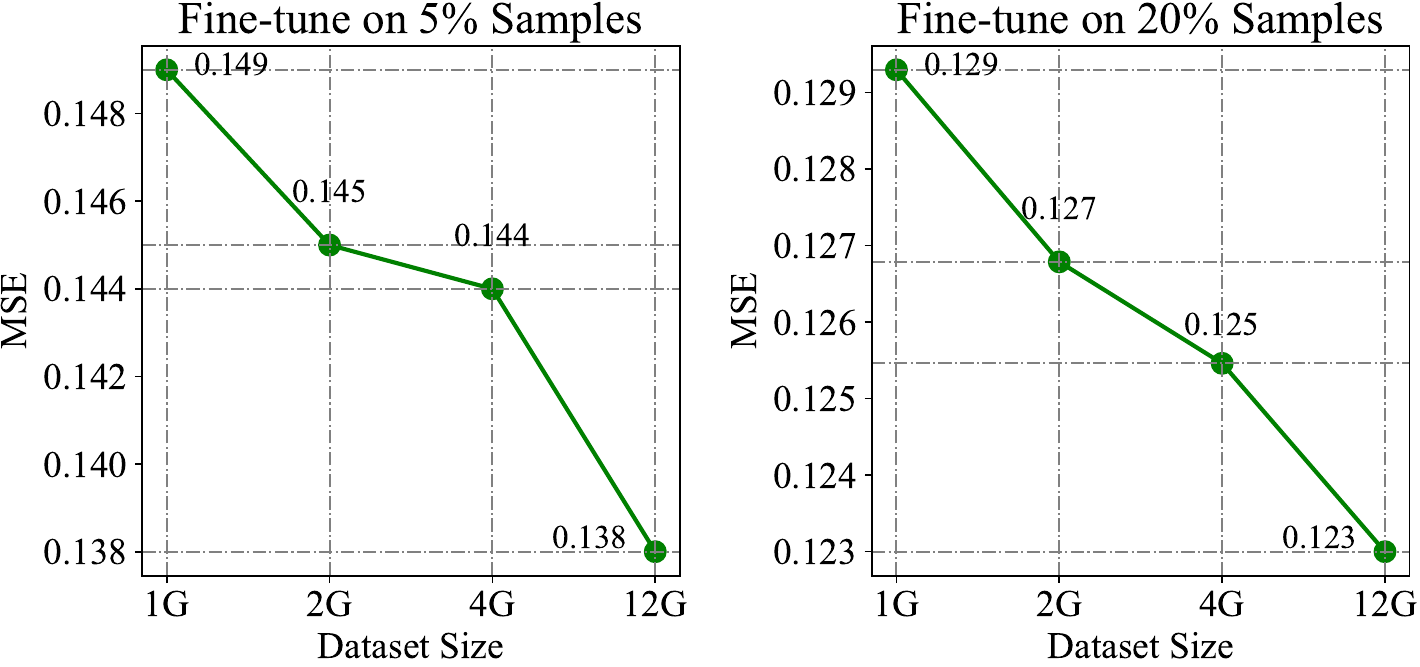}}
    \vspace{-7pt}
	\caption{Timer trained on larger dataset demonstrates better performance on downstream forecasting. Models are configured with $L=8$ and $D=1024$. Detailed results are provided in Table~\ref{tab:scale_up_full}.}
	\label{fig:scaleup_dataset}
\end{center}
\vspace{-28pt}
\end{figure}

\paragraph{Data scale} We pre-train Timer under the same model size with different UTSD sizes, which exhibits steady improvement with the enlarged pre-training scale in Figure~\ref{fig:scaleup_dataset}. The benefit is relatively small compared to expanding the model size previously, which can be attributed to the performance saturation on these datasets. Compared with large language models, the parameter scale of Timer can still be small, indicating the higher parameter efficiency in the time series modality, which is also supported by prior works~\cite{das2023decoder}. As the scaling law~\cite{kaplan2020scaling} of large models highlights the significance of synchronized scaling of data with the model parameters, there is still an urgent need to accelerate the data infrastructure in the time series field to promote the development of LTSMs.

Overall, by increasing the model size and data scale, Timer reduces the prediction error as $0.231\to0.138\ (-40.3\%)$ and $0.194\to0.123\ (-36.6\%)$ under few-shot scenarios, surpassing state-of-the-art multivariate forecaster~\cite{liu2023itransformer} training on full samples of PEMS datasets $(0.139)$.

\begin{table*}[ht]
  \vspace{-5pt}
  \caption{Downstream forecasting results under different data scarcity of the encoder-only and decoder-only Transformer respectively pre-trained on UTST-12G. Datasets are ordered by the degradation in Figure~\ref{fig:saturation}. Full results of PEMS and ETT can be found in Table~\ref{tab:forecast_full}.}
  \vskip 0.1in
  \label{tab:forecast}
  \footnotesize
  \begin{small}
  \begin{sc}
  \linespread{2}
  \renewcommand{\multirowsetup}{\centering}
  \setlength{\tabcolsep}{5.9pt}
  \begin{tabular}{c|cc|cc|cc|cc|cc|cc}
    \toprule
    Scenario & \multicolumn{4}{c|}{1\% Target} & \multicolumn{4}{c|}{5\% Target} & \multicolumn{4}{c}{20\% Target} \\
    \cmidrule(lr){1-1} \cmidrule(lr){2-5} \cmidrule(lr){6-9} \cmidrule(lr){10-13}
    Architecture & \multicolumn{2}{c|}{Encoder} & \multicolumn{2}{c|}{Decoder} & \multicolumn{2}{c|}{Encoder} & \multicolumn{2}{c|}{Decoder} & \multicolumn{2}{c|}{Encoder} & \multicolumn{2}{c}{Decoder}  \\ 
    \cmidrule(lr){1-1} \cmidrule(lr){2-3} \cmidrule(lr){4-5} \cmidrule(lr){6-7} \cmidrule(lr){8-9} \cmidrule(lr){10-11} \cmidrule(lr){12-13}
    Pre-trained & None & 12G & None & 12G & None & 12G & None & 12G & None & 12G & None & 12G  \\
    \toprule
    PEMS (Avg) & 0.286 & 0.246 & 0.328 & \textbf{0.180} & 0.220 & 0.197 & 0.215 & \textbf{0.138} & 0.173 & 0.164 & 0.153 & \textbf{0.126}\\
    \midrule
    ECL & 0.183 & 0.168 & 0.215 & \textbf{0.140} & 0.150 & 0.147 & 0.154 & \textbf{0.132} & 0.140 & 0.138 & 0.137 & \textbf{0.134}\\
    \midrule
    Traffic & 0.442 & 0.434 & 0.545 & \textbf{0.390} & 0.392 & 0.384 & 0.407 & \textbf{0.361} & 0.367 & 0.363 & 0.372 & \textbf{0.352}\\
    \midrule
    ETT (Avg) &  0.367 & 0.317 & 0.340 & \textbf{0.295} & 0.339 & 0.303 & 0.321 & \textbf{0.285} & 0.309 & 0.301 & 0.297 & \textbf{0.288}\\
    \midrule
    Weather & 0.224 & 0.165 & 0.246 & \textbf{0.166} & 0.182 & 0.154 & 0.198 & \textbf{0.151} & 0.153 & \textbf{0.149} & 0.166 & 0.151\\
    \bottomrule
  \end{tabular}
  \end{sc}
  \vspace{-10pt}
  \end{small}
\end{table*}

\subsection{Model Analysis}\label{sec:arch_compare}

\paragraph{Backbone for LTSM} Deep learning approaches have brought the boom of time series analysis, with various backbones for modeling the sequential time series modality being proposed. To validate the appropriate option for large time series models, we compare Timer with four candidates: MLP-based TiDE~\cite{das2023long}, CNN-based TCN~\cite{bai2018empirical}, RNN-based LSTM~\cite{hochreiter1997long} and encoder-only PatchTST~\cite{nie2022time}.

\begin{figure}[h]
\begin{center}
    \centerline{\includegraphics[width=0.48\textwidth]{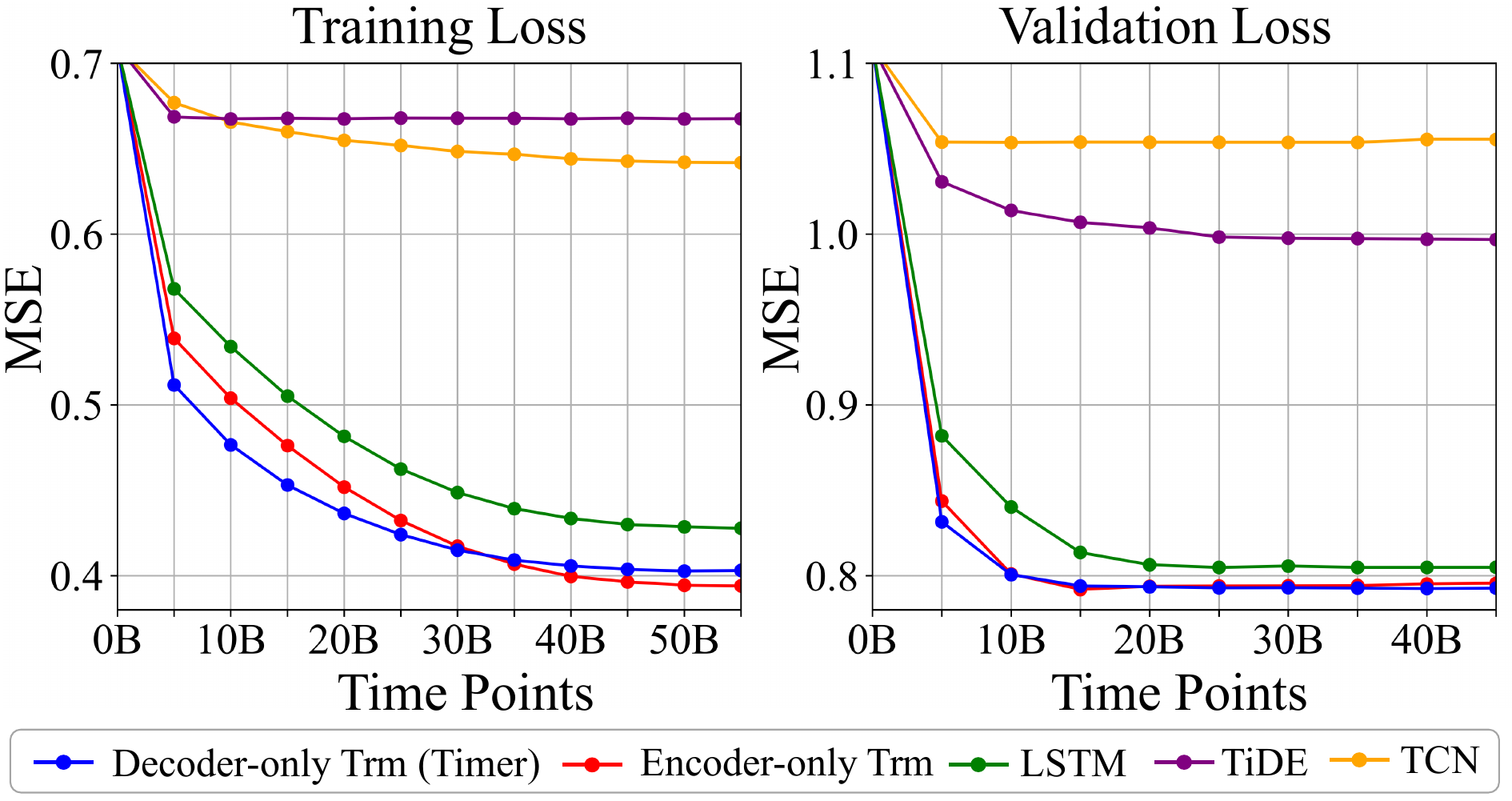}}
    \vspace{-5pt}
	\caption{Training loss of candidate backbones. Model dimension and layer number are consistently chosen for a fair comparison.}
	\label{fig:backbone}
\end{center}
\vspace{-17pt}
\end{figure}

To make sure the evaluation is comparable on different backbones, we maintain the same model configuration, including the model dimension and layer number, and pre-train these backbones on our UTSD-4G respectively. We set the token length $S=96$ and the context length as $672$ for Timer. For other non-autoregressive backbones, we pre-train them by direct multi-step forecasting in the $672$-pred-$96$ setting. The loss curves of training and validation are calculated as the MSE of the same set of model outputs (length-$96$ time series). As illustrated in Figure~\ref{fig:backbone}, Transformer exhibits excellent scalable ability as the backbone for LTSMs, whereas MLP-based and CNN-based architectures may encounter the bottleneck in accommodating diverse time series data.

\paragraph{Decoder-only v.s. Encoder-only} While a smaller training loss is achieved by the encoder-only Transformer in Figure~\ref{fig:backbone}, the progress of large language models indicates that decoder-only models may possess stronger generalization capabilities in downstream adaptation~\cite{wang2022language, dai2022can}, which is the essential purpose of LTSMs. Therefore, we proceed to compare their forecasting performance under varying degrees of data scarcity.

We elaborately evaluate two architectures on six benchmarks in Table~\ref{tab:forecast}. In the case of training from scratch (Pre-trained = None), the encoder-only Transformer will achieve better performance if the training samples are insufficient (Target = 1\%). Instead, the decoder-only architecture will demonstrate improved performance when more training samples are provided in the end-to-end scenarios. After pre-training on UTSD-12G (Pre-trained = 12G), Timer as the decoder-only Transformer achieves the best performance in most downstream scenarios, indicating better generalization than the encoder-only pre-trained model. The observations are consistent with several findings in large language models and elucidate why the encoder-only structure has become prevalent in the field of time series. Existing benchmarks can still be small and the encoder-only model can overfit in end-to-end scenarios. Meanwhile, the decoder-only Transformer, which excels at generalizing on different domains, is a promising choice for developing large time series models.

\begin{figure}[h]
\begin{center}
    \centerline{\includegraphics[width=0.48\textwidth]{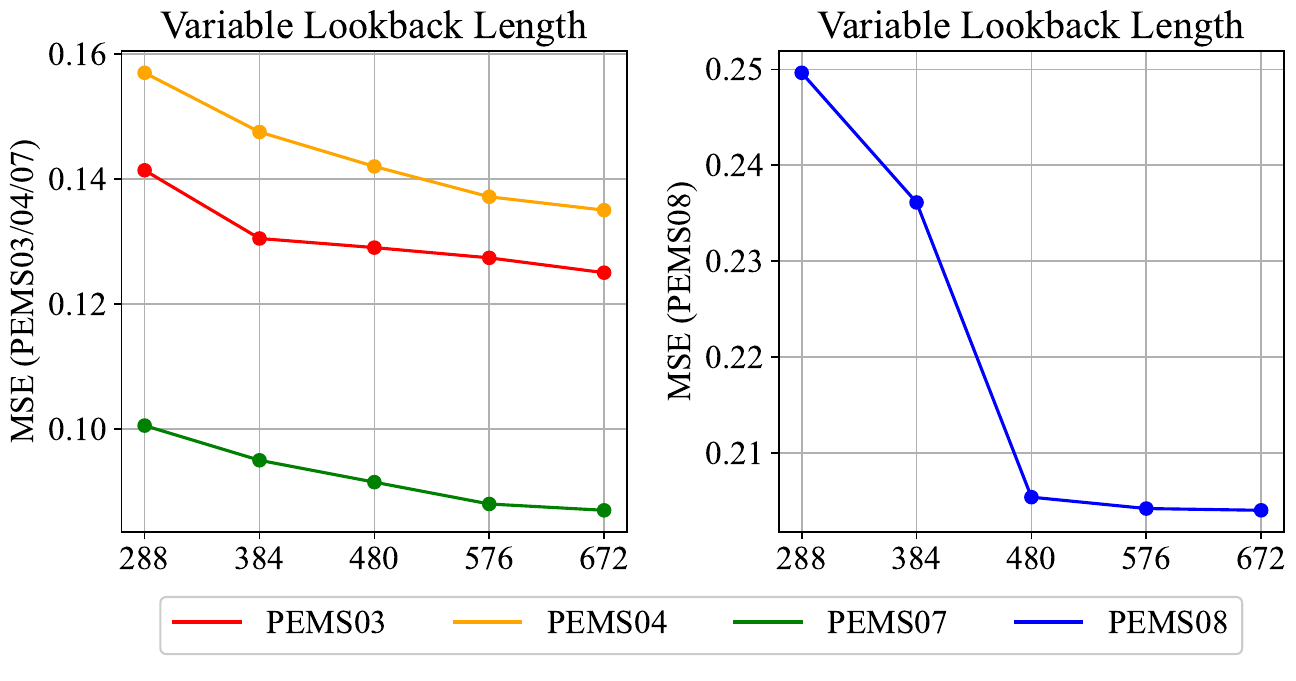}}
    \vspace{-12pt}
	\caption{Performance of one Timer for all lookback lengths.}
	\label{fig:vary_lookback}
    \end{center}
    \vspace{-27pt}
\end{figure}

\begin{figure*}[ht]
\begin{center}
    \center{\includegraphics[width=\textwidth]{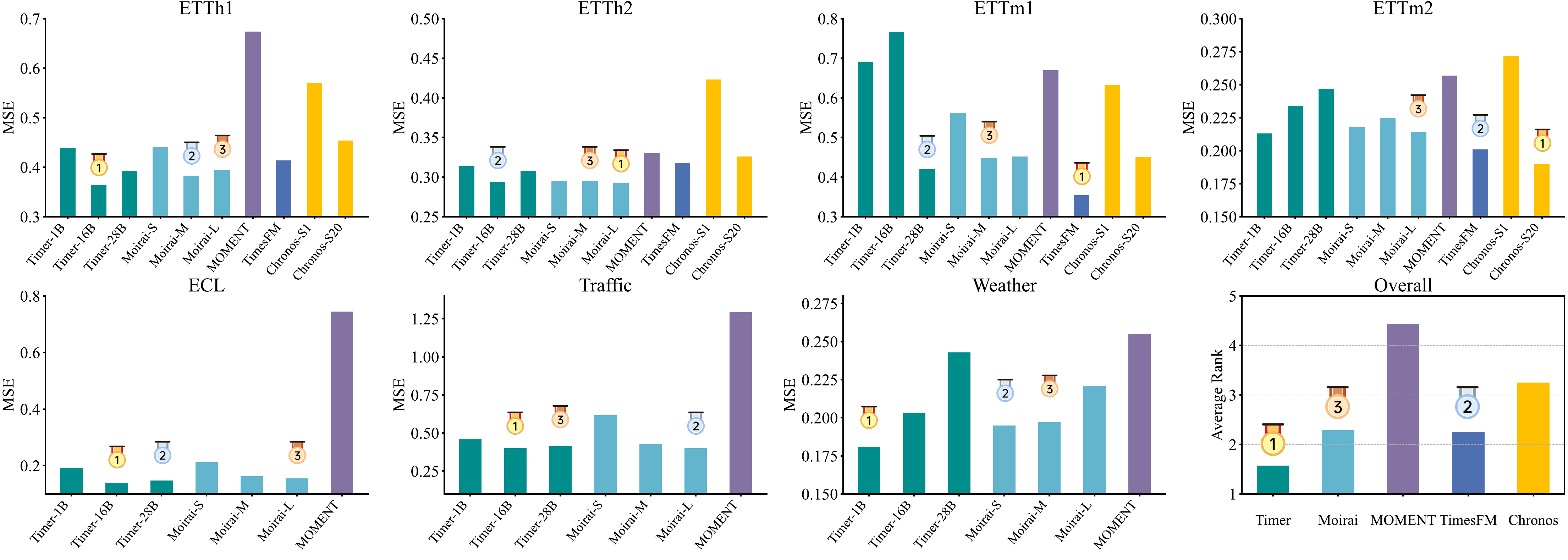}}
    \vspace{-25pt}
	\caption{Zero-shot evaluation on LTSMs. The top three models for each dataset are highlighted on the leaderboard. \emph{Average Rank} of each model is calculated on the benchmarks in which the model has participated. Detailed results are provided in Table~\ref{tab:forecast_baseline}.} 
	\label{fig:zeroshot}
\end{center}
\vspace{-12pt}
\end{figure*}

\begin{figure}[h]
\begin{center}
\vspace{-5pt}
    \centerline{\includegraphics[width=0.48\textwidth]{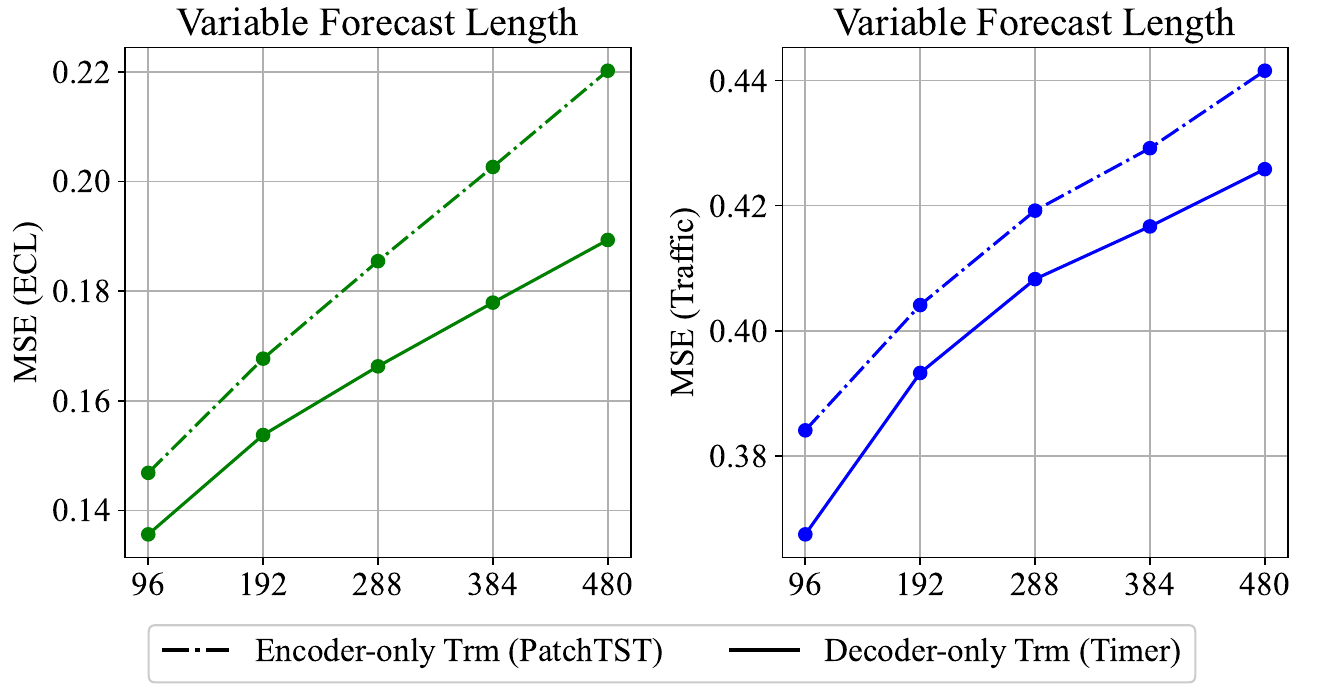}}
    \vspace{-10pt}
	\caption{Performance of Timer/PatchTST for all forecast lengths. We conduct rolling forecasting on a single $672$-pred-$96$ model.}
	\label{fig:vary_forecast}
  \end{center}
  \vspace{-24pt}
\end{figure}

\paragraph{Flexible sequence length} Typically, current deep forecasting models are trained on specific lookback and forecast lengths, limiting their versatility. Instead, the decoder-only architecture can provide the flexibility to address different sequence lengths. For instance, one Timer is applicable on different lookback lengths because of token-wise supervision outlined in Equation~\ref{equ:objective}. In addition to the feasibility, it achieves enhanced performance by increasing the lookback length in Figure~\ref{fig:vary_lookback}. As for the forecast length, increasing works~\cite{liu2024autotimes} bring the revival of autoregression (iterative multi-step prediction), enabling the generation of future predictions with arbitrary lengths. We explore this paradigm by rolling one model for all forecast lengths in Figure~\ref{fig:vary_forecast}, where the decoder-only Times exhibits smaller error accumulation, thereby achieving better performance.

\subsection{Evaluation of Large Time Series Models}
There is a growing surge in the development of large models in the field of time series~\cite{garza2023timegpt, das2023decoder, woo2024unified, ansari2024chronos, goswami2024moment}. One particularly fascinating direction of research is focused on zero-shot forecasting (ZSF), which has the potential to renovate the conventional practice of training small models or fine-tuning language models for each specific scenario. Zero-shot generalization represents a sophisticated capability of large models, necessitating substantial model capacity and pre-training on extensive datasets. Consequently, we are actively expanding our dataset by incorporating the latest data infrastructure~\cite{woo2024unified} in this field to pre-train Timer on ever larger scales (1B/16B/28B). Given the significant value to researchers and practitioners, we extensively evaluate concurrent large models and establish the first zero-shot forecasting benchmark of LTSMs as detailed in Appendix~\ref{sec:zero-shot}.

\paragraph{Quality assessments} Our evaluation assesses the quality of LTSMs in Table~\ref{tab:ltsm_comp}, including (1) fundamental attributes such as pre-training scale, parameters; (2) abilities such as applicable tasks, context length, etc. Current LTSMs essentially build upon Transformer, with a significantly smaller number of parameters compared to LLMs. There is still potential to support more tasks and longer contexts.
\vspace{-5pt}

\paragraph{Quantitative evaluations} We apply official checkpoints on seven datasets that do not appear during pre-training. The performance is fairly evaluated using MSE by predicting future $96$ points of all windows in each dataset. Figure~\ref{fig:zeroshot} presents the result and rank of each model, where the top-ranked LTSMs are Timer, Moirai~\cite{woo2024unified}, and TimesFM~\cite{das2023decoder}. However, the positive correlation between performance and pre-training scale remains relatively weak, highlighting the significance of high-quality data and synchronized scaling
of data and model size.
\vspace{-2pt}

\section{Conclusion and Future Work}\label{sec:conclusion}
Real-world time series analysis is increasingly underscoring the demand for large time series models (LTSM). In this paper, we release a time series dataset with 1 billion time points, propose a unified sequence format to address the heterogeneity of multivariate time series, and develop a generative pre-trained Transformer as a generalizable, scalable, task-general LTSM. Empirically, we evaluate our model in forecasting, imputation, and anomaly detection, yielding state-of-the-art performance and notable pre-training benefits in the data-scarce scenario. Further analysis validates the model scalability, explores the architecture for LTSMs, and highlights the versatility of our autoregressive generation. By performing zero-shot forecasting on available large models, we conduct the initial quantitative assessments among LTSMs. Quality evaluations unveil crucial pathways for future development, including better zero-shot generalization and facilitating probabilistic and long-context forecasting.

\section*{Impact Statement}
This paper aims to advance the development of large models for time series. In this work, we release a high-quality and unified time series dataset for scalable pre-training, which can serve as a foundation for pre-training and establishing new benchmarks. The outcome large model demonstrates notable effectiveness of generalization, versatility across various tasks, and scalability to refine performance, offering valuable insights for future investigations and application values for practitioners. Our paper mainly focuses on scientific research and has no obvious negative social impact.

\section*{Acknowledgements}
This work was supported by the National Natural Science Foundation of China (62022050 and U2342217), the BNRist Innovation Fund (BNR2024RC01010), and the National Engineering Research Center for Big Data Software.

\bibliography{example_paper}
\bibliographystyle{icml2024}


\newpage
\appendix
\onecolumn
\section{Unified Time Series Dataset}\label{sec:datasets_detail}
\subsection{Datasets Details}
Unified Time Series Dataset (UTSD) is meticulously assembled from a blend of publicly accessible online data repositories and empirical data derived from real-world machine operations. To enhance data integrity, missing values are systematically addressed using linear interpolation techniques. We follow the unified data storage format (parquet) used in ~\cite{woo2024unified}. For each univariate, multivariate, or irregular-sampled time series, we store them with timestamps and other meta-information in one directory using ARROW format. One dataset may composed of multiple related time series. We continue to expand the UTSD to include data from public datasets such as LOSTA\footnote{\href{https://huggingface.co/datasets/Salesforce/lotsa\_data}{https://huggingface.co/datasets/Salesforce/lotsa\_data}} for zero-shot forecasting. UTSD encompasses 29 individual datasets as listed with the asterisk mark in Table~\ref{tab:dataset_detailed_descriptions}, intricately representative of a wide range of domains.

All datasets can be classified into ten distinct domains by their source: Energy, Environment, Health, Internet of Things (IoT), Nature, Transport, Web, CloudOps, Finance, and Multiple Sources (Misc.), where the first seven domains originally come from our curated UTSD. The datasets exhibit diverse sampling frequencies, ranging from macro intervals such as yearly and quarterly to more fine-grained intervals like hourly and minutely. Notably, several datasets can demonstrate exceptionally high-frequency sampling rates, such as the MotorImagery dataset, which operates at a millisecond frequency.

In the pursuit of advanced data analysis, we have also analyzed the stationarity manifested as ADF test statistics~\cite{adftest} and forecastability~\cite{goerg2013forecastable}. The rigorous methodologies and intricate details are elaborated in Section~\ref{sec:dataset_statistics}. We utilize these statistical indicators to filter four high-quality subsets of UTSD, namely UTSD-1G, UTSD-2G, UTSD-4G, and UTSD-12G. As we expand the dataset, we continuously analyze statistical indicators and employ various methodologies to ensure the selection of high-quality datasets. LOTSA has not been sorted in this hierarchy due to its immensity.

\subsection{Statistics}\label{sec:dataset_statistics}
We analyze each dataset within our collection, examining the time series through the lenses of stationarity and forecastability. This approach allows us to characterize the level of complexity inherent to each dataset.

\paragraph{Stationarity} The stationarity of time series is a fundamental property that can be rigorously quantified using the Augmented Dickey-Fuller (ADF) test. Notably, a larger ADF test statistic typically signifies a higher degree of non-stationarity within the time series~\cite{adftest}. In the context of datasets comprising multiple time series, the challenge of aligning these series arises, particularly when they vary in length. To address this, we implement a length-weighted ADF method that evaluates the stationarity of the entire dataset, taking into consideration the varying lengths of individual series. This approach ensures that the contribution of each series to the overall stationarity metric is proportional to its length, thus reflecting its relative significance within the dataset. By doing so, the length-weighted ADF provides a more granular and accurate depiction of the stationarity of the dataset, highlighting the impact of longer series on the overall stability and ensuring that shorter series do not disproportionately affect the assessment. The weighted statistic is formulated as follows: 
\begin{equation}
T = \sum_{i=1}^C T_i, \
\operatorname{ADF-Statistic}(\mathcal{D}) = \sum_{i=1}^C \frac{T_i}{T}\operatorname{ADF-Statistic}(\mathbf{S}^{(i)}),
\end{equation}
where $\mathbf{S}_i \in \mathbb{R}^{T_i}$ denotes the $i$-th series in dataset $\mathcal{D}$, $T_i$ is the length of $\mathbf{S}_i$ and $C$ is the number of time series of dataset $\mathcal{D}$.

\paragraph{Forecastability}
Forecastability is calculated by subtracting the entropy of the series Fourier decomposition adopted from~\citet{goerg2013forecastable}, where a higher forecastability value indicates superior predictability. Just as with the assessment of stationarity, when considering a dataset composed of multiple time series of varying lengths, it is essential to adjust the measure of forecastability to account for these differences. Therefore, we extend the concept of forecastability to a weighted version, analogous to the length-weighted ADF method, to finely tune the predictability assessment to the characteristics of each series. The weighted forecastability for a dataset can be formulated as follows:
\begin{equation}
T = \sum_{i=1}^C T_i, \    
\operatorname{Forecastability}(\mathcal{D}) = \sum_{i=1}^C \frac{T_i}{T}(1 - \operatorname{Entropy}(\mathcal{F}(\mathbf{S}^{(i)}))),
\end{equation}
where $\mathbf{S}_i \in \mathbb{R}^{T_i}$ denotes the $i$-th time series in dataset $\mathcal{D}$, $T_i$ is the length of $\mathbf{S}_i$ and $C$ is the number of time series in dataset $\mathcal{D}$. $\mathcal{F}(\mathbf{S}^{(i)})$ denotes the Fourier decomposition of series $\mathbf{S}^{(i)}$.

\renewcommand\arraystretch{1.4}
\setlength{\tabcolsep}{0.8pt}
\setlength{\LTcapwidth}{\textwidth}
\begin{ThreePartTable}
\begin{TableNotes}
\footnotesize
\item [*] The asterisk marks the dataset that originally belongs to UTSD.
\end{TableNotes}
\begin{small}
\begin{sc}
\begin{longtable}{c|c|c|c|c|c|c|c}
    \caption{Dataset detailed descriptions. \emph{Time Points} denotes the total number of time points aggregating from all variates if multivariate. \emph{File Size} denotes the storage that the ARROW format of the dataset occupies. \emph{Freq.} denotes the sampling interval of time points, where ``-'' indicates no timestamp or irregular interval. \emph{ADF.} denotes the Augmented Dickey-Fuller test statistics of the dataset. \emph{Forecast.} denotes the forecastability of the dataset. \emph{Source} denotes the original paper or resource of the dataset.}\label{tab:dataset_detailed_descriptions} \\
    \toprule
    Domain & Dataset & Time Points & File Size & Freq. & ADF. & Forecast. & Source \\
    \endfirsthead
    \multicolumn{8}{c}%
    {{\bfseries Table \thetable\ continued from previous page}} \\
    \toprule
    Domain & Dataset & Time Points & File Size & Freq. & ADF. & Forecast. & Source \\
    \toprule
    \endhead
    \bottomrule
    \endfoot
    \bottomrule
    \insertTableNotes
    \endlastfoot
  \toprule
  \multirow{23}{*}{Energy} 
  & \scalebox{0.90}[0.9]{London Smart Meters\tnote{*}} & 166.50M & 636M & Hourly & -13.158 & 0.173 & \scalebox{0.8}[0.8]{\citet{godahewa2021monash}}\\
  \cmidrule(lr){2-8}
  & \scalebox{0.90}[0.9]{Wind Farms\tnote{*}} & 7.40M & 29M & 4 sec & -29.174 & 0.811 &\scalebox{0.8}[0.8]{\citet{godahewa2021monash}} \\
  \cmidrule(lr){2-8}
  & \scalebox{0.90}[0.9]{Aus. Electricity Demand\tnote{*}} & 1.16M & 5M & 30 min & -27.554 & 0.730 &\scalebox{0.8}[0.8]{\citet{godahewa2021monash}} \\
  \cmidrule(lr){2-8}
  & \scalebox{0.90}[0.9]{BDG-2 Panther} & 0.92M & 4M & Hourly & -6.593 & 0.479 &\scalebox{0.8}[0.8]{\citet{emami2023buildingsbench}} \\
  \cmidrule(lr){2-8}
  & \scalebox{0.90}[0.9]{BDG-2 Fox} & 2.32M & 9M & Hourly & -9.191 & 0.469 &\scalebox{0.8}[0.8]{\citet{emami2023buildingsbench}} \\
  \cmidrule(lr){2-8}
  & \scalebox{0.90}[0.9]{BDG-2 Rat} & 4.73M & 19M & Hourly & -6.868 & 0.456 &\scalebox{0.8}[0.8]{\citet{emami2023buildingsbench}} \\
  \cmidrule(lr){2-8}
  & \scalebox{0.90}[0.9]{BDG-2 Bear} & 1.48M & 6M & Hourly & -11.742 & 0.471 &\scalebox{0.8}[0.8]{\citet{emami2023buildingsbench}} \\
  \cmidrule(lr){2-8}
  & \scalebox{0.90}[0.9]{Low Carbon London} & 9.54M & 37M & Hourly & -12.366 & 0.134 &\scalebox{0.8}[0.8]{\citet{emami2023buildingsbench}} \\
  \cmidrule(lr){2-8}
  & \scalebox{0.90}[0.9]{SMART} & 0.10M & 1M & Hourly & -10.755 & 0.143 &\scalebox{0.8}[0.8]{\citet{emami2023buildingsbench}} \\
  \cmidrule(lr){2-8}
  & \scalebox{0.90}[0.9]{IDEAL} & 1.26M & 5M & Hourly & -11.223 & 0.106 &\scalebox{0.8}[0.8]{\citet{emami2023buildingsbench}} \\
  \cmidrule(lr){2-8}
  & \scalebox{0.90}[0.9]{Sceaux} & 0.03M & 1M & Hourly & -14.172 & 0.143 &\scalebox{0.8}[0.8]{\citet{emami2023buildingsbench}} \\
  \cmidrule(lr){2-8}
  & \scalebox{0.90}[0.9]{Borealis} & 0.08M & 1M & Hourly & -6.612 & 0.160 &\scalebox{0.8}[0.8]{\citet{emami2023buildingsbench}} \\
  \cmidrule(lr){2-8}
  & \scalebox{0.90}[0.9]{Buildings900K } & 15852.22M & 60102M & Hourly & -8.412 & 0.357 &\scalebox{0.8}[0.8]{\citet{emami2023buildingsbench}} \\
  \cmidrule(lr){2-8}
  & \scalebox{0.90}[0.9]{Covid19 Energy} & 0.03M & 1M & Hourly & -13.768 & 0.698 &\scalebox{0.8}[0.8]{\citet{wang2023benchmarks}} \\
  \cmidrule(lr){2-8}
  & \scalebox{0.90}[0.9]{GEF12} & 1.58M & 6M & Hourly & -9.576 & 0.566 &\scalebox{0.8}[0.8]{\citet{wang2023benchmarks}} \\
  \cmidrule(lr){2-8}
  & \scalebox{0.90}[0.9]{GEF14} & 0.02M & 1M & Hourly & -9.372 & 0.628 &\scalebox{0.8}[0.8]{\citet{wang2023benchmarks}} \\
  \cmidrule(lr){2-8}
  & \scalebox{0.90}[0.9]{GEF17} & 0.28M & 1M & Hourly & -5.976 & 0.599 &\scalebox{0.8}[0.8]{\citet{wang2023benchmarks}} \\
  \cmidrule(lr){2-8}
  & \scalebox{0.90}[0.9]{PDB} & 0.04M & 1M & Hourly & -6.453 & 0.622 &\scalebox{0.8}[0.8]{\citet{wang2023benchmarks}} \\
  \cmidrule(lr){2-8}
  & \scalebox{0.90}[0.9]{Spanish} & 0.07M & 1M & Hourly & -13.217 & 0.770 &\scalebox{0.8}[0.8]{\citet{wang2023benchmarks}} \\
  \cmidrule(lr){2-8}
  & \scalebox{0.90}[0.9]{ELF} & 0.02M & 1M & Hourly & -13.607 & 0.770 &\scalebox{0.8}[0.8]{\citet{wang2023benchmarks}} \\
  \cmidrule(lr){2-8}
  & \scalebox{0.90}[0.9]{KDD Cup 2022} & 4.73M & 181M & Hourly & -17.017 & 0.225 &\scalebox{0.8}[0.8]{\citet{zhou2022sdwpf}} \\
  \cmidrule(lr){2-8}
  & \scalebox{0.90}[0.9]{Residential Load Power} & 437.98M & 1671M & Minutely & -37.979 & 0.264 &\scalebox{0.8}[0.8]{\citet{Bergmeir_C_2023}} \\
  \cmidrule(lr){2-8}
  & \scalebox{0.90}[0.9]{Residential PV Power} & 373.37M & 1435M & Minutely & -31.389 & 0.421 &\scalebox{0.8}[0.8]{\citet{Bergmeir_C_2023}} \\
  \midrule
  \multirow{5}{*}{Environment} 
  & \scalebox{0.90}[0.9]{AustraliaRainfall\tnote{*}} & 11.54M & 45M & Hourly & -150.10 & 0.458 &\scalebox{0.8}[0.8]{\citet{tan2021time}} \\
  \cmidrule(lr){2-8}
  & \scalebox{0.90}[0.9]{BeijingPM25Quality\tnote{*}} & 3.66M & 14M & Hourly & -31.415 & 0.404 & \scalebox{0.8}[0.8]{\citet{tan2021time}}\\
  \cmidrule(lr){2-8}
  & \scalebox{0.90}[0.9]{BenzeneConcentration\tnote{*}} & 16.34M & 63M & Hourly & -65.187 & 0.526 &\scalebox{0.8}[0.8]{\citet{tan2021time}} \\
  \cmidrule(lr){2-8}
  & \scalebox{0.90}[0.9]{China Air Quality\tnote{*}} & 34.29M & 132M & Hourly & -12.602 & 0.529 &\scalebox{0.8}[0.8]{\citet{zheng2015forecasting}} \\
  \cmidrule(lr){2-8}
  & \scalebox{0.90}[0.9]{Beijing Air Quality\tnote{*}} & 4.62M & 18M & Hourly & -15.758 & 0.332 &\scalebox{0.8}[0.8]{\citet{misc_beijing_multi-site_air_quality_501}} \\
  \pagebreak
  \multirow{12}{*}{Health}
  & \scalebox{0.90}[0.9]{MotorImagery\tnote{*}} & 72.58M & 279M & 0.001 sec & -3.132 & 0.449 &\scalebox{0.8}[0.8]{\citet{dau2019ucr}} \\
  \cmidrule(lr){2-8}
  & \scalebox{0.90}[0.9]{SelfRegulationSCP1\tnote{*}} & 3.02M & 12M & 0.004 sec & -3.191 & 0.504 &\scalebox{0.8}[0.8]{\citet{dau2019ucr}} \\
  \cmidrule(lr){2-8}
  & \scalebox{0.90}[0.9]{SelfRegulationSCP2\tnote{*}} & 3.06M & 12M & 0.004 sec & -2.715 & 0.481 &\scalebox{0.8}[0.8]{\citet{dau2019ucr}} \\
  \cmidrule(lr){2-8}
  & \scalebox{0.90}[0.9]{AtrialFibrillation\tnote{*}} & 0.04M & 1M & 0.008 sec & -7.061 & 0.167 &\scalebox{0.8}[0.8]{\citet{dau2019ucr}} \\
  \cmidrule(lr){2-8}
  & \scalebox{0.90}[0.9]{PigArtPressure\tnote{*}} & 0.62M & 3M & - & -7.649 & 0.739 &\scalebox{0.8}[0.8]{\citet{dau2019ucr}} \\
  \cmidrule(lr){2-8}
  & \scalebox{0.90}[0.9]{PigCVP\tnote{*}} & 0.62M & 3M & - & -4.855 & 0.577 &\scalebox{0.8}[0.8]{\citet{dau2019ucr}} \\
  \cmidrule(lr){2-8}
  & \scalebox{0.90}[0.9]{IEEEPPG\tnote{*}} & 15.48M & 61M & 0.008 sec & -7.725 & 0.380 &\scalebox{0.8}[0.8]{\citet{tan2021time}} \\
  \cmidrule(lr){2-8}
  & \scalebox{0.90}[0.9]{BIDMC32HR\tnote{*}} & 63.59M & 244M & - & -14.135 & 0.523 &\scalebox{0.8}[0.8]{\citet{tan2021time}} \\
  \cmidrule(lr){2-8}
  & \scalebox{0.90}[0.9]{TDBrain\tnote{*}} & 72.30M & 283M & 0.002 sec & -3.167 & 0.967 &\scalebox{0.8}[0.8]{\citet{wang2023contrast}} \\
  \cmidrule(lr){2-8}
  & \scalebox{0.90}[0.9]{CDC Fluview ILINet } & 0.28M & 2M & Weekly & -4.381 & 0.307 &\scalebox{0.8}[0.8]{\citet{cdc_2017}} \\
  \cmidrule(lr){2-8}
  & \scalebox{0.90}[0.9]{CDC Fluview WHO NREVSS} & 0.14M & 1M & Weekly & -7.928 & 0.233 &\scalebox{0.8}[0.8]{\citet{cdc_2017}} \\
  \cmidrule(lr){2-8}
  & \scalebox{0.90}[0.9]{Project Tycho} & 1.35M & 5M & Weekly & -8.167 & 0.111 &\scalebox{0.8}[0.8]{\citet{van2018project}} \\
  \midrule
  IoT & \scalebox{0.90}[0.9]{SensorData\tnote{*}} & 165.4M & 631M & 0.02 sec & -15.892 & 0.917 & \scalebox{0.8}[0.8]{Real-world machine logs} \\
  \midrule
  \multirow{13}{*}{Nature}
  & \scalebox{0.90}[0.9]{Phoneme\tnote{*}} & 2.16M & 9M & - & -8.506 & 0.243 &\scalebox{0.8}[0.8]{\citet{dau2019ucr}} \\
  \cmidrule(lr){2-8}
  & \scalebox{0.90}[0.9]{EigenWorms\tnote{*}} & 27.95M & 107M & - & -12.201 & 0.393 &\scalebox{0.8}[0.8]{\citet{dau2019ucr}} \\
  \cmidrule(lr){2-8}
  & \scalebox{0.90}[0.9]{ERA5\tnote{*}} & 96458.81M & 368610M & Hourly & -7.970 & 0.581 &\scalebox{0.8}[0.8]{\citet{nguyen2024climatelearn}} \\
  \cmidrule(lr){2-8}
  & \scalebox{0.90}[0.9]{CMIP6} & 104593.00M & 399069M & 6 H & -7.960 & 0.573 &\scalebox{0.8}[0.8]{\citet{nguyen2024climatelearn}} \\
  \cmidrule(lr){2-8}
  & \scalebox{0.90}[0.9]{Temperature Rain\tnote{*}} & 23.25M & 93M & Daily & -10.952 & 0.133 &\scalebox{0.8}[0.8]{\citet{godahewa2021monash}} \\
  \cmidrule(lr){2-8}
  & \scalebox{0.90}[0.9]{StarLightCurves\tnote{*}} & 9.46M & 37M & - & -1.891 & 0.555 &\scalebox{0.8}[0.8]{\citet{dau2019ucr}} \\
  \cmidrule(lr){2-8}
  & \scalebox{0.90}[0.9]{Saugeen River Flow\tnote{*}} & 0.02M & 1M & Daily & -19.305 & 0.300 &\scalebox{0.8}[0.8]{\citet{godahewa2021monash}} \\
  \cmidrule(lr){2-8}
  & \scalebox{0.90}[0.9]{KDD Cup 2018\tnote{*}} & 2.94M & 12M & Hourly & -10.107 & 0.362 &\scalebox{0.8}[0.8]{\citet{godahewa2021monash}} \\
  \cmidrule(lr){2-8}
  & \scalebox{0.90}[0.9]{US Births\tnote{*}} & 0.00M & 1M & Daily & -3.352 & 0.675 &\scalebox{0.8}[0.8]{\citet{godahewa2021monash}} \\
  \cmidrule(lr){2-8}
  & \scalebox{0.90}[0.9]{Sunspot\tnote{*}} & 0.07M & 1M & Daily & -7.866 & 0.287 &\scalebox{0.8}[0.8]{\citet{godahewa2021monash}} \\
  \cmidrule(lr){2-8}
  & \scalebox{0.90}[0.9]{Worms} & 0.23M & 1M & 0.033 sec & -3.851 & 0.395 &\scalebox{0.8}[0.8]{\citet{dau2019ucr}} \\
  \cmidrule(lr){2-8}
  & \scalebox{0.90}[0.9]{Subseasonal} & 56.79M & 217M & Daily & -12.391 & 0.414 &\scalebox{0.8}[0.8]{\citet{mouatadid2024subseasonalclimateusa}} \\
  \cmidrule(lr){2-8}
  & \scalebox{0.90}[0.9]{Subseasonal Precipitation} & 9.76M & 38M & Daily & -13.567 & 0.276 &\scalebox{0.8}[0.8]{\citet{mouatadid2024subseasonalclimateusa}} \\
  \midrule
  \multirow{7}{*}{Transport}
  & \scalebox{0.90}[0.9]{Pedestrian Counts\tnote{*}} & 3.13M & 12M & Hourly & -23.462 & 0.297 &\scalebox{0.8}[0.8]{\citet{godahewa2021monash}} \\
  \cmidrule(lr){2-8}
  & \scalebox{0.90}[0.9]{PEMS 03} & 9.38M & 36M & 5 min & -19.051 & 0.411 &\scalebox{0.8}[0.8]{\citet{libcitylong}} \\
  \cmidrule(lr){2-8}
  & \scalebox{0.90}[0.9]{PEMS 04} & 15.65M & 60M & 5 min & -15.192 & 0.494 &\scalebox{0.8}[0.8]{\citet{libcitylong}} \\
  \cmidrule(lr){2-8}
  & \scalebox{0.90}[0.9]{PEMS 07} & 24.92M & 96M & 5 min & -20.603 & 0.466 &\scalebox{0.8}[0.8]{\citet{libcitylong}} \\
  \cmidrule(lr){2-8}
  & \scalebox{0.90}[0.9]{PEMS 08} & 9.11M & 35M & 5 min & -14.918 & 0.551 &\scalebox{0.8}[0.8]{\citet{libcitylong}} \\
  \cmidrule(lr){2-8}
  & \scalebox{0.90}[0.9]{PEMS Bay} & 16.94M & 65M & 5 min & -12.770 & 0.704 &\scalebox{0.8}[0.8]{\citet{libcitylong}} \\
  \cmidrule(lr){2-8}
  & \scalebox{0.90}[0.9]{Los-Loop} & 7.09M & 28M & 5 min & -16.014 & 0.657 &\scalebox{0.8}[0.8]{\citet{libcitylong}} \\
  \pagebreak
  \multirow{10}{*}{Transport}
  & \scalebox{0.90}[0.9]{Loop Seattle} & 33.95M & 130M & 5 min & -32.209 & 0.535 &\scalebox{0.8}[0.8]{\citet{libcitylong}} \\
  \cmidrule(lr){2-8}
  & \scalebox{0.90}[0.9]{SZ-Taxi} & 0.46M & 2M & 15 min & -5.900 & 0.217 &\scalebox{0.8}[0.8]{\citet{libcitylong}} \\
  \cmidrule(lr){2-8}
  & \scalebox{0.90}[0.9]{Beijing Subway} & 0.87M & 22M & 30 min & -8.571 & 0.219 &\scalebox{0.8}[0.8]{\citet{libcitylong}} \\
  \cmidrule(lr){2-8}
  & \scalebox{0.90}[0.9]{SHMetroy} & 5.07M & 20M & 15 min & -17.014 & 0.222 &\scalebox{0.8}[0.8]{\citet{libcitylong}} \\
  \cmidrule(lr){2-8}
  & \scalebox{0.90}[0.9]{HZMetro} & 0.38M & 2M & 15 min & -11.254 & 0.232 &\scalebox{0.8}[0.8]{\citet{libcitylong}} \\
  \cmidrule(lr){2-8}
  & \scalebox{0.90}[0.9]{Q-Traffic} & 264.39M & 1011M & 15 min & -15.761& 0.490 &\scalebox{0.8}[0.8]{\citet{libcitylong}} \\
  \cmidrule(lr){2-8}
  & \scalebox{0.90}[0.9]{Taxi} & 55.00M & 212M & 30 min & -8.302 & 0.146 &\scalebox{0.8}[0.8]{\citet{alexandrov2020gluonts}} \\
  \cmidrule(lr){2-8}
  & \scalebox{0.90}[0.9]{Uber TLC Daily} & 0.05M & 1M & Daily & -1.778 & 0.285 &\scalebox{0.8}[0.8]{\citet{alexandrov2020gluonts}} \\
  \cmidrule(lr){2-8}
  & \scalebox{0.90}[0.9]{Uber TLC Hourly} & 1.13M & 5M & Hourly & -9.022 & 0.124 &\scalebox{0.8}[0.8]{\citet{alexandrov2020gluonts}} \\
  \cmidrule(lr){2-8}
  & \scalebox{0.90}[0.9]{LargeST} & 4452.20M & 16988M & 5 min & -38.020 & 0.436 &\scalebox{0.8}[0.8]{\citet{liu2023largest}} \\
  \midrule
  \multirow{2}{*}{Web}
  & \scalebox{0.90}[0.9]{Web Traffic\tnote{*}} & 116.49M & 462M & Daily & -8.272 & 0.299 &\scalebox{0.8}[0.8]{\citet{godahewa2021monash}} \\
  \cmidrule(lr){2-8}
  & \scalebox{0.90}[0.9]{Wiki-Rolling} & 40.62M & 157M & Daily & -5.524 & 0.242 &\scalebox{0.8}[0.8]{\citet{alexandrov2020gluonts}} \\
  \midrule
  \multirow{3}{*}{CloudOps}
  & \scalebox{0.90}[0.9]{Alibaba Cluster Trace 2018} & 190.39M & 2909M & 5 min & -5.303 & 0.668 &\scalebox{0.8}[0.8]{\citet{woo2023pushing}} \\
  \cmidrule(lr){2-8}
  & \scalebox{0.90}[0.9]{Azure VM Traces 2017} & 885.52M & 10140M & 5 min & -11.482 & 0.290 &\scalebox{0.8}[0.8]{\citet{woo2023pushing}} \\
  \cmidrule(lr){2-8}
  & \scalebox{0.90}[0.9]{Borg Cluster Data 2011} & 1073.89M & 14362M & 5 min & -8.975 & 0.505 &\scalebox{0.8}[0.8]{\citet{woo2023pushing}} \\
  \midrule
  \multirow{5}{*}{Sales}
  & \scalebox{0.90}[0.9]{M5} & 58.33M & 224M & Daily & -6.985 & 0.247 &\scalebox{0.8}[0.8]{\citet{alexandrov2020gluonts}} \\
  \cmidrule(lr){2-8}
  & \scalebox{0.90}[0.9]{Favorita Sales} & 139.18M & 535M & Daily & -6.441 & 0.097 &\scalebox{0.8}[0.8]{Kaggle} \\
  \cmidrule(lr){2-8}
  & \scalebox{0.90}[0.9]{Favorita Transactions} & 0.08M & 1M & Daily & -5.481 & 0.362 &\scalebox{0.8}[0.8]{Kaggle} \\
  \cmidrule(lr){2-8}
  & \scalebox{0.90}[0.9]{Restaurant} & 0.29M & 2M & Daily & -4.650 & 0.126 &\scalebox{0.8}[0.8]{Kaggle} \\
  \cmidrule(lr){2-8}
  & \scalebox{0.90}[0.9]{Hierarchical Sales} & 0.21M & 1M & Daily & -8.704 & 0.078 &\scalebox{0.8}[0.8]{\citet{mancuso2021machine}} \\
  \midrule
  \multirow{2}{*}{Finance}
  & \scalebox{0.90}[0.9]{GoDaddy} & 0.26M & 2M & Monthly & -1.539 & 0.784 &\scalebox{0.8}[0.8]{Kaggle} \\
  \cmidrule(lr){2-8}
  & \scalebox{0.90}[0.9]{Bitcoin\tnote{*}} & 0.07M & 1M & Daily & -2.493 & 0.398 &\scalebox{0.8}[0.8]{\citet{godahewa2021monash}} \\
  \midrule
  \multirow{9}{*}{Misc.}
  & \scalebox{0.90}[0.9]{M1 Yearly} & 0.00M & 1M & Yearly & -0.791 & 0.473 &\scalebox{0.8}[0.8]{\citet{godahewa2021monash}} \\
  \cmidrule(lr){2-8}
  & \scalebox{0.90}[0.9]{M1 Quarterly} & 0.01M & 1M & Quarterly & -0.175 & 0.572 &\scalebox{0.8}[0.8]{\citet{godahewa2021monash}} \\
  \cmidrule(lr){2-8}
  & \scalebox{0.90}[0.9]{M1 Monthly} & 0.04M & 1M & Monthly & -1.299 & 0.588 &\scalebox{0.8}[0.8]{\citet{godahewa2021monash}} \\
  \cmidrule(lr){2-8}
  & \scalebox{0.90}[0.9]{M3 Yearly} & 0.02M & 1M & Yearly & -0.850 & 0.524 &\scalebox{0.8}[0.8]{\citet{godahewa2021monash}} \\
  \cmidrule(lr){2-8}
  & \scalebox{0.90}[0.9]{M3 Quarterly} & 0.04M & 1M & Quarterly & -0.897 & 0.624 &\scalebox{0.8}[0.8]{\citet{godahewa2021monash}} \\
  \cmidrule(lr){2-8}
  & \scalebox{0.90}[0.9]{M3 Monthly} & 0.1M & 1M & Monthly & -1.954 & 0.635 &\scalebox{0.8}[0.8]{\citet{godahewa2021monash}} \\
  \cmidrule(lr){2-8}
  & \scalebox{0.90}[0.9]{M3 Other} & 0.01M & 1M & - & -0.568 & 0.801 &\scalebox{0.8}[0.8]{\citet{godahewa2021monash}} \\
  \cmidrule(lr){2-8}
  & \scalebox{0.90}[0.9]{M4 Yearly} & 0.84M & 4M & Yearly & -0.036 & 0.533 &\scalebox{0.8}[0.8]{\citet{godahewa2021monash}} \\
  \cmidrule(lr){2-8}
  & \scalebox{0.90}[0.9]{M4 Quarterly} & 2.214M & 10M & Quarterly & -0.745 & 0.696 &\scalebox{0.8}[0.8]{\citet{godahewa2021monash}} \\
  \cmidrule(lr){2-8}
  & \scalebox{0.90}[0.9]{M4 Monthly} & 10.38M & 41M & Monthly & -1.358 & 0.665 &\scalebox{0.8}[0.8]{\citet{godahewa2021monash}} \\
  \pagebreak
  \multirow{3}{*}{Misc.}
  & \scalebox{0.90}[0.9]{M4 Weekly} & 0.37M & 2M & Weekly & -0.533 & 0.644 &\scalebox{0.8}[0.8]{\citet{godahewa2021monash}} \\
  \cmidrule(lr){2-8}
  & \scalebox{0.90}[0.9]{M4 Daily} & 9.96M & 39M & Daily & -1.332 & 0.841 &\scalebox{0.8}[0.8]{\citet{godahewa2021monash}} \\
  \cmidrule(lr){2-8}
  & \scalebox{0.90}[0.9]{M4 Hourly} & 0.35M & 2M & Hourly & -2.073 & 0.532 &\scalebox{0.8}[0.8]{\citet{godahewa2021monash}} \\
\end{longtable}
\end{sc}
\end{small}

\end{ThreePartTable}

\begin{figure*}[h]
\begin{center}
    \center{\includegraphics[width=\textwidth]{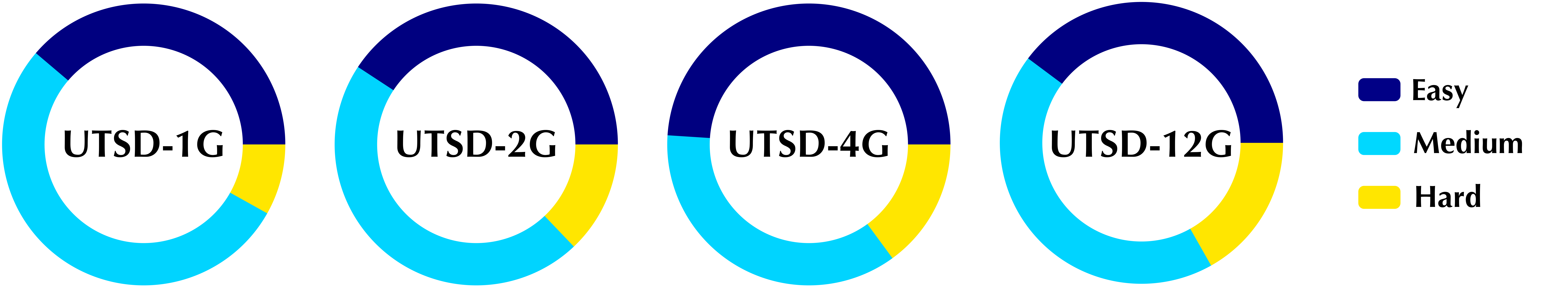}}
    \vspace{-10pt}
	\caption{Dataset complexity in each hierarchy of UTSD.}
	\label{fig:complexity}
\end{center}
\vspace{-10pt}
\end{figure*}

\subsection{UTSD Composition Analysis}\label{sec:data_filtering_strategy}

UTSD is constructed with hierarchical capacities, namely UTSD-1G, UTSD-2G, UTSD-4G, and UTSD-12G, where each smaller dataset is a subset of the larger ones. We adhere to the principle of progressively increasing the complexity and pattern diversity. Hierarchical structuring allows for a nuanced analysis that accounts for different levels of data granularity and complexity, ensuring that the pattern diversity is maintained across each hierarchy of the dataset. This approach not only facilitates a comprehensive evaluation across different scales but also ensures that each subset within the larger dataset offers a unique and incrementally challenging perspective, thus contributing to a more scalable pre-training.

\paragraph{Dataset Complexity}
We conduct a comprehensive analysis of individual datasets to obtain the stationarity and forecastability measures and construct the UTSD hierarchically regarding these indicators. Code for calculating the statistics is provided in the repository of UTSD. Consequently, based on the ${\operatorname{ADF-Statistic}}$ of each dataset, we categorized the predictive difficulty of the datasets into three levels: Easy, Medium, and Hard. The criteria are listed as follows: 
\begin{itemize}
    \item \textbf{Easy}: ${\operatorname{ADF-Statistic}} < -15.00;$
    \item \textbf{Medium}: $-15.00 \leq {\operatorname{ADF-Statistic}} < -5.00;$
    \item \textbf{Hard}: $-5.00 \leq {\operatorname{ADF-Statistic}}.$
\end{itemize}
For excessively long datasets in the temporal dimension, we additionally adopt the forecastability to assess the complexity of time series across different periods. As the capacity of UTSD  increases, the periods with low forecastability will be further incorporated correspondingly. In a nutshell, larger datasets contain a greater proportion of challenging tasks as shown in Figure~\ref{fig:complexity}, thereby escalating the complexity of the pre-training process. The hierarchy reflects an increase in the difficulty of patterns as the dataset size grows. This approach enables a structured examination of the learning challenges presented by different dataset sizes, underlining the intricate balance between data volume and pre-training difficulty.

\begin{figure*}[t]
\begin{center}
    \center{\includegraphics[width=\textwidth]{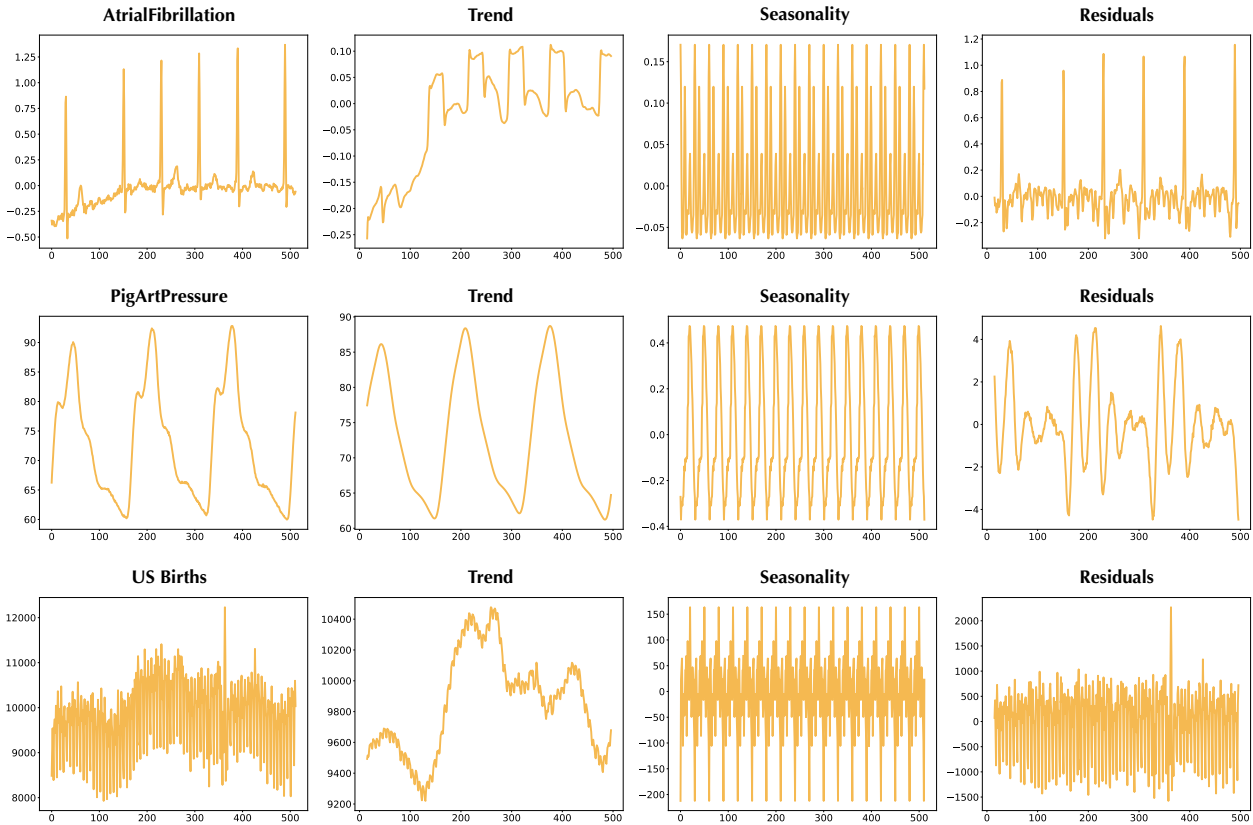}}
    \vspace{-20pt}
	\caption{Visualization of representative patterns in UTSD. Each time series is decomposed into trend, seasonal, and residual components.}
	\label{fig:dataset_showcase}
\end{center}
\vspace{-8pt}
\end{figure*}

\paragraph{Pattern Diversity}
Each dataset within the UTSD collection demonstrates unique patterns, highlighting the importance of maintaining pattern diversity. We build the UTSD dataset in a top-down manner, ensuring that each hierarchy within UTSD comprehensively represents all individual datasets and contains as many patterns as possible. As shown in Figure~\ref{fig:dataset_showcase}, we select several representative datasets for visualization analysis:
\begin{itemize}
    \item \textbf{AtrialFibrillation}: The dataset showcases a fluctuating trend with minimal seasonality. This pattern is an indicator of irregular heart rhythm characteristics, typical in medical recordings related to cardiac health. Such fluctuations, lacking a clear seasonal pattern, are crucial for understanding the unpredictable nature of atrial fibrillation.
    \item \textbf{PigArtPressure}: The dataset reveals a fluctuating trend interspersed with notable seasonality. This pattern is representative of the physiological variations in blood pressure that can occur due to environmental or internal factors. The presence of both fluctuating trends and seasonality in this dataset underscores the complex nature of biological data.
    \item \textbf{US Births}: The dataset distinctly exhibits a clear trend alongside pronounced seasonality. This pattern is characteristic of demographic data, where trends and seasonal variations can reflect socio-cultural factors and environmental influences. The consistent trend and seasonality in birth rates provide insights into population dynamics and reproductive behaviors.
\end{itemize}

To avoid selecting trivial temporal variations and provide a comprehensive representation of the varied patterns inherent in the individual datasets, we employ a downsampling technique for individual datasets. For those with a larger number of variates, we selectively choose representative variates that best encapsulate the distinct patterns of respective datasets. Similarly, for datasets with considerable temporal length, we resample them by the representative period. This methodical selection process ensures that the integrity and distinctive characteristics of each dataset are preserved, thereby maintaining the diversity of patterns across the hierarchical structure of the UTSD dataset.

\subsection{Experiments}

\paragraph{Forecasting benchmarks} In the field of time series forecasting, several classical datasets such as ETT~\cite{zhou2021informer}, ECL~\cite{wu2021autoformer}, Traffic~\cite{wu2021autoformer} and Weather~\cite{wu2021autoformer} have become widely recognized benchmarks for evaluating model performance. However, the variability in several datasets, such as ECL, is relatively homogeneous, and they do not adequately address aspects such as non-stationarity and predictability when assessing the strengths and weaknesses of models. Consequently, the development of a new benchmark is essential. Therefore, we have carefully considered factors such as domain, number of variables, frequency, non-stationarity, and predictability, and have selected a subset from the UTSD as the new benchmark. The datasets we have selected are presented in Table~\ref{tab:new_benchmarks}. Furthermore, we have evaluated our model along with other baseline models on these benchmarks. The results are presented in Table~\ref{tab:new_benchmark_results}. Admittedly, relying solely on these benchmarks is not sufficient to comprehensively assess model performance. We also look forward to the proposal of more diverse and comprehensive benchmarks in the future.
\begin{table*}[ht]
  \vspace{-5pt}
  \caption{Benchmark detailed descriptions. \emph{Time Point} denotes the total number of time points aggregating from all variates if multivariate. \emph{Frequency} denotes the sampling interval of time points, where ``-'' indicates no timestamp or irregular interval. \emph{ADF Statistic} denotes the Augmented Dickey-Fuller test statistics of the dataset. \emph{Forecastability} denotes the forecastability of the dataset.}
  \vskip 0.1in
  \label{tab:new_benchmarks}
  \footnotesize
  \begin{small}
  \begin{sc}
  \linespread{2}
  \renewcommand{\multirowsetup}{\centering}
  \renewcommand\arraystretch{1.4}
  \setlength{\tabcolsep}{5.1pt}
  \begin{tabular}{c|c|c|c|c|c|c}
    \toprule
    Domain & Dataset & Time Points & Variates & Frequency & ADF Statistic & Forecastability \\
    \toprule
     Environment & AustraliaRainfall & 11.54M & 3 & Hourly & -150.10 & 0.458 \\
     \midrule
     Transport & PedestrianCounts & 0.08M & 1 & Hourly & -23.462 & 0.297 \\
     \midrule
     IoT & SensorData & 3.24M & 18 & 0.002 Sec & -15.892 & 0.917 \\
     \midrule
     Health & BIDMC32HR & 0.04M & 1000 & - & -14.135 & 0.523 \\
    \bottomrule
  \end{tabular}
  \end{sc}
  \end{small}
  \vspace{-5pt}
\end{table*}

\begin{table*}[ht]
  \vspace{-5pt}
  \caption{Forecasting results on well-acknowledged deep forecasters and Timer, where Timer is pre-trained on the held-out datasets and then all models are superwisedly trained on the four datasets in the $672$-pred-$96$ setting.}
  \vskip 0.1in
  \label{tab:new_benchmark_results}
  \footnotesize
  \begin{small}
  \begin{sc}
  \linespread{2}
  \renewcommand{\multirowsetup}{\centering}
  \setlength{\tabcolsep}{12.4pt}
  \begin{tabular}{c|cc|cc|cc|cc}
    \toprule
    Models & \multicolumn{2}{c|}{\textbf{Timer}} & \multicolumn{2}{c|}{PatchTST} & \multicolumn{2}{c|}{iTransformer} & \multicolumn{2}{c}{DLinear}  \\ 
    \cmidrule(lr){1-1} \cmidrule(lr){2-3} \cmidrule(lr){4-5} \cmidrule(lr){6-7} \cmidrule(lr){8-9}
    Metric & MSE & MAE & MSE & MAE & MSE & MAE & MSE & MAE   \\
    \toprule
    AustraliaRainfall &  \textbf{0.800} & \textbf{0.720} & 0.802 & \textbf{0.720} & \textbf{0.800} & 0.800 & 0.804 & 0.804    \\
    \midrule
    PedestrianCounts &  \textbf{0.054} & \textbf{0.133} & 0.058 & 0.153 & 0.056 & 0.143 & 0.060 & 0.149    \\
    \midrule
    SensorData &  \textbf{0.049} & 0.094 & 0.056 & 0.094 & 0.052 & \textbf{0.091} & 0.057 & 0.111    \\
    \midrule
    BIDMC32HR &  \textbf{0.030} & \textbf{0.062} & 0.188 & 0.284 & 0.159 & 0.249 & 0.320 & 0.409    \\
    \bottomrule
  \end{tabular}
  \end{sc}
  \end{small}
\end{table*}

\paragraph{Domain transfer} To investigate the domain partitioning of UTSD, we use different domains of UTSD as the source and adapt the trained model to different target datasets to establish in-domain and out-of-domain transfer. The results in Table~\ref{tab:transfer} indicate that in-domain transfer can further enhance the downstream performance. Additionally, as the number of downstream data samples increases, the relative improvement of pre-training will gradually diminish, and even lead to negative transfer in some out-of-domain scenarios. It provides a promising direction to develop domain-specific models.

\begin{table*}[ht]
  \vspace{-5pt}
  \caption{In-domain and out-of-domain forecasting results by pre-training on the source domain and fine-tuning on the target dataset under different data scarcity. ECL and Weather belong to the Energy and Nature domains respectively.} 
  \vskip 0.1in
  \label{tab:transfer}
  \footnotesize
  \begin{small}
  \begin{sc}
  \linespread{2}
  \renewcommand{\multirowsetup}{\centering}
  \setlength{\tabcolsep}{5.2pt}
  \begin{tabular}{c|cc|cc|cc|cc|cc|cc}
    \toprule
    Target Dataset & \multicolumn{6}{c|}{Weather} & \multicolumn{6}{c}{ECL} \\
    \cmidrule(lr){1-1} \cmidrule(lr){2-7} \cmidrule(lr){8-13}
    Source Domain & \multicolumn{2}{c|}{From Scratch} & \multicolumn{2}{c|}{Energy} & \multicolumn{2}{c|}{\textbf{Nature}} & \multicolumn{2}{c|}{From Scratch} & \multicolumn{2}{c|}{Nature} & \multicolumn{2}{c}{\textbf{Energy}}  \\ 
    \cmidrule(lr){1-1} \cmidrule(lr){2-3} \cmidrule(lr){4-5} \cmidrule(lr){6-7} \cmidrule(lr){8-9} \cmidrule(lr){10-11} \cmidrule(lr){12-13}
    Metric & MSE & MAE & MSE & MAE & MSE & MAE & MSE & MAE & MSE & MAE & MSE & MAE  \\
    \toprule
    \ \ 5\% Target &  0.229 & 0.279 & 0.171 & 0.220 & \textbf{0.162} & \textbf{0.212} & 0.179 & 0.277 & 0.165 & 0.269 & \textbf{0.141} & \textbf{0.238}   \\
    \midrule
    20\% Target &  0.185 & 0.238 & 0.160 & 0.212 & \textbf{0.153} & \textbf{0.202} & 0.145 & 0.243 & 0.140 & 0.238 & \textbf{0.133} & \textbf{0.228}     \\
    \midrule
    100\% Target \ \ &   0.158 & 0.209 & 0.152 & 0.199 & \textbf{0.151} & \textbf{0.198} & \textbf{0.130} & 0.224 & 0.132 & 0.224 & 0.131 & \textbf{0.223}   \\
    \bottomrule
  \end{tabular}
  \end{sc}
  \end{small}
\end{table*}

\section{Implementation Details}

\subsection{Pre-training}\label{sec:pretrain_detail}

Based on the constructed UTSD datasets of different sizes and difficulties in the unified single series sequence (S3) format, Timer is pre-trained with increasing data sizes and model parameters to validate the scalability. Detailed configurations and parameter counts of the pre-trained models involved in this paper are provided in Table~\ref{tab:model_hyperparameters}. 

\begin{table*}[ht]
  \vspace{-5pt}
  \caption{Detailed model configurations of Timer and corresponding parameter counts. The number of heads for models is fixed as $8$.}
  \vskip 0.1in
  \label{tab:model_hyperparameters}
  \footnotesize
  \begin{small}
  \begin{sc}
  \linespread{2}
  \renewcommand{\multirowsetup}{\centering}
  \renewcommand\arraystretch{1.4}
  \setlength{\tabcolsep}{6.8pt}
  \begin{tabular}{c|cccc|cccc|cc}
    \toprule
    Scenario & \multicolumn{4}{c|}{Model Dim. Scale-up} & \multicolumn{4}{c|}{Layer Number. Scale-up} & \multicolumn{2}{c}{Others} \\
    \cmidrule(lr){1-1} \cmidrule(lr){2-5} \cmidrule(lr){6-9} \cmidrule(lr){10-11}
    Scale & 3M & 13M & 29M & 51M & 1M & 2M & 3M & 4M & 38M & 67M  \\ 
    \toprule
    Layers  &   6 & 6 & 6 & 6 & 2 & 4 & 6 & 8 & 8 & 8  \\
    \midrule
    Model Dim. & 256 & 512 & 768 & 1024 & 256 & 256 & 256 & 256 & 768 & 1024  \\
    \midrule
    FFN Dim.  &  512 & 1024 & 1536 & 2048 & 512 & 512 & 512 & 512 & 1536 & 2048  \\
    \midrule
    Parameters  &  3.21M & 12.72M & 28.51M & 50.59M & 1.10M & 2.16M & 3.21M & 4.27M & 37.97M& 67.40M \\
    \bottomrule
  \end{tabular}
  \end{sc}
  \end{small}
\end{table*}
All experiments are implemented in PyTorch~\citep{Paszke2019PyTorchAI} and trained using NVIDIA A100 Tensor Core GPU. We use AdamW~\cite{DBLP:journals/corr/KingmaB14} as the optimizer and cosine annealing algorithm for learning rate decay. The base learning rate is $5\times 10^{-5}$, and the final learning rate is $2\times 10^{-6}$ . The decay steps are proportional to the number of training steps of 10 epochs. During pre-training, we use $N=15$ as the number of tokens, and the batch size is set to 8192.

During the pre-training on the UTSD-1G to UTSD-4G, we adopt a global shuffle strategy by loading the whole time series into the memory. Due to the much greater data scale of UTSD-12G compared to any commonly used time series dataset in the past, it is difficult to load all 12GB of the pre-training dataset into memory for global shuffling. Therefore, we use a local shuffle strategy, which randomly selects and divides the 12GB pre-training dataset into three 4G subsets in the storage space through file selection and segmentation, and then takes turns loading them into memory for pre-training with global steps. In this strategy, we also ensure the continuity of learning rate decay.

\subsection{Downstream Tasks}\label{sec:finetune_detail}

We introduce the details of downstream experiments and present the generative scheme for each task, including time series forecast, imputation, and anomaly detection. Configurations for downstream adaptation are listed in Table~\ref{tab:finetune_hyperparameters}. Corresponding detailed results are provided in Section~\ref{sec:full_results}. And showcases of downstream tasks are shown in Figure~\ref{fig:show_forecast},~\ref{fig:show_anomaly_detection}, and~\ref{fig:show_imputation}.

\paragraph{Forecasting}\label{sec:forecast_detail}
The downstream forecasting task is tested on the real-world datasets, including (1) ETT~\cite{zhou2021informer} contains 7 variates of power transformers, with the period from July 2016 to July 2018, including four subsets and sampling intervals of one hour and fifteen minutes. (2) ECL~\cite{wu2021autoformer} mainly consists of hourly electricity consumption data from 321 customers (3) Traffic~\cite{wu2021autoformer} collected hourly road occupancy rates measured by 862 sensors on the San Francisco Bay Area highway from January 2015 to December 2016. (4) Weather~\cite{wu2021autoformer} consists of 21 meteorological variates collected every 10 minutes from the Max Planck Institute of Biogeochemistry meteorological station in 2020.
(5) PEMS contains California public transportation network data collected through a 5-minute window with the same four common subsets (PEMS03, PEMS04, PEMS07, PEMS08) used in SCINet~\cite{SCINet}.

\begin{table*}[ht]
  \vspace{-5pt}
  \caption{Downstream forecasting dataset descriptions. \emph{Split} denotes the number of time points in (train, validation, test) splits. \emph{Frequency} denotes the sampling interval of time points. \emph{Information}  denotes the domain in which the dataset belongs to.}
  \vskip 0.1in
  \label{tab:downstream_dataset}
  \footnotesize
  \begin{small}
  \begin{sc}
  \linespread{2}
  \renewcommand{\multirowsetup}{\centering}
  \renewcommand\arraystretch{1.4}
  \setlength{\tabcolsep}{17.9pt}
  \begin{tabular}{c|c|c|c|c}
    \toprule
    Dataset & Variate & Split & Frequency & Information \\
    \toprule
     ETTh1, ETTh2 & 7  & (8545, 2881, 2881) & Hourly & Electricity\\
     \midrule
     ETTm1, ETTm2 & 7  & (34465, 11521, 11521) & 15min & Electricity\\
    \midrule
    ECL & 321  & (18317, 2633, 5261) & Hourly & Electricity \\
    \midrule
    Traffic & 862  & (12185, 1757, 3509) & Hourly & Transportation \\
    \midrule
    Weather & 21  & (36792, 5271, 10540) & 10min & Weather\\
    \midrule
    PEMS03 & 358  & (15617, 5135, 5135) & 5min & Transportation\\
    \midrule
    PEMS04 & 307  & (10172,3375, 3375) & 5min & Transportation\\
    \midrule
    PEMS07 & 883  & (16911, 5622, 5622) & 5min & Transportation\\
    \midrule
    PEMS08 & 170  & (10690, 3548, 3548) & 5min & Transportation\\
    \bottomrule
  \end{tabular}
  \end{sc}
  \end{small}
\end{table*}

We adopt the autoregressive generation training objective~\cite{bengio2000neural} for downstream forecasting datasets in the fine-tuning stage. Specifically, we divide the lookback length into $N=7$ tokens with the segment length $S=96$. The model naturally outputs $N$ next tokens, which we calculate the mean squared error (MSE) of the $N$ tokens with corresponding ground truth and backpropagate the loss. During inference, we conduct iterative multi-step forecasting by concatenating the forecasted result with the lookback series and repeatedly adopting the model to generate the next token until the total length of predicted tokens reaches the expected length. If exceeding the predicted length, we will crop the excess value of the end. 

For constructing data-scarce scenarios, we perform retrieval with the uniform interval in the training split according to the sampling ratio and conduct random shuffling at the end of each epoch to train the model. The construction pipeline with the fixed random seed ensures the reproducibility of our experimental results. In order to maintain comparability with previous benchmarks, we keep the same validation and testing sets of original downstream datasets and train the baseline model and Timer with the same set of training samples.

\paragraph{Imputation}\label{sec:imputation_detail}
Considering the real-world scenario that missing values at time points often appear in succession, we adjust the previous point-level imputation proposed by TimesNet~\cite{wu2022timesnet} and increase the difficulty of the task, that is, changing the masked unit from point to time series segment. The protocol poses challenges to recovering successive points, which underscore higher demands for the model capacity to restore a span of series variations. Concretely, for a time series consisting of $N=8$ segments with a length of $24$, we randomly mask several segments as zeros except for the first segment, ensuring that the first segment is observed by the model to learn about initial series variations for imputation. For the training objective of downstream adaption, we adopt the denoising autoencoding~\cite{raffel2020exploring}, which takes the masked parts as special tokens and unmasked segments as tokens as the model input. Due to the generative capability of Timer acquired by pre-training, we regard outputted tokens as the next predicted tokens and backpropagate the reconstruction error between the generated next token of the masked segment with the ground truth. During inference, we take the MSE of the masked segments as the indicator to evaluate the imputation performance. Based on the above protocol, we conduct the imputation task on the same datasets of the forecasting task in Table~\ref{tab:downstream_dataset}.

\begin{table*}[ht]
  \vspace{-5pt}
  \caption{Detailed explanation of model hyperparameters and corresponding parameter quantities. We adopt the learning rate schedule strategy with exponential decay at a base of $0.5$ under all three downstream tasks.}
  \vskip 0.1in
  \label{tab:finetune_hyperparameters}
  \footnotesize
  \begin{small}
  \begin{sc}
  \linespread{2}
  \renewcommand{\multirowsetup}{\centering}
  \setlength{\tabcolsep}{10.4pt}
  \renewcommand\arraystretch{1.4}
  \begin{tabular}{c|c|c|c|c|c|c|c|c}
    \toprule
    \multirow{2}{*}{Tasks} & \multicolumn{4}{c}{Model Hyper-parameter} & \multicolumn{4}{c}{Training Process} \\
    \cmidrule(lr){2-5}\cmidrule(lr){6-9}
    & $L_{min}$ & $L_{max}$ & $d_{min}$ $^\dag$ & $d_{max}$ $^\dag$ & LR$^\ast$ & Loss & Batch Size & Epochs\\
    \toprule
    Forecasting & 2 & 8 & 256 & 2048 & 3e-5 & MSE & 2048 & 10 \\
    \midrule
    Imputation & 4 & 4 & 256 & 256 & 3e-5 & MSE & 32 & 10 \\
    \midrule
    Anomaly Detection & 4 & 4 & 256 & 256 & 3e-5 & MSE & 128 & 10 \\
    \bottomrule
  \end{tabular}
      \begin{tablenotes}
        \footnotesize
        \item[] $\ast$ LR means the initial learning rate.
  \end{tablenotes}
  \end{sc}
  \end{small}
\end{table*}

\paragraph{Anomaly detection}\label{sec:anomaly_detail}
For anomaly detection, prevalent protocols represented by Anomaly Transformer~\cite{xu2021anomaly} and TimesNet~\cite{wu2022timesnet} adopt the reconstructive approach that learns a feature extractor to reconstruct raw series, and the output is regarded as standard values. With all the mean squared errors between the standard and input series from the datasets, a specific threshold with the given quantile is determined to label the anomalies. 

Considering the prevalent scenarios of anomaly detection by monitoring real-time measurements, the quick judgment of on-the-fly time series anomaly can be more practical in the real world. Therefore, we propose a predictive protocol of anomaly detection based on generative models. Concretely, we use the observed segments to predict the future segment, and the predicted segment will be established as the standard to be compared with the actual value received. We adopt the UCR Anomaly Archive proposed by~\citet{wu2021current}. The task is to find the position of an anomaly in the test series based on a single normal series for training, which is an extremely data-scarce scenario with only one available sample. For downstream adaption, we adopt the same next token prediction as the pre-training, that is, training Timer with the lookback series containing $N=7$ segments of the length $S=96$ to generate the next token with length $96$, which is regarded as the standard value. After training, we record the MSE of all segments in the test set and sort them in descending order. We find the first segment hit the anomaly interval labeled in the dataset within the first $\alpha$ quantile, and we record the quantile. Based on the above protocol, the real-time judgment ability of the model for sudden anomalies can be predictively examined. Detailed quantiles of Timer in $250$ tasks are provided in Table~\ref{tab:anomaly_detection}. With more complex time series anomalies introduced in UCR Anomaly Archive, we hope to establish a reasonable and challenging benchmark in the field of anomaly detection.

\paragraph{Zero-shot forecasting} \label{sec:zero-shot}
We conduct zero-shot forecasting experiments on seven datasets from iTransformer~\cite{liu2023itransformer}. Notably, PEMS datasets are not included, as they have already appeared in the LOSTA dataset for pre-training. We apply the same data-split strategy as Autoformer~\cite{wu2021autoformer} and calculate the averaged MSE of all predict-$96$ windows in the test split. We evaluate five open-source large time series models, including Timer, Moirai~\cite{woo2024unified}, TimesFM~\cite{das2023decoder}, Chronos~\cite{ansari2024chronos}, and MOMENT~\cite{goswami2024moment}. We further assess the qualities in Table~\ref{tab:ltsm_comp}, which includes more LTSMs and summarizes several attributes and abilities of large models.
\begin{itemize}
    \item \textbf{MOMENT}: MOMENT\footnote{\href{https://huggingface.co/AutonLab/MOMENT-1-large}{https://huggingface.co/AutonLab/MOMENT-1-large}} trained by masking modeling is applied to zero-shot forecasting by concatenating the lookback series with a mask with the length to be predicted. The mask through the model is regarded as the prediction.
    \item \textbf{Chronos}: Chronos\footnote{\href{https://huggingface.co/amazon/chronos-t5-large}{https://huggingface.co/amazon/chronos-t5-large}} is a probabilistic forecaster. \emph{Chronos-S1} means sampling one prediction trajectory and \emph{Chronos-S20} means sampling 20 trajectories and using the average trajectory.
    \item \textbf{TimesFM}: We use the official checkpoint from HuggingFace\footnote{\href{https://huggingface.co/google/timesfm-1.0-200m}{https://huggingface.co/google/timesfm-1.0-200m}}, which supports various input and output lengths.
    \item \textbf{Moirai}: The Moirai family\footnote{\href{https://huggingface.co/collections/Salesforce/moirai-10-r-models-65c8d3a94c51428c300e0742}{https://huggingface.co/collections/Salesforce/moirai-10-r-models-65c8d3a94c51428c300e0742}} has three different sizes, labeled as \emph{Moirai-S}, \emph{Moirai-M}, and \emph{Moirai-L}.
    \item \textbf{Timer}: We provide three versions with increased scopes of pre-training. \emph{Timer-1B} is pre-trained on UTSD; \emph{Timer-16B} is pre-trained on UTSD and Buildings900K~\cite{emami2023buildingsbench}; and \emph{Timer-28B} is pre-trained on UTSD and LOTSA.
\end{itemize}

\begin{table*}[ht]
  \vspace{-15pt}
  \caption{Quality evaluation of large time series models. \emph{Architecture} denotes the Transformer category. \emph{Model size} presents the parameter counts. \emph{Token type} presents the graininess of time series tokens. \emph{Context length} means the maximum/fixed input length of the model.}
  \vspace{-5pt}
  \label{tab:ltsm_comp}
  \vskip 0.15in
  \centering
  \begin{small}
  \begin{sc}
  \renewcommand{\multirowsetup}{\centering}
  \setlength{\tabcolsep}{2.5pt}
  \renewcommand\arraystretch{1.2}
  \begin{tabular}{c|ccccccc}
    \toprule
    \multirow{2}{*}{Method} & \textbf{Timer} & Moirai & MOMENT & Chronos & Lag-LLama & TimesFM & TimeGPT-1   \\ 
     & \textbf{(Ours)}  & \citeyearpar{woo2024unified}& \citeyearpar{goswami2024moment}  & \citeyearpar{ansari2024chronos} &
     \citeyearpar{rasul2023lag} &  \citeyearpar{das2023decoder}  & \citeyearpar{garza2023timegpt} \\
    \toprule
     \multirow{2}{*}{Architecture} & Decoder  & Encoder & Encoder & Encoder  & Decoder & Decoder & Encoder\\
     &  &  &  & Decoder & & & Decoder \\
    \midrule
    \multirow{2}{*}{Model size} & 29M, 50M,  & 14M, 91M, & 40M, 125M & 20M, 46M,  & 200M  & 17M, 70M,  & UnKnown \\
     & 67M & 311M & 385M & 200M, 710M & & 200M  & \\
    \midrule
    \multirow{3}{*}{Supported tasks} & Forecast  & Forecast & \scalebox{0.8}{Forecast Imputation} & Forecast & Forecast & Forecast  & Forecast
 \\
     & Imputation & &\scalebox{0.8}{Classification}& & &  & Detection\\
     & Detection & & \scalebox{0.8}{Detection}& & &  & \\
    \midrule
    Pre-training Scale & 28B & 27.65B & 1.13B & 84B & 0.36B & 100B & 100B \\
    \midrule
    Token type & Segment & Segment & Segment & Point & Point & Segment & Segment \\
    \midrule
    Context length & $\le$1440 & $\le$5000 & = 512 & $\le$512 & $\le$1024 & $\le$512 & Unknown  \\
    \midrule
    Variable length & True & True & False & True & True & True & True \\
    \midrule
    Probabilistic & False & True & False & True & True & True & True \\
    \bottomrule
  \end{tabular}
  \end{sc}
  \end{small}
  \vspace{-5pt}
\end{table*}

\section{Full Results}\label{sec:full_results}
\subsection{Time Series Forecasting}
We provide all the results of the forecasting task in Figure~\ref{fig:forecast_promotion}. As shown in Table~\ref{tab:forecast_promotion_full}, we include six representative real-world datasets, demonstrating that Timer achieves state-of-the-art forecasting performance and the large-scale pre-training helps to alleviate performance degradation as the available downstream samples decrease.

\begin{table*}[ht]
  \vspace{-5pt}
  \caption{Full forecasting results of Timer obtained by training from scratch (None) and fine-tuning from UTSD-12G pre-trained model. The bold values we use indicate that the pre-trained model results have positive benefits compared to from scratch. We attach the current state-of-the-art results as \emph{SOTA} in this table, including PatchTST~\cite{nie2022time} on ETTh1 and Weather, as well as iTransformer~\cite{liu2023itransformer} on ECL, Traffic, PEMS03, and PEMS04. We adopt the unified lookback length as $672$ and the forecast length as $96$.}
  \vskip 0.1in
  \label{tab:forecast_promotion_full}
  \footnotesize
  \begin{small}
  \begin{sc}
  \linespread{2}
  \renewcommand{\multirowsetup}{\centering}
  \setlength{\tabcolsep}{6.1pt}
  \renewcommand\arraystretch{1.4}
  \begin{tabular}{c|cc|cc|cc|cc|cc|cc}
    \toprule
    Dataset & \multicolumn{2}{c|}{ETTh1} & \multicolumn{2}{c|}{ECL} & \multicolumn{2}{c|}{Traffic} & \multicolumn{2}{c|}{Weather} & \multicolumn{2}{c|}{PEMS03} & \multicolumn{2}{c}{PEMS04} \\
    \cmidrule(lr){1-1} \cmidrule(lr){2-3} \cmidrule(lr){4-5} \cmidrule(lr){6-7} \cmidrule(lr){8-9} \cmidrule(lr){10-11} \cmidrule(lr){12-13}
    Pre-trained & None & 12G  & None & 12G  & None & 12G  & None & 12G & None & 12G & None & 12G  \\
    \toprule
    100\%  & 0.363 & \textbf{0.358} & 0.132 & 0.136 & 0.352 & \textbf{0.351} & 0.165 & \textbf{0.154} & 0.126 & \textbf{0.118} & 0.125 & \textbf{0.107}\\
    \midrule
    75\%  &   0.364 & \textbf{0.358} & 0.132 & 0.137 & 0.353 & \textbf{0.351} & 0.162 & \textbf{0.157} & 0.124 & \textbf{0.114} & 0.126 & \textbf{0.110}   \\
    \midrule
    50\%  &   0.370 & \textbf{0.356} & 0.132 & 0.135 & 0.356 & \textbf{0.352} & 0.161 & \textbf{0.151}  & 0.129 & \textbf{0.114} & 0.131 & \textbf{0.110}  \\
    \midrule
    25\%  &   0.387 & \textbf{0.359} & 0.135 & \textbf{0.134} & 0.368 & \textbf{0.352} & 0.162 & \textbf{0.153}  & 0.133 & \textbf{0.114} & 0.141 & \textbf{0.117} \\
    \midrule
    20\%  &   0.385 & \textbf{0.359} & 0.137 & \textbf{0.134} & 0.372 & \textbf{0.352} & 0.166 & \textbf{0.151}  & 0.135 & \textbf{0.116} & 0.145 & \textbf{0.120}  \\
    \midrule
    15\%  & 0.391 & \textbf{0.360} & 0.141 & \textbf{0.134} & 0.379 & \textbf{0.353} & 0.174 & \textbf{0.152}  & 0.138 & \textbf{0.118} & 0.152 & \textbf{0.123} \\
    \midrule
    10\%  & 0.426 & \textbf{0.361} & 0.144 & \textbf{0.133} & 0.387 & \textbf{0.353} & 0.182 & \textbf{0.152}  & 0.140 & \textbf{0.120} & 0.165 & \textbf{0.126}\\
    \midrule
    5\% &  0.426 & \textbf{0.362} & 0.154 & \textbf{0.132} & 0.407 & \textbf{0.361} & 0.198 & \textbf{0.151}  & 0.158 &\textbf{ 0.125} & 0.195 & \textbf{0.135}  \\
    \midrule
    4\% &   0.424 & \textbf{0.362} & 0.161 & \textbf{0.135} & 0.416 & \textbf{0.366} & 0.208 & \textbf{0.152}  & 0.166 & \textbf{0.127} & 0.210 & \textbf{0.138} \\
    \midrule
    3\% & 0.427 & \textbf{0.363} & 0.169 & \textbf{0.134} & 0.431 & \textbf{0.369} & 0.218 & \textbf{0.153}  & 0.180 & \textbf{0.131} & 0.234 & \textbf{0.143}  \\
    \midrule
    2\% & 0.427 & \textbf{0.363} & 0.186 & \textbf{0.137} & 0.467 & \textbf{0.380} & 0.230 & \textbf{0.159}  & 0.201 & \textbf{0.137} & 0.257 & \textbf{0.152} \\
    \midrule
    1\% & 0.428 & \textbf{0.366} & 0.215 & \textbf{0.140} & 0.545 & \textbf{0.390} & 0.246 & \textbf{0.166}  & 0.249 & \textbf{0.151} & 0.320 & \textbf{0.172} \\
    \toprule
    SOTA & \multicolumn{2}{c|}{0.370} & \multicolumn{2}{c|}{0.129} & \multicolumn{2}{c|}{0.360} & \multicolumn{2}{c|}{0.149} & \multicolumn{2}{c|}{0.132} & \multicolumn{2}{c}{0.115} \\
    \bottomrule
  \end{tabular}
  \end{sc}
  \end{small}
\end{table*}

\subsection{Imputation}
\begin{figure*}[ht]
\begin{center}
    \center{\includegraphics[width=\textwidth]{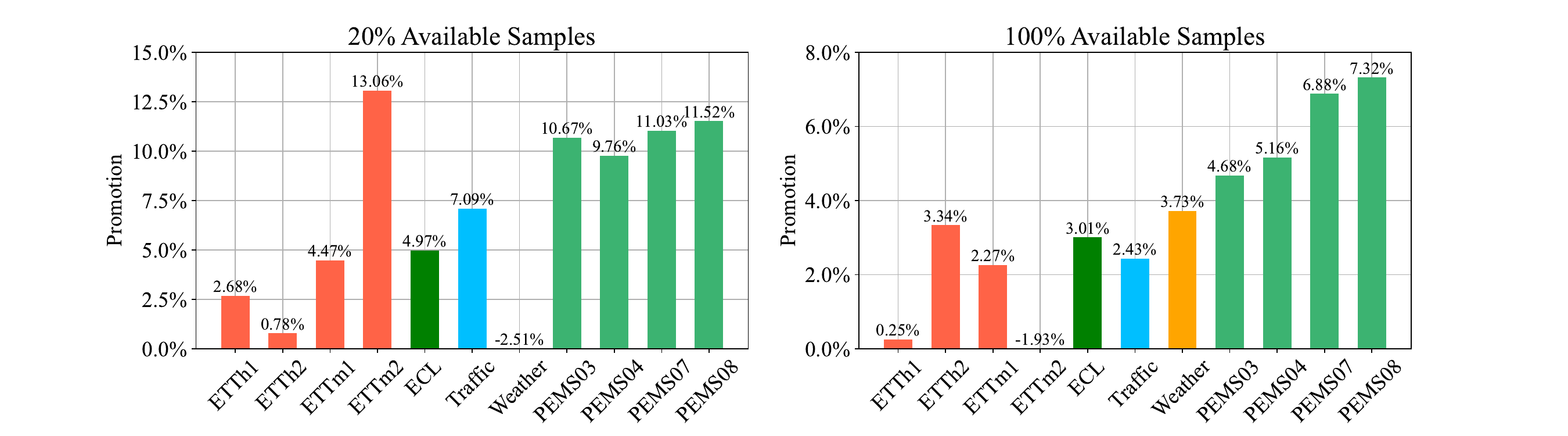}}
    \vspace{-25pt}
	\caption{Pre-training benefit of Timer on the downstream imputation task with $20\%$ and $100\%$ available samples. Complete results details used to calculate the pre-training benefits relative to training from scratch in the figures are listed in Table~\ref{tab:imputation_full}-\ref{tab:imputation_timesnet}.}
	\label{fig:imputation_full}
\end{center}
\end{figure*}

In this section, we provide the detailed results of the imputation task, including Timer trained from scratch and adapting pre-trained models with $5\%$ available samples in Table~\ref{tab:imputation_full}, $20\%$ samples in Table~\ref{tab:imputation_full2}, and full samples in Table~\ref{tab:imputation_full3} on the downstream task. We also report the results of TimesNet at the above three ratios in Table~\ref{tab:imputation_timesnet}. Based on the result, we provided an improvement in imputation performance before and after pre-training with $\{5\%, 20\%, 100\%\}$ samples in Figure~\ref{fig:imputation}~and~\ref{fig:imputation_full}, reflecting the benefits of autoregressive pre-training in segment-wise imputation task.

\begin{table*}[ht]
  \caption{Downstream imputation with $5\%$ samples. Pre-training benefit $\Delta\%$ is calculated as the ratio of decreased imputing error in MSE. In the case of $5\%$ samples, our pre-trained model outperforms TimesNet (Table~\ref{tab:imputation_timesnet}) in all $44$ settings on datasets and masked ratios.}
  \vskip 0.1in
  \label{tab:imputation_full}
  \footnotesize
  \begin{small}
  \begin{sc}
  \linespread{2}
  \renewcommand{\multirowsetup}{\centering}
  \renewcommand\arraystretch{1.4}
  \setlength{\tabcolsep}{5.35pt}
  \begin{tabular}{c|ccc|ccc|ccc|ccc}
    \toprule
    Mask Ratio & \multicolumn{3}{c|}{12.5\%} & \multicolumn{3}{c|}{25.0\%} & \multicolumn{3}{c|}{37.5\%} & \multicolumn{3}{c}{50.0\%} \\
    \cmidrule(lr){1-1} \cmidrule(lr){2-4} \cmidrule(lr){5-7} \cmidrule(lr){8-10} \cmidrule(lr){11-13}
    Pre-trained & None & 12G & $\Delta\%$ & None & 12G & $\Delta$\% & None & 12G & $\Delta\%$ & None & 12G & $\Delta\%$ \\
    \toprule
    ETTh1  & 0.301 & 0.292 & \textbf{+3.08} & 0.313 & 0.299 & \textbf{+4.46} & 0.322 & 0.307 & \textbf{+4.59} & 0.325 & 0.325 & 0.00\\
    \midrule
    ETTh2  &   0.172 & 0.168 & \textbf{+2.64} & 0.182 & 0.180 & \textbf{+1.26} & 0.197 & 0.190 & \textbf{+3.22} & 0.216 & 0.215 & \textbf{+0.47}  \\
    \midrule
    ETTm1  &   0.397 & 0.347 & \textbf{+12.52} & 0.403 & 0.332 & \textbf{+17.72} & 0.428 & 0.374 & \textbf{+12.77} & 0.473 & 0.425 & \textbf{+10.13}  \\
    \midrule
    ETTm2  &   0.118 & 0.116 & \textbf{+1.59} & 0.127 & 0.121 & \textbf{+4.69} & 0.134 & 0.131 & \textbf{+2.22} & 0.147 & 0.144 & \textbf{+1.99} \\
    \midrule
    ECL  &   0.152 & 0.140 &  \textbf{+7.67} & 0.162 & 0.150 & \textbf{+7.27} & 0.172 & 0.161 & \textbf{+6.76} & 0.185 & 0.174 & \textbf{+6.17}   \\
    \midrule
    Traffic  & 0.538 & 0.460 & \textbf{+14.60} & 0.567 & 0.487 & \textbf{+14.14} & 0.598 & 0.520 & \textbf{+13.16} & 0.633 & 0.558 & \textbf{+11.91} \\
    \midrule
    Weather  & 0.113 & 0.117 & -3.18 & 0.116 & 0.114 & \textbf{+2.31} & 0.128 & 0.124 & \textbf{+3.28} & 0.155 & 0.136 & \textbf{+12.42} \\
    \midrule
    PEMS03 &  0.160 & 0.135 & \textbf{+15.78} & 0.196 & 0.168 & \textbf{+14.60} & 0.257 & 0.223 & \textbf{+13.51} & 0.354 & 0.306 & \textbf{+13.49}  \\
    \midrule
    PEMS04 &   0.193 & 0.161 & \textbf{+16.80} & 0.238 & 0.202 & \textbf{+15.30} & 0.305 & 0.258 & \textbf{+15.28} & 0.410 & 0.348 & \textbf{+15.14} \\
    \midrule
    PEMS07 & 0.166 & 0.139 & \textbf{+16.19} & 0.210 & 0.183 & \textbf{+12.89} & 0.278 & 0.243 & \textbf{+12.72} & 0.378 & 0.326 & \textbf{+13.76}  \\
    \midrule
    PEMS08 &  0.185 & 0.157 & \textbf{+15.33} & 0.232 & 0.195 & \textbf{+15.98} & 0.303 & 0.265 & \textbf{+12.75} & 0.417 & 0.362 & \textbf{+13.26} \\
    
    \bottomrule
  \end{tabular}
  \end{sc}
  \end{small}
\end{table*}
\begin{table*}[ht]
  \caption{Downstream imputation with $20\%$ samples. Pre-training benefit $\Delta\%$ is calculated as the ratio of decreased imputing error in MSE. In the case of $20\%$ samples, our pre-trained model outperforms TimesNet in $86.4\%$ of $44$ settings on datasets and masked ratios.}
  \vskip 0.1in
  \label{tab:imputation_full2}
  \footnotesize
  \begin{small}
  \begin{sc}
  \linespread{2}
  \renewcommand{\multirowsetup}{\centering}
  \setlength{\tabcolsep}{6pt}
  \renewcommand\arraystretch{1.4}
  \begin{tabular}{c|ccc|ccc|ccc|ccc}
    \toprule
    Mask Ratio & \multicolumn{3}{c|}{12.5\%} & \multicolumn{3}{c|}{25.0\%} & \multicolumn{3}{c|}{37.5\%} & \multicolumn{3}{c}{50.0\%} \\
    \cmidrule(lr){1-1} \cmidrule(lr){2-4} \cmidrule(lr){5-7} \cmidrule(lr){8-10} \cmidrule(lr){11-13}
    Pre-trained & None & 12G & $\Delta\%$ & None & 12G & $\Delta$\% & None & 12G & $\Delta\%$ & None & 12G & $\Delta\%$ \\
    \toprule
    ETTh1  & 0.289  & 0.278  & \textbf{3.83} & 0.293  & 0.287  & \textbf{1.91} & 0.305  & 0.297  & \textbf{2.56} & 0.322  & 0.314  & \textbf{2.44}\\
    \midrule
    ETTh2  &   0.168  & 0.166  & \textbf{1.21} & 0.180  & 0.178  & \textbf{1.02} & 0.192  & 0.190  & \textbf{0.73} & 0.208  & 0.208  & \textbf{0.17}  \\
    \midrule
    ETTm1  &   0.349  & 0.328  & \textbf{5.98} & 0.335  & 0.326  & \textbf{2.72} & 0.378  & 0.360  & \textbf{4.84} & 0.426  & 0.407  & \textbf{4.34}  \\
    \midrule
    ETTm2  &   0.139  & 0.133  & \textbf{4.77} & 0.158  & 0.123  & \textbf{22.30} & 0.176  & 0.136  & \textbf{22.95} & 0.146  & 0.143  & \textbf{2.22} \\
    \midrule
    ECL  &   0.136  & 0.130  & \textbf{5.01} & 0.146  & 0.138  & \textbf{5.30} & 0.157  & 0.149  & \textbf{5.04} & 0.170  & 0.162  & \textbf{4.54}   \\
    \midrule
    Traffic  & 0.451  & 0.420  & \textbf{6.89} & 0.481  & 0.446  & \textbf{7.26} & 0.513  & 0.477  & \textbf{7.09} & 0.550  & 0.511  & \textbf{7.10} \\
    \midrule
    Weather  & 0.125  & 0.129  & -3.40 & 0.125  & 0.147  & -17.46 & 0.154  & 0.125  & \textbf{18.77} & 0.141  & 0.153  & -7.93 \\
    \midrule
    PEMS03 &  0.134  & 0.120  & \textbf{10.41} & 0.169  & 0.150  & \textbf{11.35} & 0.221  & 0.198  & \textbf{10.61} & 0.305  & 0.273  & \textbf{10.32}  \\
    \midrule
    PEMS04 &   0.162  & 0.146  & \textbf{9.93} & 0.203  & 0.184  & \textbf{9.61} & 0.262  & 0.236  & \textbf{9.86} & 0.354  & 0.320  & \textbf{9.65} \\
    \midrule
    PEMS07 & 0.140  & 0.125  & \textbf{10.96} & 0.182  & 0.162  & \textbf{10.85} & 0.240  & 0.214  & \textbf{10.77} & 0.327  & 0.290  & \textbf{11.57}  \\
    \midrule
    PEMS08 &  0.155  & 0.139  & \textbf{10.39} & 0.198  & 0.174  & \textbf{12.11} & 0.268  & 0.236  & \textbf{11.94} & 0.366  & 0.324  & \textbf{11.63} \\
    \bottomrule
  \end{tabular}
  \end{sc}
  \end{small}
\end{table*}
\begin{table*}[ht]
  \caption{Downstream imputation with $100\%$ samples. Pre-training benefit $\Delta\%$ is calculated as the ratio of decreased imputing error in MSE. In the case of $100\%$ samples, our pre-trained model outperforms TimesNet in $56.8\%$ of $44$ settings on datasets and masked ratios.}
  \vskip 0.1in
  \label{tab:imputation_full3}
  \footnotesize
  \begin{small}
  \begin{sc}
  \linespread{2}
  \renewcommand{\multirowsetup}{\centering}
  \setlength{\tabcolsep}{6pt}
  \renewcommand\arraystretch{1.4}
  \begin{tabular}{c|ccc|ccc|ccc|ccc}
    \toprule
    Mask Ratio & \multicolumn{3}{c|}{12.5\%} & \multicolumn{3}{c|}{25.0\%} & \multicolumn{3}{c|}{37.5\%} & \multicolumn{3}{c}{50.0\%} \\
    \cmidrule(lr){1-1} \cmidrule(lr){2-4} \cmidrule(lr){5-7} \cmidrule(lr){8-10} \cmidrule(lr){11-13}
    Pre-trained & None & 12G & $\Delta\%$ & None & 12G & $\Delta$\% & None & 12G & $\Delta\%$ & None & 12G & $\Delta\%$ \\
    \toprule
    ETTh1  & 0.274 & 0.273 & \textbf{0.34} & 0.283 & 0.283 & -0.04 & 0.295 & 0.294 & \textbf{0.52} & 0.313 & 0.312 & \textbf{0.17}\\
    \midrule
    ETTh2  &   0.207 & 0.177 & \textbf{14.44} & 0.186 & 0.186 & -0.49 & 0.192 & 0.195 & -1.18 & 0.210 & 0.209 & \textbf{0.57}  \\
    \midrule
    ETTm1  &   0.342 & 0.352 & -3.04 & 0.359 & 0.345 & \textbf{3.87} & 0.400 & 0.371 & \textbf{7.09} & 0.418 & 0.413 & \textbf{1.15}  \\
    \midrule
    ETTm2  &   0.149 & 0.161 & -8.01 & 0.153 & 0.171 & -11.46 & 0.173 & 0.176 & -1.53 & 0.183 & 0.158 & \textbf{13.27} \\
    \midrule
    ECL  &   0.125 & 0.122 & \textbf{2.98} & 0.134 & 0.130 & \textbf{3.06} & 0.144 & 0.139 & \textbf{3.12} & 0.157 & 0.152 &\textbf{2.87}   \\
    \midrule
    Traffic  & 0.402 & 0.392 & \textbf{2.50} & 0.424 & 0.414 & \textbf{2.48} & 0.454 & 0.443 & \textbf{2.46} & 0.488 & 0.477 & \textbf{2.29} \\
    \midrule
    Weather  & 0.144 & 0.157 & -8.67 & 0.159 & 0.146 & \textbf{8.01} & 0.162 & 0.147 & \textbf{9.41} & 0.168 & 0.158 & \textbf{6.15} \\
    \midrule
    PEMS03 &  0.113 & 0.108 &\textbf{4.65} & 0.143 & 0.135 & \textbf{5.30} & 0.188 & 0.179 & \textbf{4.88} & 0.258 & 0.248 & \textbf{3.90}  \\
    \midrule
    PEMS04 &  0.142 & 0.134 & \textbf{5.23} & 0.176 & 0.166 & \textbf{5.39} & 0.227 & 0.216 & \textbf{5.24} & 0.311 & 0.296 & \textbf{4.77} \\
    \midrule
    PEMS07 & 0.121 & 0.114 & \textbf{5.81} & 0.155 & 0.144 & \textbf{6.50} & 0.204 & 0.189 & \textbf{7.36} & 0.277 & 0.256 & \textbf{7.85}  \\
    \midrule
    PEMS08 &  0.137 & 0.129 & \textbf{5.89} & 0.169 & 0.157 & \textbf{7.29} & 0.224 & 0.206 & \textbf{7.70} & 0.314 & 0.288 & \textbf{8.39} \\
    
    \bottomrule
  \end{tabular}
  \end{sc}
  \end{small}
\end{table*}
\begin{table*}[ht]
  \caption{Full results of downstream imputation of TimesNet under different data scarcities as the baseline.}
  \vskip 0.1in
  \label{tab:imputation_timesnet}
  \footnotesize
  \begin{small}
  \begin{sc}
  \linespread{2}
  \renewcommand{\multirowsetup}{\centering}
  \setlength{\tabcolsep}{4.7pt}
  \renewcommand\arraystretch{1.4}
  \begin{tabular}{c|cccc|cccc|cccc}
    \toprule
    Sample Ratio & \multicolumn{4}{c|}{5\%} & \multicolumn{4}{c|}{20\%} & \multicolumn{4}{c}{100\%} \\
    \cmidrule(lr){1-1} \cmidrule(lr){2-5} \cmidrule(lr){6-9} \cmidrule(lr){10-13} 
    Mask Ratio & 12.5\% & 25.0\% & 37.5\% & 50.0\% & 12.5\% & 25.0\% & 37.5\% & 50.0\% & 12.5\% & 25.0\% & 37.5\% & 50.0\% \\
    \toprule
    ETTh1  & 0.676 & 0.671 & 0.678 & 0.682 & 0.684 & 0.687 & 0.675 & 0.679 & 0.284 & 0.296 & 0.269 & 0.289 \\
    \midrule
    ETTh2  &   0.258 & 0.249 & 0.272 & 0.276 & 0.252 & 0.245 & 0.251 & 0.268 & 0.178 & 0.199 & 0.219 & 0.253   \\
    \midrule
    ETTm1  &   0.665 & 0.734 & 0.441 & 0.483 & 0.254 & 0.344 & 0.314 & 0.444 & 0.185 & 0.232 & 0.273 & 0.373   \\
    \midrule
    ETTm2  &   0.138 & 0.135 & 0.143 & 0.153 & 0.104 & 0.107 & 0.119 & 0.148 & 0.084 & 0.090 & 0.096 & 0.112  \\
    \midrule
    ECL  &   0.226 & 0.222 & 0.230 & 0.230 & 0.221 & 0.224 & 0.226 & 0.231 & 0.200 & 0.207 & 0.209 & 0.211    \\
    \midrule
    Traffic  & 0.802 & 0.794 & 0.801 & 0.809 & 0.801 & 0.791 & 0.798 & 0.805 & 0.773 & 0.775 & 0.624 & 0.565  \\
    \midrule
    Weather  & 0.155 & 0.141 & 0.162 & 0.168 & 0.135 & 0.124 & 0.132 & 0.157 & 0.104 & 0.114 & 0.111 & 0.127  \\
    \midrule
    PEMS03 &  0.173 & 0.192 & 0.239 & 0.321 & 0.156 & 0.291 & 0.254 & 0.318 & 0.142 & 0.148 & 0.195 & 0.273   \\
    \midrule
    PEMS04 &  0.215 & 0.243 & 0.291 & 0.379 & 0.179 & 0.222 & 0.266 & 0.350 & 0.123 & 0.167 & 0.210 & 0.285  \\
    \midrule
    PEMS07 & 0.166 & 0.195 & 0.247 & 0.335 & 0.161 & 0.195 & 0.253 & 0.310 & 0.113 & 0.142 & 0.191 & 0.272   \\
    \midrule
    PEMS08 &  0.293 & 0.265 & 0.346 & 0.404 & 0.210 & 0.214 & 0.309 & 0.378 & 0.147 & 0.185 & 0.239 & 0.326  \\
    
    \bottomrule
  \end{tabular}
  \end{sc}
  \end{small}
\end{table*}

\subsection{Anomaly Detection}

In this section, we provide detailed results of anomaly detection in Table~\ref{tab:anomaly_detection}, including the results of Timer from scratch and pre-trained. We conducted experiments on all $250$ datasets of UCR Anomaly Archive and calculated the corresponding $\alpha$ quantiles. The results show that the pre-trained Timer can detect time series anomalies with smaller $\alpha$ on most datasets.

\begin{table}[ht]
\caption{Full results of anomaly detection on UCR Anomaly Archive, which contains $250$ datasets (arranged in 25 rows for a total of 10 rows). We provide the quantile ($\%$) of each dataset, where the bold parts represent the better results that benefited from pre-training.}
\vskip 0.1in
\label{tab:anomaly_detection}
\setlength{\tabcolsep}{2.5pt}
\renewcommand{\arraystretch}{1.5}
\begin{sc}
\resizebox{\linewidth}{!}{
\begin{tabular}{c|c|ccccccccccccccccccccccccc}
\toprule[1.2pt]  
\multicolumn{2}{c}{\scalebox{1.35}{Index}}
& \scalebox{1.35}{1} & \scalebox{1.35}{2} & \scalebox{1.35}{3} & 
\scalebox{1.35}{4} & \scalebox{1.35}{5} & 
\scalebox{1.35}{6} & \scalebox{1.35}{7} & 
\scalebox{1.35}{8} & \scalebox{1.35}{9} & 
\scalebox{1.35}{10} & \scalebox{1.35}{11} & \scalebox{1.35}{12} & \scalebox{1.35}{13} & 
\scalebox{1.35}{14} & \scalebox{1.35}{15} & 
\scalebox{1.35}{16} & \scalebox{1.35}{17} & 
\scalebox{1.35}{18} & \scalebox{1.35}{19} & 
\scalebox{1.35}{20} & \scalebox{1.35}{21} & \scalebox{1.35}{\textbf{22}} & \scalebox{1.35}{23} & 
\scalebox{1.35}{24} & \scalebox{1.35}{25}\\ 
\toprule[1.2pt]
\multirow{10}{*}{\scalebox{1.35}{\rotatebox{90}{Timer (From Scratch)}}}
& \scalebox{1.35}{1}  
 & \scalebox{1.35}{1.1} & \scalebox{1.35}{16.2} & \scalebox{1.35}{6.7} & \scalebox{1.35}{1.2} & \scalebox{1.35}{19.0} & \scalebox{1.35}{23.8} & \scalebox{1.35}{16.7} & \scalebox{1.35}{19.0} & \scalebox{1.35}{14.3} & \scalebox{1.35}{2.4} & \scalebox{1.35}{0.5} & \scalebox{1.35}{13.3} & \scalebox{1.35}{2.0} & \scalebox{1.35}{1.8} & \scalebox{1.35}{0.0} & \scalebox{1.35}{0.4} & \scalebox{1.35}{0.4} & \scalebox{1.35}{30.7} & \scalebox{1.35}{3.0} & \scalebox{1.35}{3.0} & \scalebox{1.35}{7.6} & \scalebox{1.35}{1.3} & \scalebox{1.35}{3.0} & \scalebox{1.35}{47.6} & \scalebox{1.35}{14.3}
 \\
& \scalebox{1.35}{2}  
 & \scalebox{1.35}{1.7} & \scalebox{1.35}{81.7} & \scalebox{1.35}{28.3} & \scalebox{1.35}{25.0} & \scalebox{1.35}{2.4} & \scalebox{1.35}{2.1} & \scalebox{1.35}{1.7} & \scalebox{1.35}{3.3} & \scalebox{1.35}{1.7} & \scalebox{1.35}{4.2} & \scalebox{1.35}{13.3} & \scalebox{1.35}{0.4} & \scalebox{1.35}{8.1} & \scalebox{1.35}{8.9} & \scalebox{1.35}{28.5} & \scalebox{1.35}{7.7} & \scalebox{1.35}{89.7} & \scalebox{1.35}{0.7} & \scalebox{1.35}{0.9} & \scalebox{1.35}{22.8} & \scalebox{1.35}{15.3} & \scalebox{1.35}{4.2} & \scalebox{1.35}{2.6} & \scalebox{1.35}{2.6} & \scalebox{1.35}{9.0}
\\
& \scalebox{1.35}{3} 
 & \scalebox{1.35}{5.1} & \scalebox{1.35}{1.3} & \scalebox{1.35}{62.5} & \scalebox{1.35}{2.8} & \scalebox{1.35}{9.5} & \scalebox{1.35}{19.9} & \scalebox{1.35}{71.8} & \scalebox{1.35}{16.0} & \scalebox{1.35}{5.0} & \scalebox{1.35}{2.7} & \scalebox{1.35}{7.4} & \scalebox{1.35}{12.8} & \scalebox{1.35}{4.1} & \scalebox{1.35}{1.2} & \scalebox{1.35}{53.0} & \scalebox{1.35}{3.3} & \scalebox{1.35}{33.3} & \scalebox{1.35}{94.0} & \scalebox{1.35}{1.5} & \scalebox{1.35}{0.3} & \scalebox{1.35}{0.3} & \scalebox{1.35}{20.6} & \scalebox{1.35}{1.0} & \scalebox{1.35}{30.0} & \scalebox{1.35}{16.3}
\\
& \scalebox{1.35}{4} 
 & \scalebox{1.35}{0.1} & \scalebox{1.35}{24.3} & \scalebox{1.35}{18.9} & \scalebox{1.35}{11.7} & \scalebox{1.35}{21.9} & \scalebox{1.35}{92.6} & \scalebox{1.35}{25.3} & \scalebox{1.35}{18.7} & \scalebox{1.35}{0.3} & \scalebox{1.35}{5.2} & \scalebox{1.35}{0.2} & \scalebox{1.35}{3.4} & \scalebox{1.35}{1.3} & \scalebox{1.35}{11.1} & \scalebox{1.35}{15.1} & \scalebox{1.35}{16.3} & \scalebox{1.35}{10.5} & \scalebox{1.35}{0.4} & \scalebox{1.35}{1.2} & \scalebox{1.35}{23.8} & \scalebox{1.35}{0.4} & \scalebox{1.35}{3.0} & \scalebox{1.35}{95.0} & \scalebox{1.35}{1.7} & \scalebox{1.35}{0.4}
\\ 
& \scalebox{1.35}{5} 
 & \scalebox{1.35}{32.5} & \scalebox{1.35}{0.7} & \scalebox{1.35}{1.3} & \scalebox{1.35}{62.7} & \scalebox{1.35}{21.7} & \scalebox{1.35}{1.3} & \scalebox{1.35}{43.8} & \scalebox{1.35}{36.5} & \scalebox{1.35}{0.6} & \scalebox{1.35}{9.1} & \scalebox{1.35}{3.5} & \scalebox{1.35}{1.2} & \scalebox{1.35}{19.0} & \scalebox{1.35}{21.4} & \scalebox{1.35}{14.3} & \scalebox{1.35}{23.8} & \scalebox{1.35}{16.7} & \scalebox{1.35}{2.4} & \scalebox{1.35}{0.5} & \scalebox{1.35}{14.0} & \scalebox{1.35}{2.0} & \scalebox{1.35}{1.3} & \scalebox{1.35}{0.0} & \scalebox{1.35}{0.4} & \scalebox{1.35}{0.4}
\\ 
& \scalebox{1.35}{6}  
 & \scalebox{1.35}{22.8} & \scalebox{1.35}{4.5} & \scalebox{1.35}{3.0} & \scalebox{1.35}{3.0} & \scalebox{1.35}{1.3} & \scalebox{1.35}{3.0} & \scalebox{1.35}{40.5} & \scalebox{1.35}{4.8} & \scalebox{1.35}{1.7} & \scalebox{1.35}{96.7} & \scalebox{1.35}{30.0} & \scalebox{1.35}{27.1} & \scalebox{1.35}{2.4} & \scalebox{1.35}{2.1} & \scalebox{1.35}{1.7} & \scalebox{1.35}{3.3} & \scalebox{1.35}{1.7} & \scalebox{1.35}{6.2} & \scalebox{1.35}{10.0} & \scalebox{1.35}{0.4} & \scalebox{1.35}{9.7} & \scalebox{1.35}{9.3} & \scalebox{1.35}{29.7} & \scalebox{1.35}{0.9} & \scalebox{1.35}{91.0}
\\
& \scalebox{1.35}{7} 
 & \scalebox{1.35}{0.7} & \scalebox{1.35}{1.4} & \scalebox{1.35}{23.5} & \scalebox{1.35}{11.1} & \scalebox{1.35}{4.2} & \scalebox{1.35}{2.6} & \scalebox{1.35}{1.3} & \scalebox{1.35}{17.9} & \scalebox{1.35}{5.1} & \scalebox{1.35}{1.3} & \scalebox{1.35}{68.8} & \scalebox{1.35}{2.8} & \scalebox{1.35}{9.5} & \scalebox{1.35}{15.8} & \scalebox{1.35}{59.1} & \scalebox{1.35}{11.9} & \scalebox{1.35}{2.6} & \scalebox{1.35}{0.5} & \scalebox{1.35}{16.4} & \scalebox{1.35}{20.7} & \scalebox{1.35}{1.3} & \scalebox{1.35}{0.5} & \scalebox{1.35}{36.4} & \scalebox{1.35}{3.3} & \scalebox{1.35}{37.5}
\\
& \scalebox{1.35}{8} 
 & \scalebox{1.35}{96.4} & \scalebox{1.35}{1.5} & \scalebox{1.35}{0.3} & \scalebox{1.35}{0.3} & \scalebox{1.35}{15.0} & \scalebox{1.35}{0.7} & \scalebox{1.35}{30.3} & \scalebox{1.35}{23.1} & \scalebox{1.35}{0.1} & \scalebox{1.35}{17.9} & \scalebox{1.35}{16.7} & \scalebox{1.35}{29.3} & \scalebox{1.35}{28.6} & \scalebox{1.35}{57.1} & \scalebox{1.35}{14.8} & \scalebox{1.35}{21.3} & \scalebox{1.35}{0.3} & \scalebox{1.35}{4.6} & \scalebox{1.35}{0.2} & \scalebox{1.35}{3.2} & \scalebox{1.35}{1.6} & \scalebox{1.35}{15.0} & \scalebox{1.35}{14.1} & \scalebox{1.35}{14.8} & \scalebox{1.35}{9.4}
\\
& \scalebox{1.35}{9} 
 & \scalebox{1.35}{0.3} & \scalebox{1.35}{21.3} & \scalebox{1.35}{79.6} & \scalebox{1.35}{2.9} & \scalebox{1.35}{14.1} & \scalebox{1.35}{4.0} & \scalebox{1.35}{1.6} & \scalebox{1.35}{6.0} & \scalebox{1.35}{0.1} & \scalebox{1.35}{49.3} & \scalebox{1.35}{31.7} & \scalebox{1.35}{30.1} & \scalebox{1.35}{0.0} & \scalebox{1.35}{0.0} & \scalebox{1.35}{0.1} & \scalebox{1.35}{0.0} & \scalebox{1.35}{0.2} & \scalebox{1.35}{0.0} & \scalebox{1.35}{0.0} & \scalebox{1.35}{0.0} & \scalebox{1.35}{0.0} & \scalebox{1.35}{0.1} & \scalebox{1.35}{17.5} & \scalebox{1.35}{13.3} & \scalebox{1.35}{7.3}
\\
& \scalebox{1.35}{10}   
& \scalebox{1.35}{0.1} & \scalebox{1.35}{10.4} & \scalebox{1.35}{1.0} & \scalebox{1.35}{10.0} & \scalebox{1.35}{18.6} & \scalebox{1.35}{16.1} & \scalebox{1.35}{9.6} & \scalebox{1.35}{0.9} & \scalebox{1.35}{40.1} & \scalebox{1.35}{0.7} & \scalebox{1.35}{24.4} & \scalebox{1.35}{0.7} & \scalebox{1.35}{2.5} & \scalebox{1.35}{2.4} & \scalebox{1.35}{9.8} & \scalebox{1.35}{1.1} & \scalebox{1.35}{7.4} & \scalebox{1.35}{7.6} & \scalebox{1.35}{1.2} & \scalebox{1.35}{0.9} & \scalebox{1.35}{4.4} & \scalebox{1.35}{10.9} & \scalebox{1.35}{19.7} & \scalebox{1.35}{53.8} & \scalebox{1.35}{30.6}
\\ \midrule
\multirow{10}{*}{\scalebox{1.35}{\rotatebox{90}{Timer (Pre-trained)}}}
& \scalebox{1.35}{1}  
 & \scalebox{1.35}{3.9} & \scalebox{1.35}{\textbf{5.8}} & \scalebox{1.35}{\textbf{1.5}} & \scalebox{1.35}{\textbf{1.2}} & \scalebox{1.35}{23.8} & \scalebox{1.35}{\textbf{19.0}} & \scalebox{1.35}{\textbf{7.1}} & \scalebox{1.35}{47.6} & \scalebox{1.35}{\textbf{11.9}} & \scalebox{1.35}{\textbf{2.4}} & \scalebox{1.35}{\textbf{0.5}} & \scalebox{1.35}{\textbf{6.7}} & \scalebox{1.35}{\textbf{2.0}} & \scalebox{1.35}{32.9} & \scalebox{1.35}{\textbf{0.0}} & \scalebox{1.35}{\textbf{0.4}} & \scalebox{1.35}{\textbf{0.4}} & \scalebox{1.35}{\textbf{26.8}} & \scalebox{1.35}{\textbf{3.0}} & \scalebox{1.35}{\textbf{3.0}} & \scalebox{1.35}{\textbf{6.1}} & \scalebox{1.35}{\textbf{1.3}} & \scalebox{1.35}{\textbf{1.5}} & \scalebox{1.35}{\textbf{2.4}} & \scalebox{1.35}{\textbf{4.8}}
\\
& \scalebox{1.35}{2}  
 & \scalebox{1.35}{\textbf{1.7}} & \scalebox{1.35}{\textbf{10.0}} & \scalebox{1.35}{\textbf{3.3}} & \scalebox{1.35}{\textbf{6.2}} & \scalebox{1.35}{4.8} & \scalebox{1.35}{\textbf{2.1}} & \scalebox{1.35}{\textbf{1.7}} & \scalebox{1.35}{\textbf{3.3}} & \scalebox{1.35}{\textbf{1.7}} & \scalebox{1.35}{\textbf{2.1}} & \scalebox{1.35}{\textbf{6.7}} & \scalebox{1.35}{\textbf{0.4}} & \scalebox{1.35}{\textbf{2.3}} & \scalebox{1.35}{\textbf{4.7}} & \scalebox{1.35}{28.9} & \scalebox{1.35}{\textbf{3.4}} & \scalebox{1.35}{\textbf{89.3}} & \scalebox{1.35}{\textbf{0.7}} & \scalebox{1.35}{1.4} & \scalebox{1.35}{\textbf{21.0}} & \scalebox{1.35}{\textbf{9.0}} & \scalebox{1.35}{\textbf{3.3}} & \scalebox{1.35}{\textbf{2.6}} & \scalebox{1.35}{\textbf{2.6}} & \scalebox{1.35}{\textbf{7.7}}
\\
& \scalebox{1.35}{3} 
 & \scalebox{1.35}{\textbf{5.1}} & \scalebox{1.35}{\textbf{1.3}} & \scalebox{1.35}{72.9} & \scalebox{1.35}{\textbf{2.8}} & \scalebox{1.35}{\textbf{2.4}} & \scalebox{1.35}{36.5} & \scalebox{1.35}{\textbf{45.5}} & \scalebox{1.35}{22.4} & \scalebox{1.35}{6.3} & \scalebox{1.35}{3.4} & \scalebox{1.35}{9.8} & \scalebox{1.35}{18.8} & \scalebox{1.35}{7.7} & \scalebox{1.35}{1.9} & \scalebox{1.35}{63.6} & \scalebox{1.35}{8.3} & \scalebox{1.35}{56.2} & \scalebox{1.35}{\textbf{77.4}} & \scalebox{1.35}{\textbf{1.5}} & \scalebox{1.35}{\textbf{0.3}} & \scalebox{1.35}{\textbf{0.3}} & \scalebox{1.35}{24.7} & \scalebox{1.35}{\textbf{0.7}} & \scalebox{1.35}{\textbf{15.8}} & \scalebox{1.35}{\textbf{8.7}}
\\
& \scalebox{1.35}{4} 
 & \scalebox{1.35}{0.6} & \scalebox{1.35}{\textbf{0.3}} & \scalebox{1.35}{19.0} & \scalebox{1.35}{14.3} & \scalebox{1.35}{66.9} & \scalebox{1.35}{92.9} & \scalebox{1.35}{27.8} & \scalebox{1.35}{\textbf{14.3}} & \scalebox{1.35}{\textbf{0.3}} & \scalebox{1.35}{\textbf{3.9}} & \scalebox{1.35}{\textbf{0.2}} & \scalebox{1.35}{5.9} & \scalebox{1.35}{6.0} & \scalebox{1.35}{17.0} & \scalebox{1.35}{\textbf{5.6}} & \scalebox{1.35}{\textbf{13.9}} & \scalebox{1.35}{\textbf{5.3}} & \scalebox{1.35}{0.9} & \scalebox{1.35}{\textbf{1.2}} & \scalebox{1.35}{38.1} & \scalebox{1.35}{\textbf{0.4}} & \scalebox{1.35}{\textbf{1.5}} & \scalebox{1.35}{\textbf{18.3}} & \scalebox{1.35}{\textbf{1.7}} & \scalebox{1.35}{\textbf{0.4}}
\\ 
& \scalebox{1.35}{5} 
 & \scalebox{1.35}{\textbf{31.7}} & \scalebox{1.35}{\textbf{0.7}} & \scalebox{1.35}{\textbf{1.3}} & \scalebox{1.35}{\textbf{26.4}} & \scalebox{1.35}{\textbf{12.4}} & \scalebox{1.35}{1.9} & \scalebox{1.35}{54.2} & \scalebox{1.35}{77.5} & \scalebox{1.35}{\textbf{0.6}} & \scalebox{1.35}{\textbf{1.5}} & \scalebox{1.35}{\textbf{0.6}} & \scalebox{1.35}{\textbf{1.2}} & \scalebox{1.35}{\textbf{16.7}} & \scalebox{1.35}{\textbf{16.7}} & \scalebox{1.35}{\textbf{4.8}} & \scalebox{1.35}{35.7} & \scalebox{1.35}{\textbf{9.5}} & \scalebox{1.35}{\textbf{2.4}} & \scalebox{1.35}{\textbf{0.5}} & \scalebox{1.35}{\textbf{7.3}} & \scalebox{1.35}{\textbf{2.0}} & \scalebox{1.35}{29.4} & \scalebox{1.35}{\textbf{0.0}} & \scalebox{1.35}{\textbf{0.4}} & \scalebox{1.35}{\textbf{0.4}}
\\ 
& \scalebox{1.35}{6}  
 & \scalebox{1.35}{\textbf{20.6}} & \scalebox{1.35}{\textbf{3.0}} & \scalebox{1.35}{\textbf{3.0}} & \scalebox{1.35}{\textbf{1.5}} & \scalebox{1.35}{\textbf{1.3}} & \scalebox{1.35}{\textbf{1.5}} & \scalebox{1.35}{\textbf{2.4}} & \scalebox{1.35}{\textbf{2.4}} & \scalebox{1.35}{\textbf{1.7}} & \scalebox{1.35}{\textbf{18.3}} & \scalebox{1.35}{\textbf{1.7}} & \scalebox{1.35}{\textbf{6.2}} & \scalebox{1.35}{\textbf{2.4}} & \scalebox{1.35}{\textbf{2.1}} & \scalebox{1.35}{\textbf{1.7}} & \scalebox{1.35}{\textbf{3.3}} & \scalebox{1.35}{\textbf{1.7}} & \scalebox{1.35}{\textbf{2.1}} & \scalebox{1.35}{\textbf{6.7}} & \scalebox{1.35}{\textbf{0.4}} & \scalebox{1.35}{\textbf{2.7}} & \scalebox{1.35}{\textbf{4.3}} & \scalebox{1.35}{\textbf{28.5}} & \scalebox{1.35}{1.3} & \scalebox{1.35}{\textbf{90.6}}
\\
& \scalebox{1.35}{7} 
 & \scalebox{1.35}{\textbf{0.7}} & \scalebox{1.35}{\textbf{0.9}} & \scalebox{1.35}{\textbf{21.0}} & \scalebox{1.35}{\textbf{9.0}} & \scalebox{1.35}{\textbf{3.3}} & \scalebox{1.35}{\textbf{1.3}} & \scalebox{1.35}{\textbf{1.3}} & \scalebox{1.35}{\textbf{3.8}} & \scalebox{1.35}{\textbf{5.1}} & \scalebox{1.35}{\textbf{1.3}} & \scalebox{1.35}{85.4} & \scalebox{1.35}{\textbf{2.8}} & \scalebox{1.35}{\textbf{3.1}} & \scalebox{1.35}{28.9} & \scalebox{1.35}{\textbf{25.5}} & \scalebox{1.35}{20.7} & \scalebox{1.35}{3.0} & \scalebox{1.35}{4.5} & \scalebox{1.35}{\textbf{13.3}} & \scalebox{1.35}{\textbf{15.8}} & \scalebox{1.35}{2.6} & \scalebox{1.35}{1.2} & \scalebox{1.35}{47.0} & \scalebox{1.35}{\textbf{3.3}} & \scalebox{1.35}{47.9}
\\
& \scalebox{1.35}{8} 
 & \scalebox{1.35}{\textbf{47.6}} & \scalebox{1.35}{\textbf{1.5}} & \scalebox{1.35}{\textbf{0.3}} & \scalebox{1.35}{\textbf{0.3}} & \scalebox{1.35}{19.7} & \scalebox{1.35}{1.0} & \scalebox{1.35}{\textbf{13.3}} & \scalebox{1.35}{\textbf{7.2}} & \scalebox{1.35}{0.1} & \scalebox{1.35}{\textbf{0.1}} & \scalebox{1.35}{\textbf{15.6}} & \scalebox{1.35}{96.5} & \scalebox{1.35}{96.3} & \scalebox{1.35}{74.1} & \scalebox{1.35}{\textbf{13.7}} & \scalebox{1.35}{\textbf{19.0}} & \scalebox{1.35}{\textbf{0.3}} & \scalebox{1.35}{\textbf{3.3}} & \scalebox{1.35}{\textbf{0.2}} & \scalebox{1.35}{5.4} & \scalebox{1.35}{8.9} & \scalebox{1.35}{\textbf{13.4}} & \scalebox{1.35}{\textbf{14.1}} & \scalebox{1.35}{\textbf{14.5}} & \scalebox{1.35}{\textbf{4.2}}
\\
& \scalebox{1.35}{9} 
 & \scalebox{1.35}{\textbf{0.3}} & \scalebox{1.35}{\textbf{1.2}} & \scalebox{1.35}{\textbf{63.5}} & \scalebox{1.35}{\textbf{0.3}} & \scalebox{1.35}{19.2} & \scalebox{1.35}{\textbf{1.7}} & \scalebox{1.35}{\textbf{0.3}} & \scalebox{1.35}{18.6} & \scalebox{1.35}{\textbf{0.1}} & \scalebox{1.35}{\textbf{29.8}} & \scalebox{1.35}{38.9} & \scalebox{1.35}{\textbf{16.2}} & \scalebox{1.35}{\textbf{0.0}} & \scalebox{1.35}{\textbf{0.0}} & \scalebox{1.35}{\textbf{0.0}} & \scalebox{1.35}{\textbf{0.0}} & \scalebox{1.35}{6.5} & \scalebox{1.35}{\textbf{0.0}} & \scalebox{1.35}{\textbf{0.0}} & \scalebox{1.35}{\textbf{0.0}} & \scalebox{1.35}{\textbf{0.0}} & \scalebox{1.35}{\textbf{0.1}} & \scalebox{1.35}{\textbf{16.5}} & \scalebox{1.35}{\textbf{11.0}} & \scalebox{1.35}{\textbf{5.7}}
\\
& \scalebox{1.35}{10}
 & \scalebox{1.35}{\textbf{0.1}} & \scalebox{1.35}{\textbf{9.2}} & \scalebox{1.35}{\textbf{0.6}} & \scalebox{1.35}{\textbf{8.0}} & \scalebox{1.35}{\textbf{6.9}} & \scalebox{1.35}{\textbf{9.8}} & \scalebox{1.35}{\textbf{1.7}} & \scalebox{1.35}{1.2} & \scalebox{1.35}{\textbf{38.6}} & \scalebox{1.35}{1.1} & \scalebox{1.35}{\textbf{23.3}} & \scalebox{1.35}{\textbf{0.7}} & \scalebox{1.35}{\textbf{1.5}} & \scalebox{1.35}{\textbf{2.3}} & \scalebox{1.35}{10.2} & \scalebox{1.35}{1.1} & \scalebox{1.35}{8.5} & \scalebox{1.35}{\textbf{7.4}} & \scalebox{1.35}{\textbf{0.7}} & \scalebox{1.35}{\textbf{0.6}} & \scalebox{1.35}{5.5} & \scalebox{1.35}{17.0} & \scalebox{1.35}{28.8} & \scalebox{1.35}{\textbf{39.7}} & \scalebox{1.35}{\textbf{9.7}}
\\ \bottomrule[1.2pt]          
\end{tabular}}
\end{sc}
\end{table}
\subsection{Scalability}

We provide detailed downstream forecasting results conducted on PEMS subsets with the scaling of model size (Figure~\ref{fig:scaleup_model}) and data size (Figure~\ref{fig:scaleup_dataset}). As shown in Table~\ref{tab:scale_up_full}, it supports the scalability of our decoder-only Timer trained in GPT-style, following the scaling law~\cite{kaplan2020scaling} towards large time series models.

\begin{table*}[ht]
  \caption{Detailed results for scaling up the pre-trained scale and the parameter of Timer.}
  \vskip 0.1in
  \label{tab:scale_up_full}
  \footnotesize
  \begin{small}
  \begin{sc}
  \linespread{2}
  \renewcommand{\multirowsetup}{\centering}
  \setlength{\tabcolsep}{7.6pt}
  \renewcommand\arraystretch{1.4}
  \begin{tabular}{c|c|c|c|c|c|c|c|c|c|c|c|c}
    \toprule
    \multicolumn{2}{c|}{Pre-trained} & \multicolumn{4}{c|}{4G} & \multicolumn{3}{c|}{4G} & 1G & 2G & 4G & 12G \\
    \cmidrule{1-2} \cmidrule{3-6} \cmidrule{7-9} \cmidrule{10-13}
    \multicolumn{2}{c|}{Model Dim.} & \multicolumn{4}{c|}{256} & 512 & 768 & 1024 & \multicolumn{4}{c}{1024} \\
    \cmidrule{1-2} \cmidrule{3-6} \cmidrule{7-9} \cmidrule{10-13}
    \multicolumn{2}{c|}{Layers} & 2 & 4 & 6 & 8 & \multicolumn{3}{c|}{6} & \multicolumn{4}{c}{8} \\
    \toprule
    \multirow{4}{*}{\rotatebox{90}{5\% Samples}}
    & PEMS03 &  0.188 & 0.174 & 0.168 & \textbf{0.160} & 0.146 & 0.138 & \textbf{0.133} & 0.130 & 0.128 & 0.128 & \textbf{0.125}   \\
    & PEMS04 &   0.223 & 0.208 & 0.200 & \textbf{0.190} & 0.166 & 0.154 & \textbf{0.145} & 0.145 & 0.142 & 0.143 & \textbf{0.135} \\
    & PEMS07 & 0.147 & 0.131 & 0.123 & \textbf{0.120} & 0.106 & 0.097 & \textbf{0.092} & 0.092 & 0.090 & 0.090 & \textbf{0.087}   \\
    & PEMS08 &  0.367 & 0.339 & 0.322 & \textbf{0.319} & 0.289 & 0.256 & \textbf{0.239} & 0.228 & 0.221 & 0.216 & \textbf{0.204} \\
    \midrule
    \multirow{4}{*}{\rotatebox{90}{20\% Samples}}
    & PEMS03 &  0.154 & 0.141 & 0.137 & \textbf{0.134} & 0.127 & 0.124 & \textbf{0.123} & 0.121 & 0.120 & 0.117 & \textbf{0.114}   \\
    & PEMS04 &   0.182 & 0.162 & 0.155 & \textbf{0.150} & 0.140 & 0.132 & \textbf{0.124} & 0.124 & 0.123 & 0.122 & \textbf{0.115} \\
    & PEMS07 & 0.115 & 0.104 & 0.098 & \textbf{0.095} & 0.086 & 0.082 & \textbf{0.080} & 0.079 & 0.079 & 0.078 & \textbf{0.076}   \\
    & PEMS08 &  0.326 & 0.277 & 0.247 & \textbf{0.238} & 0.206 & \textbf{0.193} & 0.194 & 0.193 & 0.185 & \textbf{0.185} & 0.187 \\
    \bottomrule
  \end{tabular}
  \end{sc}
  \end{small}
\end{table*}

\begin{table*}[ht]
  \vspace{-5pt}
  \caption{Additional downstream $672$-pred-$96$ forecasting results of the subsets of PEMS and ETT under different data scarcity of the encoder-only and decoder-only Transformer. The bold part indicates that the result performs best in the current dataset and sample ratio.}
  \vskip 0.1in
  \label{tab:forecast_full}
  \footnotesize
  \begin{small}
  \begin{sc}
  \linespread{2}
  \renewcommand{\multirowsetup}{\centering}
  \setlength{\tabcolsep}{5.8pt}
  \renewcommand\arraystretch{1.4}
  \begin{tabular}{c|cc|cc|cc|cc|cc|cc}
    \toprule
    Scenario & \multicolumn{4}{c|}{1\% Target} & \multicolumn{4}{c|}{5\% Target} & \multicolumn{4}{c}{20\% Target} \\
    \cmidrule(lr){1-1} \cmidrule(lr){2-5} \cmidrule(lr){6-9} \cmidrule(lr){10-13}
    Architecture & \multicolumn{2}{c|}{Encoder} & \multicolumn{2}{c|}{Decoder} & \multicolumn{2}{c|}{Encoder} & \multicolumn{2}{c|}{Decoder} & \multicolumn{2}{c|}{Encoder} & \multicolumn{2}{c}{Decoder}  \\ 
    \cmidrule(lr){1-1} \cmidrule(lr){2-3} \cmidrule(lr){4-5} \cmidrule(lr){6-7} \cmidrule(lr){8-9} \cmidrule(lr){10-11} \cmidrule(lr){12-13}
    Pre-trained & None & 12G & None & 12G & None & 12G & None & 12G & None & 12G & None & 12G  \\
    \toprule
    ETTh1  &  0.446 & 0.413 & 0.428 & \textbf{0.366} & 0.437 & 0.405 & 0.426 & \textbf{0.362} & 0.409 & 0.404 & 0.385 & \textbf{0.359}  \\
    \midrule
    ETTh2  &   0.338 & 0.304 & 0.315 & \textbf{0.284} & 0.329 & 0.293 & 0.314 & \textbf{0.280} & 0.308 & 0.299 & 0.294 & \textbf{0.284}  \\
    \midrule
    ETTm1  &   0.463 & 0.370 & 0.407 & \textbf{0.345} & 0.391 & 0.340 & 0.354 & \textbf{0.321} & 0.344 & 0.323 & 0.332 & \textbf{0.321} \\
    \midrule
    ETTm2  &  0.220 & \textbf{0.181} & 0.207 & 0.183 & 0.197 & \textbf{0.174} & 0.190 & 0.176 & \textbf{0.177} & 0.179 & \textbf{0.177} & 0.187  \\
    \midrule
    PEMS03 &   0.225 & 0.196 & 0.249 & \textbf{0.151} & 0.165 & 0.160 & 0.158 & \textbf{0.125} & 0.144 & 0.145 & 0.135 & \textbf{0.116} \\
    \midrule
    PEMS04 &   0.253 & 0.226 & 0.320 & \textbf{0.172} & 0.198 & 0.184 & 0.195 & \textbf{0.135} & 0.167 & 0.161 & 0.145 & \textbf{0.120} \\
    \midrule
    PEMS07 &  0.170 & 0.156 & 0.179 & \textbf{0.112} & 0.126 & 0.125 & 0.114 & \textbf{0.087} & 0.102 & 0.103 & 0.093 & \textbf{0.077}  \\
    \midrule
    PEMS08 &  0.496 & 0.405 & 0.563 & \textbf{0.286} & 0.389 & 0.319 & 0.391 & \textbf{0.204} & 0.280 & 0.246 & 0.241 & \textbf{0.193}  \\
    \bottomrule
  \end{tabular}
  \end{sc}
  \end{small}
\end{table*}

\subsection{Zero-shot forecasting}\label{sec:zero-shot-result}
In this section, we provide detailed results of zero-shot forecasting in Table~\ref{tab:forecast_baseline}. We conducted experiments on seven datasets that are not included in the pre-training corpora of LTSMs. The results show that the top-ranked LTSMs are Timer, Moirai, and TimesFM. The performance of probabilistic forecaster Chronos can be improved by sampling more trajectories. It can still be an issue that scaling behavior is not evident on some datasets in the zero-shot scenario, and the failure of multi-step prediction can also appear in some models, indicating the development of zero-shot LTSMs is still in the early stage.

\begin{table*}[ht]
  \vspace{-5pt}
  \caption{Zero-shot forecasting evaluation. We extensively evaluate available large time series models. We provided the average rank on all downstream datasets, where the lower is better. For the probabilistic forecaster Chronos: S1 means sampling one trajectory and S20 means sampling 20 trajectories and using the average. '-' indicates that multi-step error accumulation leads to failure predictions.}
  \vskip 0.1in
  \label{tab:forecast_baseline}
  \footnotesize
  \begin{small}
  \begin{sc}
  \linespread{2}
  \renewcommand{\multirowsetup}{\centering}
  \setlength{\tabcolsep}{3.5pt}
  \begin{tabular}{c|c|c|c|c|c|c|c|c|c|c}
    \toprule
    Model & \scalebox{0.8}{Timer-1B} & \scalebox{0.8}{Timer-16B} & \scalebox{0.8}{Timer-28B} & \scalebox{0.8}{Moirai-S} & \scalebox{0.8}{Moirai-M} & \scalebox{0.8}{Moirai-L} & \scalebox{0.8}{MOMENT} & \scalebox{0.8}{TimesFM} & \scalebox{0.8}{Chronos-S1} & \scalebox{0.8}{Chronos-S20}\\
    \toprule
    ETTh1 & 0.438  & 0.364  & 0.393  & 0.441  & 0.383  & 0.394  & 0.674 & 0.414  & 0.571 & 0.454\\
    \midrule
    ETTh2 & 0.314  & 0.294  & 0.308  & 0.295  & 0.295  & 0.293  & 0.330 & 0.318  & 0.423 & 0.326  \\
    \midrule
    ETTm1 & 0.690  & 0.766  & 0.420  & 0.562  & 0.448  & 0.452  & 0.670 & 0.354  & 0.632 & 0.451 \\
    \midrule
    ETTm2 & 0.213  & 0.234  & 0.247  & 0.218  & 0.225  & 0.214  & 0.257 & 0.201  & 0.272 & 0.190  \\
    \midrule
    ECL & 0.192  & 0.139  & 0.147  & 0.212  & 0.162  & 0.155  & 0.744 & - & - & -  \\
    \midrule
    Traffic & 0.458  & 0.399  & 0.414  & 0.616  & 0.425  & 0.399  & 1.293 & - & - & - \\
    \midrule
    Weather & 0.181  & 0.203  & 0.243  & 0.195  & 0.197  & 0.221  & 0.255 & - & - & - \\
    \toprule
    Rank (Avg.) &
    \multicolumn{3}{c|}{\textbf{1.571}}  & \multicolumn{3}{c|}{2.286} & \multicolumn{1}{c|}{4.429}  & \multicolumn{1}{c|}{2.250} & \multicolumn{1}{c|}{5.500} & \multicolumn{1}{c}{3.250}   \\
    \bottomrule
  \end{tabular}
  \end{sc}
  \vspace{-10pt}
  \end{small}
\end{table*}

\section{Showcase}
\begin{figure*}[ht]
\begin{center}
    \center{\includegraphics[width=\textwidth]{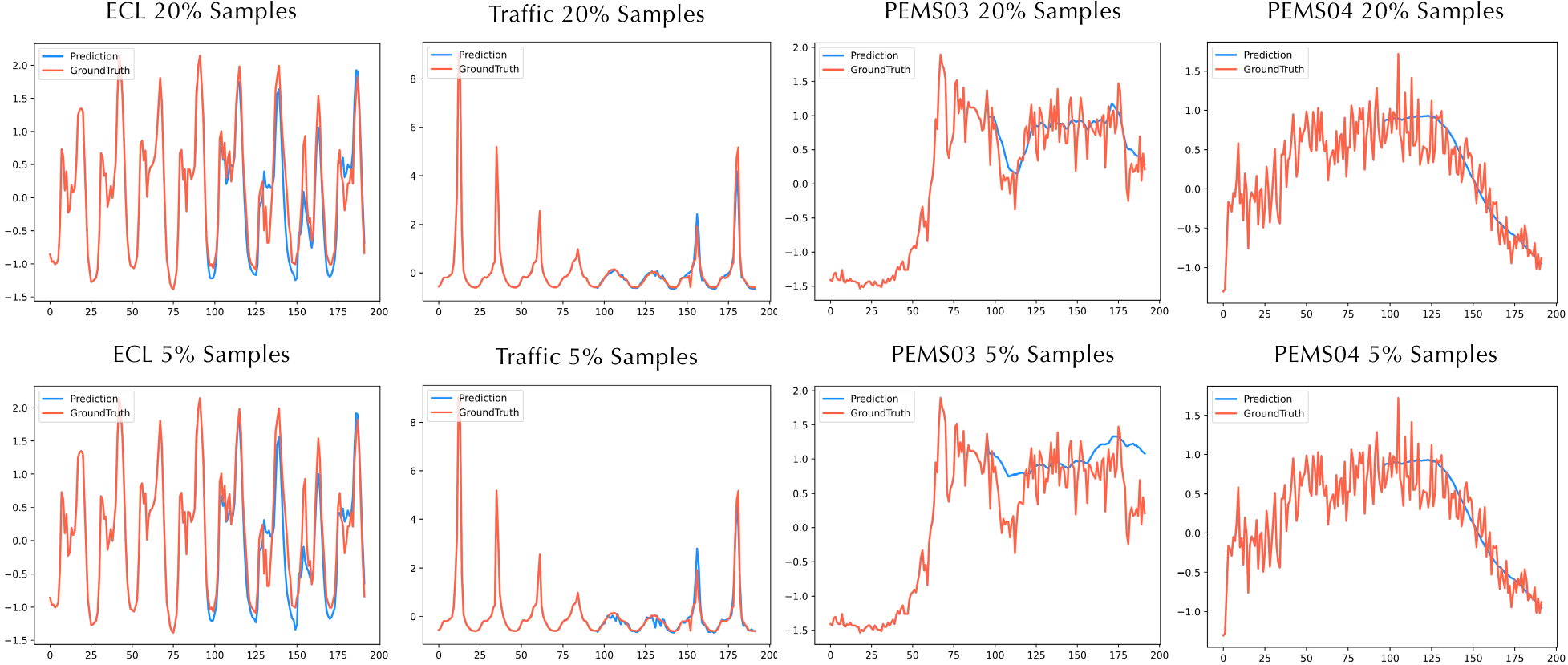}}
    \vspace{-20pt}
	\caption{Visualization of input-$672$-predict-$96$ forecasting results of Timer trained with $5\%$ and $20\%$ samples.}
	\label{fig:show_forecast}
\end{center}
\end{figure*}

\begin{figure*}[ht]
\begin{center}
    \center{\includegraphics[width=\textwidth]{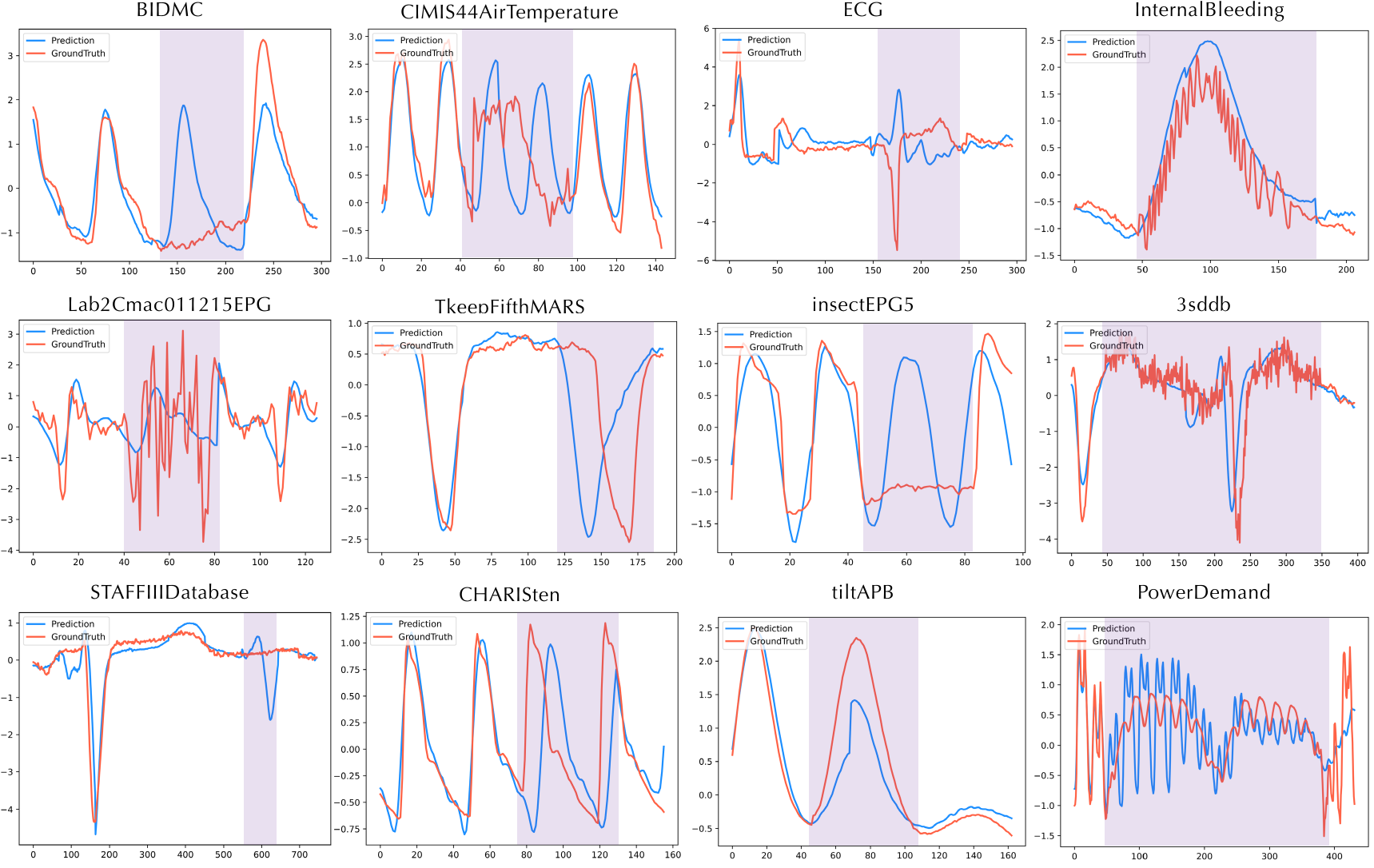}}
    \vspace{-20pt}
	\caption{Visualization of anomaly detection results of Timer on partial UCR Anomaly Archive~\cite{wu2021current}. The masked part represents the abnormal position, and the model locates the abnormal interval by generating results that deviate from the abnormal series.}
	\label{fig:show_anomaly_detection}
\end{center}
\end{figure*}

\begin{figure*}[ht]
\begin{center}
    \center{\includegraphics[width=\textwidth]{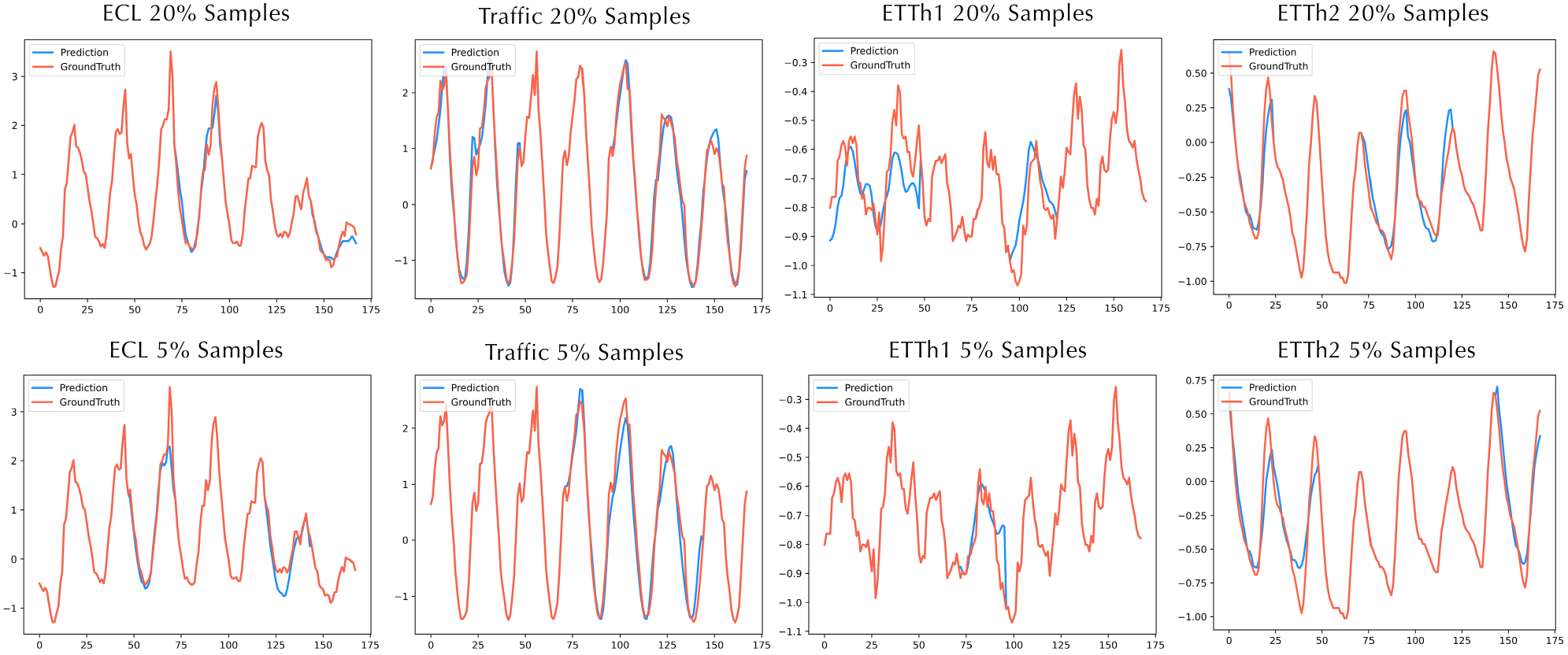}}
    \vspace{-20pt}
	\caption{Visualization of imputation results of Timer trained with $5\%$ and $20\%$ samples.}
	\label{fig:show_imputation}
\end{center}
\end{figure*}

To present a clear performance of our proposed Timer, we provide visualizations for downstream forecasting, imputation, and anomaly detection tasks in Figure~\ref{fig:show_forecast},~\ref{fig:show_anomaly_detection}~and~\ref{fig:show_imputation}. The forecasting and imputation contain experimental results at different sample ratios. For anomaly detection, we provide the position of the anomaly and the generated normal series by Timer.

\section{Limitations}
UTSD is constructed with hierarchical capacities. Though it is helpful to study the scalability of the model, it is not big enough since we have witnessed recent work claims the pre-training on ten and even a hundred billion time points. Therefore, we advocate for the ongoing expansion of data infrastructure while upholding high quality and hierarchy, which may significantly advance the time series community. In terms of the method, this work aims at an early but important development of large models. Despite the generalization, scalability, and task-generality that Timer has achieved, time series classification has not been unified in our generative formulations and Timer does not yet support probabilistic forecasting and specially adapts for multiple variables. It also leaves for better zero-shot generalization and advanced abilities, such as in-context learning and multi-modality, which are scheduled to be developed by ever-large pre-training.

\section{Societal Impacts}

\paragraph{Real-world applications} This paper develops large models for the field of time series. We present a general-purpose time series analysis model to handle data-scarce scenarios. Given the state-of-the-art performance of Timer, this model may be applied to many real-world applications, which helps our society prevent risks in advance and make better decisions with limited available samples. Our paper mainly focuses on scientific research and has no obvious negative social impact.

\paragraph{Academic research} 
In this paper, we release a high-quality dataset for scalable pre-training. Different from prior works, the dataset is not merely aggregation but follows deftly curation. Based on it, the research on scalable time series architectures and pre-training techniques can be facilitated. Towards large time series models, the proposed Timer shows its generalization and versatility in many tasks. The regime of generative pre-training and autoregression can be instructive for future research.

\end{document}